\documentclass[times,twocolumn,final]{elsarticle}
\usepackage[margin=2.0cm]{geometry}
\usepackage{hyperref}
\hypersetup{colorlinks,allcolors=black}
\usepackage{framed,multirow}

\usepackage{comment}
\newcommand{\mycomment}[1]{}
\usepackage[mathscr]{euscript}
\usepackage{blindtext, amsmath}
\usepackage{graphicx, tabto, hyperref}
\usepackage{algorithm,algpseudocode}
\usepackage{amssymb}
\usepackage{latexsym}
\setcitestyle{square}
\usepackage{url}
\usepackage{cite}
\usepackage{epsfig}
\usepackage[font=small,labelfont=bf]{caption}
\usepackage{xcolor}
\usepackage{ltablex,tabularx,setspace,arydshln}
\usepackage{subfigure,calc,amstext,amsthm,pslatex,multicol}
\usepackage{colortbl,graphics,textcomp}
\usepackage{enumitem}
\usepackage{booktabs}
\usepackage{bigstrut}
\usepackage{mathrsfs}

\usepackage[normalem]{ulem} 

\definecolor{darkgreen}{rgb}{0,.4,0}
\definecolor{darkcyan}{rgb}{0,.4,.4}

\journal{Arxiv}
\begin{document}
\begin{frontmatter}
\title{\textbf{A Layer-Wise Tokens-to-Token Transformer Network for Improved Historical Document Image Enhancement}}
\author[1]{Risab Biswas\corref{cor1}}
\cortext[cor1]{Corresponding author: Tel.: +91-90-645-37732}
\ead{risab.biswas@p360.com}
\author[2]{Swalpa Kumar Roy}
\ead{swalpa@cse.jgec.ac.in}

\author[3]{Umapada Pal}
\ead{umapada@isical.ac.in}
\address[1]{Artificial Intelligence Group, Optiks Innovations Pvt. Ltd (P360), Mumbai, Maharashtra 400066, India.}
\address[2]{Department of Computer Science and Engineering, Alipurduar Government Engineering and Management College, West Bengal 736206, India.}
\address[3]{Computer Vision and Pattern Recognition Unit, Indian Statistical Institute, Kolkata 700108, India.}

\begin{abstract}
Document image enhancement is a fundamental and important stage for attaining the best performance in any document analysis assignment because there are many degradation situations that could harm document images, making it more difficult to recognize and analyze them. In this paper, we propose \textbf{T2T-BinFormer} which is a novel document binarization encoder-decoder architecture based on a Tokens-to-token vision transformer. Each image is divided into a set of tokens with a defined length using the ViT model, which is then applied several times to model the global relationship between the tokens. However, the conventional tokenization of input data does not adequately reflect the crucial local structure between adjacent pixels of the input image, which results in low efficiency. Instead of using a simple ViT and hard splitting of images for the document image enhancement task, we employed a progressive tokenization technique to capture this local information from an image to achieve more effective results. Experiments on various DIBCO and H-DIBCO benchmarks demonstrate that the proposed model outperforms the existing CNN and ViT-based state-of-the-art methods. In this research, the primary area of examination is the application of the proposed architecture to the task of document binarization. The source code will be made available at {\url{https://github.com/RisabBiswas/T2T-BinFormer}}.
\end{abstract}
\end{frontmatter}
\section{Introduction}
\label{sec:Intro}
Studies on document image analysis and pattern recognition place high importance on preserving and improving the readability of document images, particularly historical versions. Streaks, blots, artifacts, pen strokes, ink bleed, inadequate maintenance, aging effects, warping effects, etc. can all be different types of degradation on document images. Because of this, the enhancement of document images remains a challenging task. These anomalies may have a considerable influence on the subsequent downstream data processing activities, such as optical character recognition, document layout analysis, etc.
Various thresholding techniques were used in early enhancement methods, such as~\citep{otsu1979threshold}, \citep{sauvola2000adaptive}. Convolutional neural network (CNN) \citep{lecun1995convolutional} based methods have recently been popular for tackling enhancement-related subproblems. Generative adversarial networks (GAN) \citep{goodfellow2014generative}, particularly contrastive GAN (cGAN), are also seen as an effective way to solve image generation issues. Transformers' recent success in natural language processing (NLP)~\citep{devlin2018bert} has raised interest in the possibility of these systems to solve computer vision problems including image recognition~\citep{dosovitskiy2020image}, object detection~\citep{carion2020end}, the identification of handwritten manuscripts~\citep{li2021trocr}, and other related problems. The proposed self-attention mechanism in~\citep{vaswani2017attention} facilitates the global linkages between contextual elements. Patches of a defined size are produced by the vision transformers (ViT's) from an image as opposed to CNNs, which evaluate pixel arrays. ViT divides an image into fixed-size patches e.g. ($4\times4$), ($8\times8$), ($16\times16$), etc., and embeds each of them into a latent representation (a.k.a tokens) while also incorporating the positional information which then becomes the input to the transformer encoder. The previously proposed method for document image binarization using an end-to-end ViT-based architecture has achieved significantly improved results compared to other methods, however, the architecture fails to capture the local information from the patches. 

We propose a novel document binarization framework that combines both for an improved document binarization task that achieves significant improvement over the existing methodologies. Our architecture draws inspiration from the idea that taking an image's local features into account, in conjunction with the global contextual information, can significantly help in improving the binarization results. Our hypothesis is that ViT's basic tokenization of input images using a hard split limits the model from effectively modeling the local structure of the images and thus provides an opportunity to further enhance the existing binarization framework based on ViT. Therefore, we incorporated the Tokens-to-Token ViT (T2T-ViT) in the encoding stage instead of the vanilla ViT. In this research, we backed up our hypothesis with experimental evidence as well. The proposed approach is aimed at extracting both the global and local features from the images which will serve as the basis to demonstrate that the simple tokenization in ViT-based architecture fails to capture the local knowledge from the image patches. These local features act as crucial information to the self-attention module of the ViT. At the conclusion of the encoding stage, the generated tokens contain both the global and local contextual information of the image patch. The proposed paradigm achieves state-of-the-art results in solving the document image binarization challenges. Our research experiments illustrate the same.

In the proposed network, the degraded input image is first split into patches and is attached to a latent token representation. Instead of the naive tokenization used in vanilla ViT, we employed the progressive tokenization method described in \citep{yuan2021tokens}, also known as the Tokens-to-Token module, to combine nearby tokens into a single token. This method may gradually shorten tokens by modeling the local structural information of the tokens that surround them. The decoder creates enhanced image patches after the encoding process, which are then reassembled to make the final enhanced output image. Unlike the previously proposed models for document image binarization using a vision transformer, which uses a simple tokenization method, resulting in the extraction of global contextual features, while not being able to retrieve the intricate details that reside in the local information of the patches. The architecture of the proposed methodology ensures that the patch's local structure is captured as well during the encoding stage. 

The proposed network has been shown to perform better than the Vanilla ViT for the document binarization task because the T2T block can better model the local structure information in document images, which is very crucial for this task. The T2T block does this by recursively aggregating neighboring tokens into one token, such that the local structure represented by surrounding tokens can be modeled and the length of tokens can be reduced. This results in making it easier for the encoder block to identify the text in document images, especially those that contain many "white areas" (empty zones without text). Additionally, the proposed method is more efficient than the other Vanilla ViT-based model, requiring fewer parameters and computation resources. This makes it a more practical choice for document image binarization from a computation perspective as well. \\
The contributions of the proposed methodology are summarized below:

\begin{itemize}
    \item We introduce an end-to-end, simple, and effective binarization method using a Tokens-to-token vision transformer. To our knowledge, this is the first Tokens-to-Token-based ViT architecture for document binarization problems. This architecture can also be used for other document image enhancement tasks such as dewarping, deblurring, etc. with some minor modifications.  

    \item The tokens-to-token vision transformer produced state-of-the-art results compared to the simple ViT-based architecture used for document image binarization. In fact, based on our experimental results, the proposed architecture beats the existing benchmark on a majority of the test datasets. In the proposed architecture, instead of simply dividing the input degraded image into patches and feeding them to the transformer, the architecture produces tokens using a progressive tokenization technique.  
    
    \item We conducted extensive experiments and provided a comprehensive and in-depth case study to show the model's successes in solving the document image binarization challenge from both a qualitative and quantitative standpoint on the DIBCO and the H-DIBCO datasets.
    
\end{itemize}
The remaining paper is organized as follows: In Section~\ref{sec:Works} we discuss the related research. The proposed architecture is described in Section~\ref{sec:Method}. Section~\ref{sec:Exp} provides the experimental results and comparisons. Finally, a comprehensive conclusion is outlined in Section~\ref{sec:Con}.   

\section{Related Works}
\label{sec:Works}
This section will cover several binarization approaches for document images. Both the traditional approaches and the most current advancements in this area are covered:

\subsection{Document Image Binarization}
The earliest traditional approaches for document image binarization were based on thresholding, some of them includes~\citep{otsu1979threshold} and~\citep{sauvola2000adaptive}, \citep{niblack1985introduction}. Local adaptive techniques, such as \citep{niblack1985introduction} perform better in situations involving poor-quality photos. Binarization of document images may also be done using learning algorithms like  \citep{papamarkos2001technique}. To enhance the quality of binarization, current deep-learning networks have also been integrated with threshold-based methods. Examples of the same are DeepOtsu~\citep{he2019deepotsu} and SauvolaNet~\citep{li2021sauvolanet}. \citep{pastor2015insights} utilizes CNNs for solving binarization problems. \citep{tensmeyer2017document} proposed a scalable fully convolutional network (FCN) for document image binarization. Expanding on the idea of using CNN-based networks, \citep{bezmaternykh2019u} describes a faster, more precise, and more efficient technique based on U-Net. Additionally, generative models (GAN) are widely applied to solve a variety of image restoration problems. Some of the popular GAN-based techniques include \citep{souibgui2020gan}, \citep{zhao2019document}. 
\begin{figure*}[ht!]
    \centering
    \includegraphics[clip=true, trim = 00 00 00 00, width=1.00\textwidth]{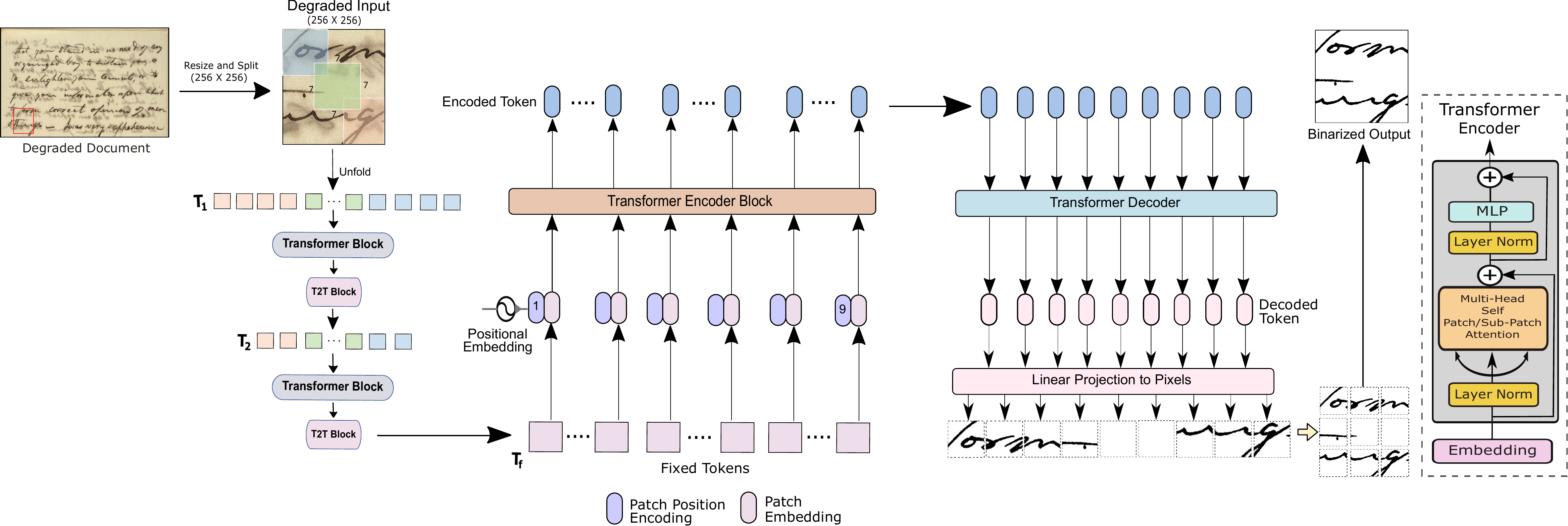} 
    \captionsetup{justification=centering}
    \caption{\centering Architecture of the proposed network for document image binarization. The encoding stage is shown on the left, and the decoding stage is shown on the right. The transformer encoder block is graphically illustrated as well.}
    \label{fig:T2TBinarization}
\end{figure*}
\subsection{Transformers for Document Image Binarization}
Recent advancements in deep learning applications have been driven by transformers. This drives numerous efforts to use them for vision tasks such as classification~\citep{biswas2020identification}, medical image segmentation~\citep{biswas2023polypsam}, object detection~\citep{zhao2019object}, drug discovery~\citep{9181309} and so forth. Transformers have also been utilized for tasks like image restoration \citep{liang2021swinir} and image de-warping \citep{li2021selfdoc}. \citep{souibgui2022docentr} proposed a fully transformer-based approach for document image enhancement, without the need for any CNN. However since their approach is entirely based on conventional ViT without any design change, it fails to capture the local information from the patches. Extending to that, we propose \textbf{T2T-BinFormer}, a more effective Tokens-to-token ViT for more effective document image binarization. 

\section{Proposed Methodology}
\label{sec:Method}
The proposed methodology is an efficient document image binarization framework that employs a Tokens-to-token vision transformer (T2T-ViT) based encoder-decoder architecture. The entire architecture is divided into two stages - \textit{Encoding} and \textit{Decoding} respectively. Fig.~\ref{fig:T2TBinarization} illustrates the architecture of the proposed binarization framework. The components of the architecture are discussed in detail from Subsection~\ref{sub_sec:Patch/Sub-Patch} to \ref{sub_sec:Loss/Optimization}:

\subsection{Patch Tokenization and Position Embedding}
\label{sub_sec:Patch/Sub-Patch}
\begin{figure}[ht]
    \centering
    \includegraphics[clip=true, width=1.0\columnwidth]{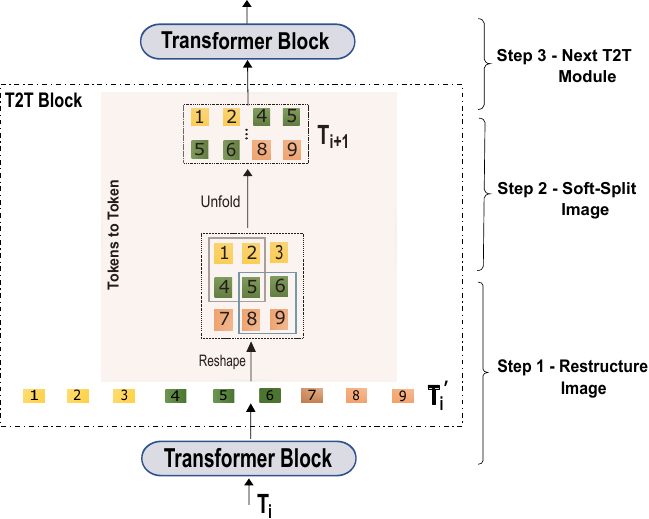}
    \caption{\centering Illustration of T2T process as proposed by Yuan et al.~\citep{yuan2021tokens}}
    \label{fig:T2TBlock}
\end{figure}

The input image, $\mathbf{X} \in \rm \mathbb{R}^{(H\times{W}\times{C})}$, where $\rm H$, $\rm W$ is the height and width, respectively and $C$ be the number of channels, here $C$ = 3, converted into tokens using T2T (Token-to-Token) Module. We apply the same operations as mentioned in \citep{yuan2021tokens} for converting the input degraded image into tokens or embeddings. Thus the input image can be written as $\mathbf{X}\in \mathbb{R}^{(\rm H\times{W}\times{3})}$. The input image $\mathbf{X}$ is downsampled by unfolding, leading to overlapping image data in each token as shown in Fig.~\ref{fig:T2TBlock}. After applying several T2T blocks the input image is converted into tokens of fixed length, which is the output of the Tokens-to-Token module, say $\mathbf{X'}\in \mathbb{R}^{(\rm H\times{W}\times{3})}$. These tokens include both the global and local structural information of the image. Positional embeddings are linearly appended to the tokens to guarantee that the pictures' positional information is preserved. The parameters of the learnable positional embedding vectors are randomly initialized. The equation below illustrates the positional embeddings for the patch representations:  
\begin{equation}
\label{ref:patch_pos_embedding}
\mathbf{X'_{token}} = {\rm X'} \oplus \rm{PE}_{Patch}
\end{equation}
where $\rm {PE}_{Patch} \in \mathbb{R}^{(n_{patch} \times \rm D)}$ and $\oplus$ represent the patch positional encodings and element-wise addition, respectively. $\rm D$ represents the dimension of each patch embedding. 
\subsection{Transformer Encoder Block}
\label{sub_sec:Encoder}
$\mathbf{X'_{token}}$ shown in Eq.~(\ref{ref:patch_pos_embedding}) is the input to the transformer encoder block. There might exist one or more transformer encoder blocks. The transformer encoder follows the identical structure as \citep{dosovitskiy2020image}. Each transformer encoder comprises a multi-head self-attention \citep{vaswani2017attention} to capture the feature dependencies along with the multi-layered perceptron
(MLP) layer. The overall tokenized matrix, $\mathbf{X'_{token}}$ passes through a layer normalization layer (LN) and then enters the multi-head self-attention (MSA) block of the transformer encoder. In the proposed transformer network, we have 8 self-attention heads working in parallel inside each transformer encoder block, this can be considered to be a hyperparameter. The objective of this step is to capture the contextual knowledge from the image. The below equation mathematically represents the above steps: 
\begin{equation}
    \mathbf{Y_{token}} = \rm MSA(LN(\mathbf{X'_{token}}))
    \label{equ:msa_out}
\end{equation}
\begin{equation}
    \mathbf{Y'_{token}} = \mathbf{Y_{token}} \oplus \rm MLP(LN(\mathbf{Y_{token}}))
    \label{equ:enco_out}
\end{equation}

\subsection{Binarization using Transformer Decoder Block}
\label{sub_sec:Decoder}
At the end of Section~\ref{sub_sec:Encoder}, we have the encoded token representation ($\mathbf{Y'_{token}}$) of the input image. This encoded vector contains both global and local structural information about the image. This section deals with the decoding stage of our architecture, which is also the final step of the proposed binarization model. The encoded token is the input to the series of transformer blocks used in the decoder stage as illustrated in Fig.~\ref{fig:T2TBinarization}. The decoder processes the encoded tokens in the same way the global transformer does in the encoding stage. These tokens are projected with a linear layer to the required pixel values after being transmitted in the transformer decoder blocks. As a result, each output element is converted to a vector $\mathbf{{Y}}_{\rm pred\_pixel}$ that represents a patch in the final image. As the last step, the binarized patches are rearranged to construct the original-sized output of the corresponding degraded image.

\subsection{Loss and Optimization Function}
\label{sub_sec:Loss/Optimization}
For the training purpose, we use \emph{Mean Square Error} (MSE) loss function, since MSE can handle the outliers more optimally. The ground truth (GT) image is divided into patches and flattened into a vector. Then we calculate the MSE loss between the output vector obtained from the decoder and the GT vector. Let the GT vector be $\mathbf{{GT}}_{\rm patches}$. The below equation illustrates the above calculation mathematically: 
\begin{equation}
\rm{MSE} = {\frac{1} {n}}\sum_{i=1}^{n}(\mathbf{{GT}}_{patches}-\mathbf{{Y}}_{\rm pred\_pixel})^2
\end{equation}
For optimization, we are using {Weight decay Adam Optimizer (AdamW)}\citep{kingma2014adam} optimization function. This may lessen the risk of the model overfitting.
\section{Experimental Setup}
\label{sec:Exp}
\subsection{Dataset and Evaluation Metrics}
We used a variety of DIBCO and H-DIBCO datasets for training, testing, and validating our proposed model. We used the following DIBCO datasets: 2009~\citep{5277767}, 2010\citep{5693650}, 2011\citep{6065249}, 2012\citep{6424498}, 2013\citep{6628857}, 2014\citep{6981120}, 2018\citep{8583809}, and 2019\citep{8978205}. These datasets contain both handwritten and machine-typed document images. We used the "leave-one-out strategy" for our experiment. This means that we use all of the documents from a single (H-)DIBCO year as the training dataset and all of the remaining documents from other years as the testing dataset. In the following subsections, we will assess the model's performance both quantitatively and qualitatively. To perform a qualitative assessment of how well the binarization model is working, we used the standard metrics such as the peak signal-to-noise ratio (\textbf{PSNR}), F-Measure (\textbf{FM}), pseudo-F-measure $(\rm \mathbf{F_{ps}})$, and distance reciprocal distortion (\textbf{DRD}).

\subsection{Implementation Details}
Each degraded image and its corresponding ground truth images are transformed into ($256 \times 256$) images during data preprocessing. These preprocessed images are then utilized for training, testing, and validation. This resizing is a matter of choice, which means, the original DIBCO image can be resized to any other dimension, such as ($128 \times $128) or ($512 \times 512$). We also used data augmentation to increase the data volume. This includes - flipping, random cropping, and horizontal and vertical rotation. The internal structure of \textbf{T2T-BinFormer} is described in Table~\ref{Tab:Table1}. Patch size $\rm P$ is chosen as 16. The model's learning rate is 1.5e-4, eps is 1e-08, and weight decay is 0.05. We used a 16-image batch size and trained each model iteration for 200 epochs on a single NVIDIA RTX 4070Ti GPU. All the training and inference algorithm is carried out in PyTorch.
\begin{table}[htbp]
    \centering
    \caption{Hyper-Parameters Settings for the Proposed Model.}
    \resizebox{\columnwidth}{!}{
        \begin{tabular}{|l|c|c|c|c|c|}
            \toprule
            \textbf{Block} & \textbf{Patch Size} & \textbf{\#Layers} & \textbf{\#Heads} & \textbf{Dim} & \textbf{MLP Dim} \\
            \midrule
            \textbf{Encoder} & $16 \times 16$ & 12 & 8 & 768 & 2048 \\
            \textbf{Decoder} & $16 \times 16$ & 1 & 8 & 64 & 2048 \\
            \hline
        \end{tabular}
    }
    \label{Tab:Table1}
\end{table}
\section{Experimental Results and Comparison}


\subsection{Quantitative Results and Comparison}
\label{sub_sec:QuantAnalysis}
After selecting the best hyperparameters and training the model, we run experiments on a variety of datasets from both the DIBCO and H-DIBCO and evaluate the results with those obtained using other state-of-the-art techniques. We begin our quantitative comparative analysis with the DIBCO 2009 dataset which contains 10 sample images of both handwritten and machine-typed types. The outcome of the proposed method and the other state-of-the-art techniques are shown in Table~\ref{Tab:Table2009}. The performance of the proposed model is mentioned in the last row, in bold. The findings demonstrate that, for this dataset, the proposed model significantly outperformed all other techniques in terms of PSNR, FM, $\rm F_{ps}$, and DRD. It also shows that the proposed method performed better than the existing vanilla ViT-based model. Following that, we tested our model on the DIBCO 2010, 2011, 2012, 2013, 2014, 2018, and the most recent 2019 datasets, which include 10, 16, 14, 16, 10, 10, and 20 images, respectively. The results of these can be found in Tables~\ref{Tab:Table2010}, \ref{Tab:Table2011}, \ref{Tab:Table2012}, \ref{Tab:Table2013}, \ref{Tab:Table2014}, \ref{Tab:Table2018} and \ref{Tab:Table2019}, respectively. The output tables exhibit the obtained results and draw a comparative analysis with the previously proposed models. We can infer that the \textbf{T2T-BinFormer} model gives the comparative performance in terms of PSNR, FM, $\rm F_{ps}$, and DRD in the majority of the cases thus making it a viable solution for document binarization task. 
\begin{table}[htbp]
\caption{\centering Comparative results of the proposed method and other state-of-the-art methods on the DIBCO 2009 dataset.}
\centering
\resizebox{\columnwidth}{!}{
\begin{tabular}{l|l|c|c|c|c}
\toprule
\textbf{Method}   &\textbf{Model}             & \textbf{PSNR$\uparrow$}        &\textbf{FM$\uparrow$}        &\textbf{Fps$\uparrow$}   &\textbf{DRD$\downarrow$}   \\ \midrule
Bradley\citep{bradley2007adaptive}     & Thresh. & 15.21      & 80.70      & 82.10 & 0.53              \\
Otsu\citep{otsu1979threshold}     & Thresh.          & 15.31      & 78.60      &80.50  &0.55    \\
Sauvola et al.\citep{sauvola2000adaptive}     & Thresh. & 16.32      & 85.00      &89.50  &0.50              \\
Gatos et al.\citep{gatos2006adaptive}     & Thresh.          & 17.02      & 87.27      & 89.28  & 0.50    \\
Winner\citep{5277767}     &CNN          & 18.66      & 91.24 &- & -   \\
Zhao et al.\citep{zhao2019document}     & GAN & 20.30      & 94.10      &95.26 & 1.82            \\
Kang et al.\citep{kang2021complex}     & U-net CNN          & 20.91      & \textbf{96.67}     & - & -      \\
SauvolaNet.\citep{li2021sauvolanet}     & GAN  &21.53   & 95.51      & 96.58 & 0.37               \\
DocEnTr\citep{souibgui2022docentr}     & Transformer      & 21.32   & 95.50  & 96.70  &\textbf{0.18}             \\
\toprule 
\textbf{Proposed}    &\textbf{Transformer} &\textbf{22.04}  &96.20  & \textbf{97.62}  &0.22
\\ \hline 
\end{tabular}}
\label{Tab:Table2009}
\end{table}

\begin{table}[htbp]
\caption{\centering Comparative results of the proposed method and other state-of-the-art methods on the DIBCO 2010 dataset.}
\centering
\resizebox{\columnwidth}{!}{
\newcolumntype{M}[1]{>{\centering\arraybackslash}m{#1}}
\begin{tabular}{l|l|c|c|c|c}
\toprule
\textbf{Method}   &\textbf{Model}             & \textbf{PSNR$\uparrow$}        &\textbf{FM$\uparrow$}        &\textbf{Fps$\uparrow$}   &\textbf{DRD$\downarrow$}   \\ \midrule
DE-GAN\citep{souibgui2020gan}   & GAN          & 15.80      & 83.00      & 82.96 & 0.18     \\
Sauvola et al.\citep{sauvola2000adaptive}     & Thresh. & 15.86      & 74.00      &83.39  &0.11              \\
Bradley\citep{bradley2007adaptive}     & Thresh. & 16.78      & 82.90      & 88.39 & 0.13              \\
Gatos et al.\citep{gatos2006adaptive}     & Thresh.          & 16.18      & 77.90      & 85.36  & \textbf{0.08 }   \\
Otsu\citep{otsu1979threshold}     & Thresh.          & 17.53      & 85.50      &90.69  &0.15    \\
Winner\citep{su2010binarization}     & Thresh. & 19.78      & 91.50      &93.58  &-              \\
Xiong et al.\citep{xiong2021enhanced}    & Estimation          & 20.97      & 93.73      & 95.18  & -    \\
DocEnTr\citep{souibgui2022docentr}     & Transformer      & 21.90   & 95.10  & 96.85  &0.13            \\
\toprule 
\textbf{Proposed}    &\textbf{Transformer} &\textbf{23.00}  &\textbf{96.17} &\textbf{97.67}  &0.22
\\ \hline 
\end{tabular}}
\label{Tab:Table2010}
\end{table}

\begin{table}[htbp]
\caption{\centering Comparative results of the proposed method and other state-of-the-art methods on the DIBCO 2011 dataset.}
\centering
\resizebox{\columnwidth}{!}{
\newcolumntype{M}[1]{>{\centering\arraybackslash}m{#1}}
\begin{tabular}{l|l|c|c|c|c}
\toprule
\textbf{Methods} & \textbf{Models}  &\textbf{PSNR$\uparrow$}   & \textbf{FM$\uparrow$}       &\textbf{Fps$\uparrow$}   & \textbf{DRD$\downarrow$} \\ \midrule
Sauvola et al.\citep{sauvola2000adaptive}     & Thresh. & 15.60      & 82.10      & 87.70 & 8.50              \\
Otsu\citep{otsu1979threshold}     & Thresh.          & 15.72      & 82.10      & 84.80  & 9.00    \\
Winner\citep{6065249}     &CNN          & 16.10      & 80.90 &- & 104.40   \\
Zhao et al.\citep{zhao2019document}     & GAN & 19.58      & 92.62      & 95.38 & 2.55               \\
Kang et al.\citep{kang2021complex}     & CNN          & 19.90      & 95.50      &-  & 1.80      \\
Vo et al.\citep{vo2018binarization}  & DSN   &20.10   & 93.30   & 96.40 & 2.00               \\
Sungho et al.\citep{suh2020two}     & GAN          & 20.22      & 93.57  &95.93  &1.99      \\
Bhunia et al.\citep{bhunia2019improving} &Adver. Learning   &20.40   & 93.70   & 96.80 & 1.80               \\
SauvolaNet.\citep{li2021sauvolanet}     & GAN  &20.55   & 94.32      & 96.40 & 1.97               \\
Dang et al.\citep{dang2021document}     & CNN          & 22.09      & 95.61      & 97.34 & 1.48      \\
DocEnTr\citep{souibgui2022docentr}     & Transformer      & 20.81   & 94.37  & 96.15  &1.63             \\
Text-DIA\citep{souibgui2022text}     & Transformer      & 21.29   & 95.01  & 96.86  &1.48             \\
\toprule
\textbf{Proposed}    &\textbf{Transformer} &\textbf{22.17}  & \textbf{96.19}    &\textbf{97.63}  &\textbf{0.15}
\\ \hline
\end{tabular}}
\label{Tab:Table2011}
\end{table}
  
\begin{table}[htbp]
\caption{\centering Comparative results of the proposed method and other state-of-the-art methods on the DIBCO 2012 dataset.}
\centering
\resizebox{\columnwidth}{!}{
\newcolumntype{M}[1]{>{\centering\arraybackslash}m{#1}}
\begin{tabular}{l|l|c|c|c|c}
\toprule
\textbf{Methods}   &\textbf{Models}             & \textbf{PSNR$\uparrow$}        &\textbf{FM$\uparrow$}        &\textbf{Fps$\uparrow$}   &\textbf{DRD$\downarrow$}   \\ \midrule
Otsu\citep{otsu1979threshold}     & Thresh.          & 15.03      & 80.18      & 82.65  & 26.46    \\
Sauvola et al.\citep{sauvola2000adaptive}     & Thresh. & 16.71      & 82.89      & 87.95 & 6.59              \\
Winner\citep{6424498}     &CNN          & 20.60      & 92.53 &96.15 & 1.55   \\	
Kang et al.\citep{kang2021complex}     & CNN          & 21.37      & 95.16      &96.44  & 1.13      \\
Xiong et al.\citep{xiong2021enhanced}    & Estimation          & 21.68      & 94.26      &95.16  &2.08    \\
Zhao et al.\citep{zhao2019document}     & GAN & 21.91      & 94.96      & 96.15 & 1.55               \\
Jemni et al.\citep{jemni2022enhance}     & GAN  &22.00   & 95.18      & 94.63   &1.62      \\
DocEnTr\citep{souibgui2022docentr}     & Transformer      & 22.29   & 95.31  & 96.29  &1.60             \\
Text-DIA\citep{souibgui2022text}     & Transformer      & 23.66   & 96.52  & 97.52  &1.10             \\
\toprule
\textbf{Proposed}    &\textbf{Transformer} &\textbf{23.95}  & \textbf{96.80}    &\textbf{98.04}  &\textbf{0.20}
\\ \hline
\end{tabular}}
\label{Tab:Table2012}
\end{table}

\begin{table}[htbp]
\caption{\centering Comparative results of the proposed method and other state-of-the-art methods on the DIBCO 2013 dataset.}
\centering
\resizebox{\columnwidth}{!}{
\newcolumntype{M}[1]{>{\centering\arraybackslash}m{#1}}
\begin{tabular}{l|l|c|c|c|c}
\toprule
\textbf{Methods} & \textbf{Models}  &\textbf{PSNR$\uparrow$}   & \textbf{FM$\uparrow$}       &\textbf{Fps$\uparrow$}   & \textbf{DRD$\downarrow$} \\ \midrule
Otsu\citep{otsu1979threshold}     & Thresh.  & 16.63     & 83.94      & 96.52  & 10.98    \\
Sauvola et al.\citep{sauvola2000adaptive}  & Thresh.  & 16.94      & 85.02      & 89.77  & 7.58          \\
Guo et al.\citep{guo2019nonlinear}   &Non-linear Diffusion   & 17.37    & 82.35      & 85.16      & 8.09 \\
Gatos et al.\citep{gatos2006adaptive}     & Thresh.          & 17.58      & 83.47      & 87.03  & 0.17    \\
DeepOtsu\citep{he2019deepotsu}     & CNN   &20.65   & 93.79      & 91.90 & 2.60               \\
Winner\citep{6628857}     & CNN      & 20.68    & 92.12  & 94.19  & 3.10   \\
Vo et al.\citep{vo2018binarization}  &Hierarchical DSN   &21.40   & 94.40   & 96.00 & 1.80               \\
Kang et al.\citep{kang2021complex}     & U-Net CNN      &22.97      &95.88       &96.38    & 1.46      \\
Dang et al.\citep{dang2021document}     & CNN          & 23.14      & 95.96      & 98.13  & 1.43      \\
DocEnTr\citep{souibgui2022docentr}    & Transformer      & 23.24   & 96.42  & 97.69  &0.14            \\
\toprule
\textbf{Proposed}    &\textbf{Transformer} &\textbf{23.99}  & \textbf{97.10}    & \textbf{98.23}  &\textbf{0.07}
\\ \hline 
\end{tabular}}
\label{Tab:Table2013}
\end{table}

\begin{table}[htbp]
\caption{\centering Comparative results of the proposed method and other state-of-the-art methods on the DIBCO 2014 dataset.}
\centering
\resizebox{\columnwidth}{!}{
\newcolumntype{M}[1]{>{\centering\arraybackslash}m{#1}}
\begin{tabular}{l|l|c|c|c|c}
\toprule
\textbf{Methods} & \textbf{Models}  &\textbf{PSNR$\uparrow$}   & \textbf{FM$\uparrow$}       &\textbf{Fps$\uparrow$}   & \textbf{DRD$\downarrow$} \\ \midrule
Sauvola et al.\citep{sauvola2000adaptive}  & Thresh.  & 17.48      & 83.72      & 87.49  & 5.05          \\
Gatos et al.\citep{gatos2006adaptive}     & Thresh.          & 18.24      & 85.88      & 88.07  & 0.24   \\
Otsu\citep{otsu1979threshold}     & Thresh.  & 18.73     & 91.62      & 95.69  & 2.65    \\
Su et al.\citep{su2012robust}    &Thresh.  & 20.31      & 94.38      & 95.94  & 1.95              \\
Zhao et al.\citep{zhao2019document}     & GAN  & 21.88      & 96.09  & 98.25 & 1.20               \\
DeepOtsu\citep{he2019deepotsu}     & CNN   &22.10   & 95.90      & 97.20 & 0.90               \\
Howe et al.\citep{6065266}    &Thresh.  & 22.24      & 96.49      & 97.30  & 1.08              \\
Vo et al.\citep{vo2018binarization}  &Hierarchical DSN   & 23.23    & 96.66      & 97.59      & 0.79 \\
DocEnTr\citep{souibgui2022docentr}    & Transformer      & 22.99   & 97.16  & 98.28  &\textbf{0.16}            \\
\toprule
\textbf{Proposed}    &\textbf{Transformer} &\textbf{23.48}  & \textbf{97.50}    & \textbf{98.50}  &0.21
\\ \hline
\end{tabular}}
\label{Tab:Table2014}
\end{table}

\begin{table}[htbp]
\caption{\centering Comparative results of the proposed method and other state-of-the-art methods on DIBCO 2018 dataset.}
\centering
\resizebox{\columnwidth}{!}{
\newcolumntype{M}[1]{>{\centering\arraybackslash}m{#1}}
\begin{tabular}{l|l|c|c|c|c}
\toprule
\textbf{Methods} & \textbf{Models}  &\textbf{PSNR$\uparrow$}   & \textbf{FM$\uparrow$}       &\textbf{Fps$\uparrow$}   & \textbf{DRD$\downarrow$} \\ \midrule
Otsu\citep{otsu1979threshold}     & Thresh.          & 9.74      & 51.45      & 53.05  & 59.07    \\
Sauvola et al.\citep{sauvola2000adaptive}     & Thresh. & 13.78      & 67.81      & 74.08 & 17.69              \\
DE-GAN\citep{souibgui2020gan}   & GAN          & 16.16      & 77.59      & 85.74 & 7.93     \\
Kang et al.\citep{kang2021complex}     &CNN          & 19.39      & 89.71      & 91.62 & 2.51      \\
Zhao et al.\citep{zhao2019document}     & GAN & 18.37      & 87.73      & 90.60 & 4.58               \\
Winner\citep{8270159}     & CNN          & 19.11      & 88.34 &90.24 & 4.92   \\
DocEnTr\citep{souibgui2022docentr}    & Transformer      & 19.47   & 92.53  & 95.15  &2.37             \\
Text-DIA\citep{souibgui2022text}     & Transformer      & 19.95   & 91.32  & 94.44  &3.21             \\
Jemni et al.\citep{jemni2022enhance}     & GAN          & 20.18      & 92.41      & 94.35 &2.60      \\
\toprule
\textbf{Proposed}    &\textbf{Transformer} &\textbf{22.33}  & \textbf{95.60}    &\textbf{96.97}  &\textbf{0.13}
\\ \hline
\end{tabular}}
\label{Tab:Table2018}
\end{table}
\begin{table}[htbp]
\caption{\centering Comparative results of the proposed method and other state-of-the-art methods on DIBCO 2019 dataset.}
\centering
\resizebox{\columnwidth}{!}{
\newcolumntype{M}[1]{>{\centering\arraybackslash}m{#1}}
\begin{tabular}{l|l|c|c|c|c}
\toprule
\textbf{Methods} & \textbf{Models}  &\textbf{PSNR$\uparrow$}   & \textbf{FM$\uparrow$}       &\textbf{Fps$\uparrow$}   & \textbf{DRD$\downarrow$} \\ \midrule
Niblack\citep{niblack1985introduction}     & Thresh.          & 8.62      & 43.00      & 43.00  & \textbf{0.25}    \\
Otsu\citep{otsu1979threshold}     & Thresh.          & 9.05      & 48.00      & 48.00  & 0.29    \\
Bradley\citep{bradley2007adaptive}     & Thresh.          & 10.98      & 58.00      & 58.00  & 0.30    \\
Sauvola et al.\citep{sauvola2000adaptive}     & Thresh. & 13.11      & 65.00      & 66.00 & 0.31              \\
Gatos et al.\citep{gatos2006adaptive}     & Thresh.          & 12.81      & 65.00      & 65.00  & 0.31    \\
DocEnTr\citep{souibgui2022docentr}    & Transformer      & 13.85   & 59.00  & 60.00  &0.30             \\
\toprule
\textbf{Proposed}    &\textbf{Transformer} &\textbf{14.49}  & \textbf{65.70}    &\textbf{67.82}  &0.29
\\ \hline
\end{tabular}}
\label{Tab:Table2019}
\end{table}
These improvements can be quantified as - (1) For DIBCO 2009, we notice that the performance of the model outperforms the previous best model-based method by 0.72 and 3.32\% absolute improvement and percentage improvement respectively on PSNR. (2) We also calculated the percentage improvement from the last best method in terms of PSNR for other DIBCO sets. The results of which are - 5.02\% for 2010, 1.23\% for 2011, 1.22\% for 2012, 2.34\% for 2013, 0.64\% for 2014, 10\% for 2018 and 4.62\% for 2019. The quantitative results produced by the model evidently revealed that the proposed network makes good use of the Tokens-to-Token module instead of the simple tokenization used in current transformer-based approaches. This also helps in understanding that the local structure captured by the proposed architecture plays a crucial role in the model's performance. 

\subsection{Qualitative Results and Comparison}
\begin{figure}[htbp]
\begin{center}
\captionsetup{justification=centering}
\resizebox{\columnwidth}{!}{
\begin{tabular}{cccc}
\includegraphics[width=1\columnwidth]{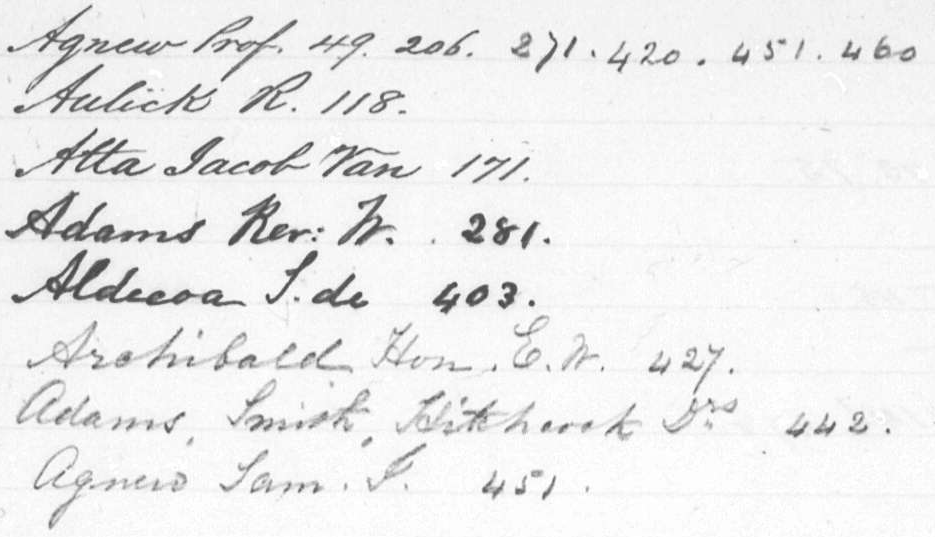} &
\includegraphics[width=1\columnwidth]{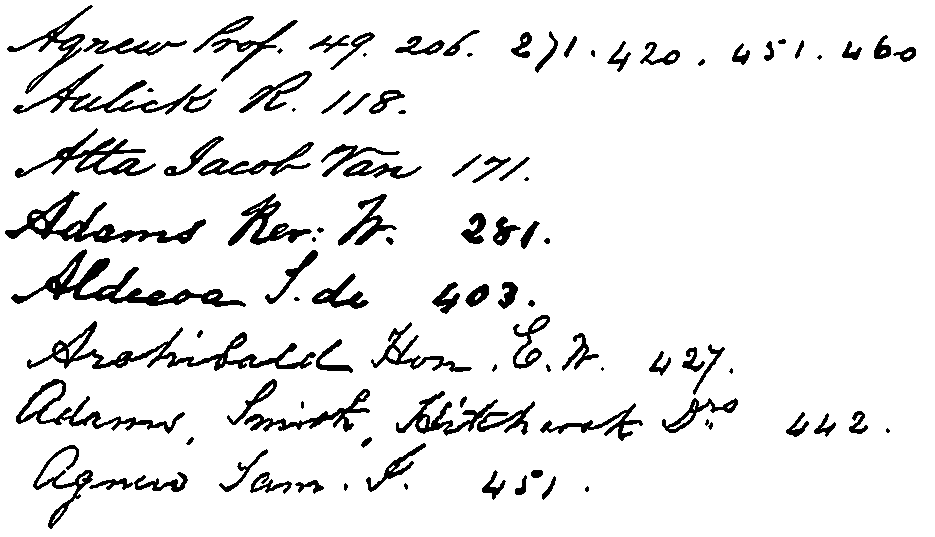} &
\includegraphics[width=1\columnwidth]{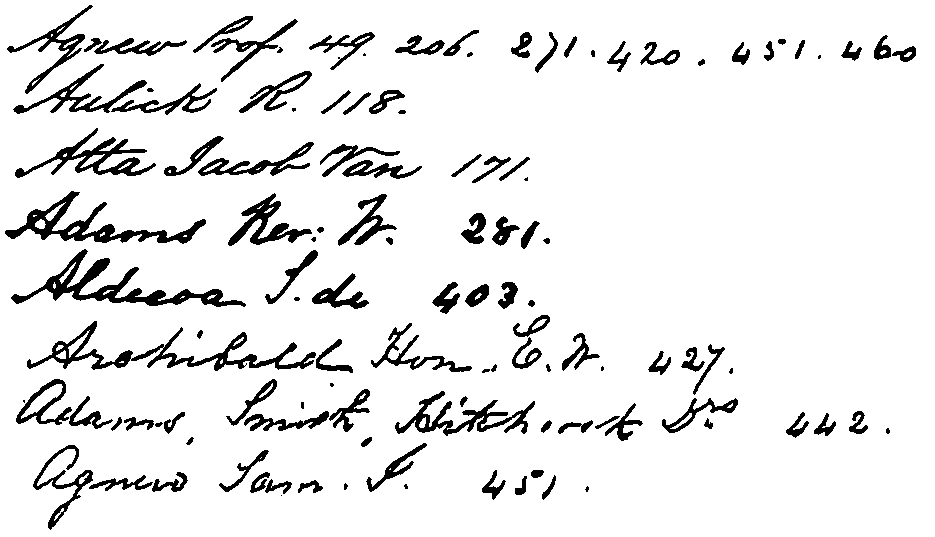} \\ \\
\includegraphics[width=1\columnwidth]{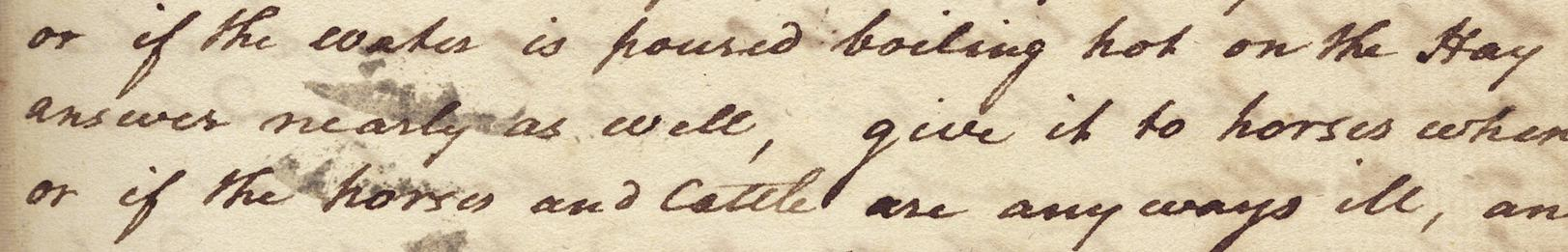} &
\includegraphics[width=1\columnwidth]{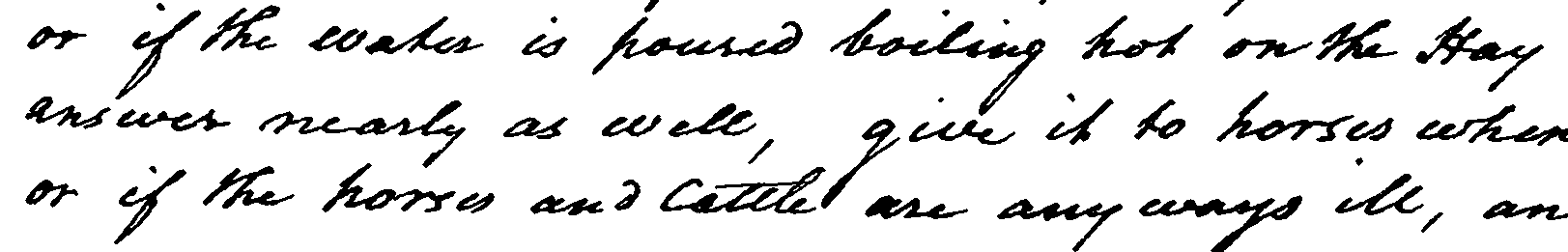} &
\includegraphics[width=1\columnwidth]{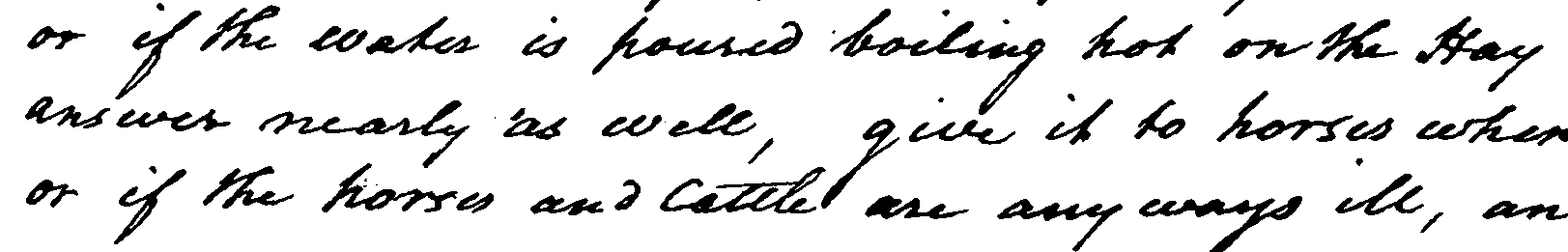} \\ \\
\includegraphics[width=1\columnwidth]{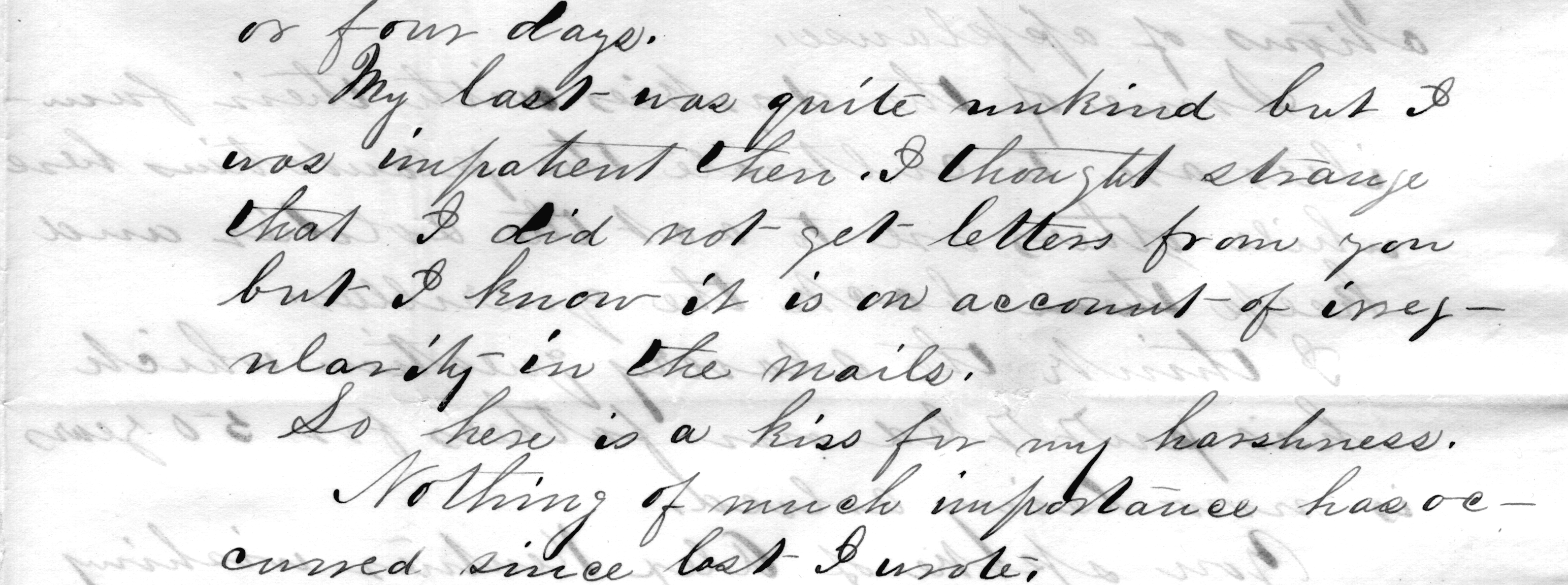} &
\includegraphics[width=1\columnwidth]{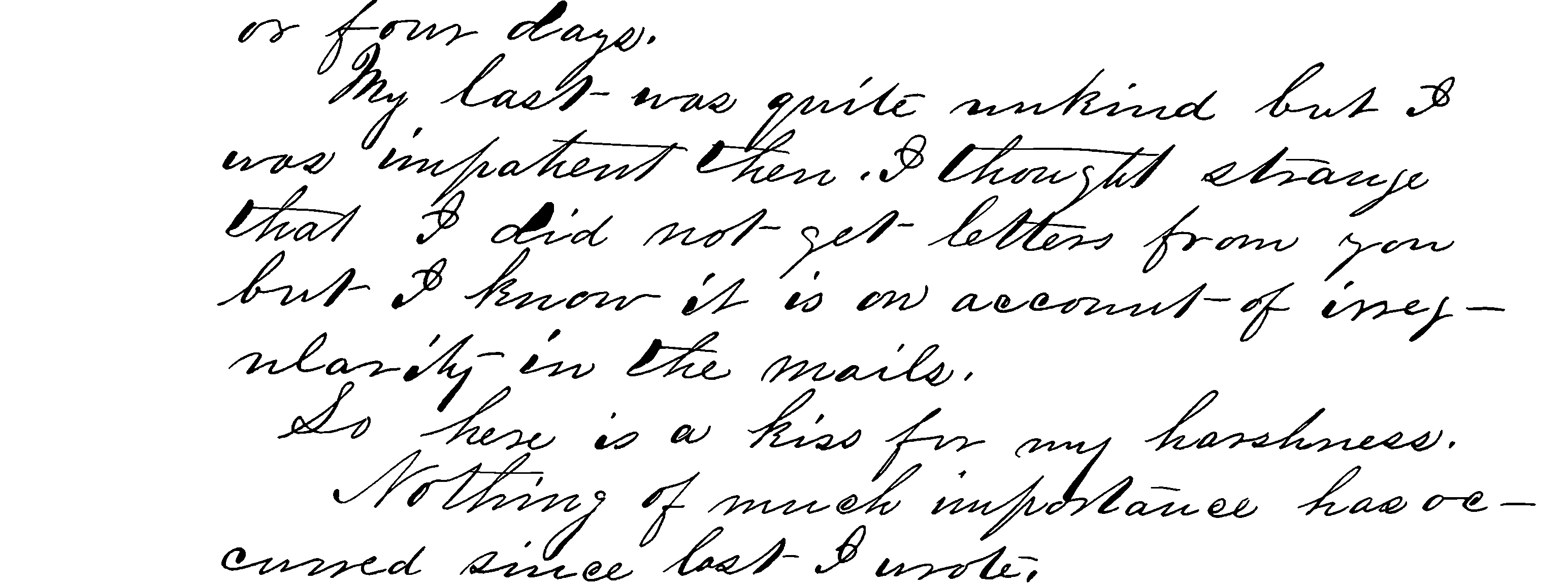} &
\includegraphics[width=1\columnwidth]{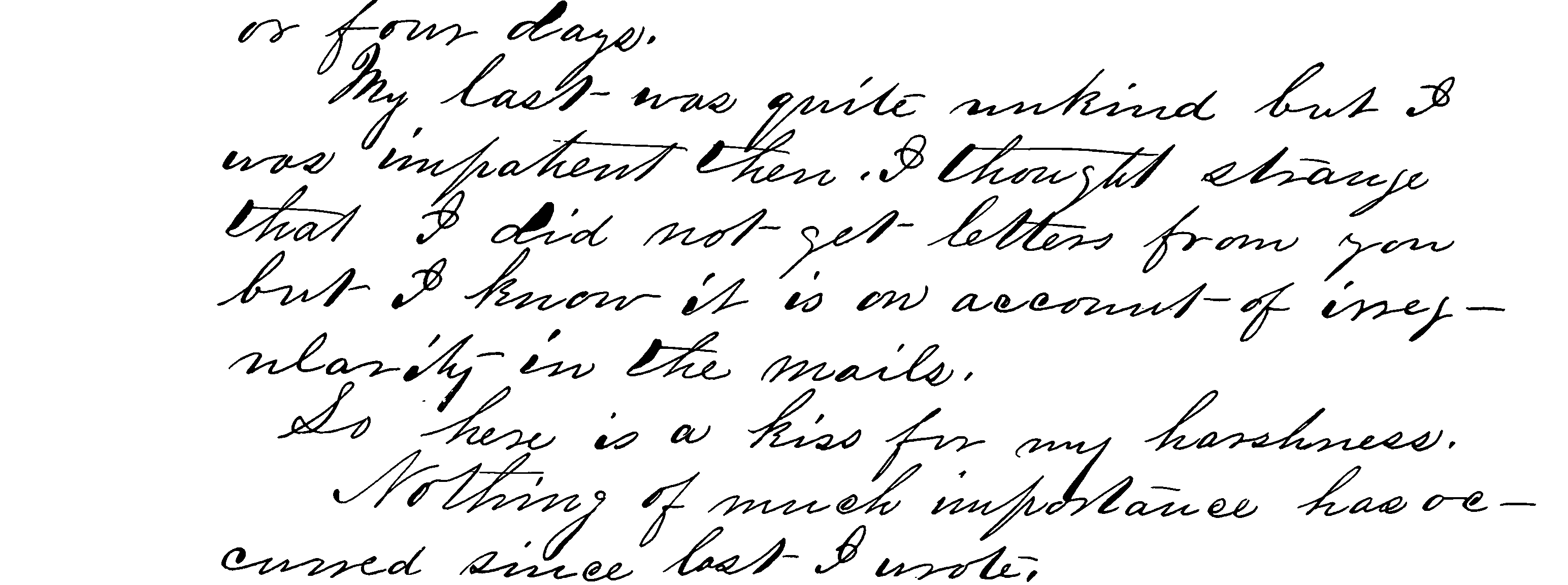} \\ \\
\includegraphics[width=1\columnwidth]{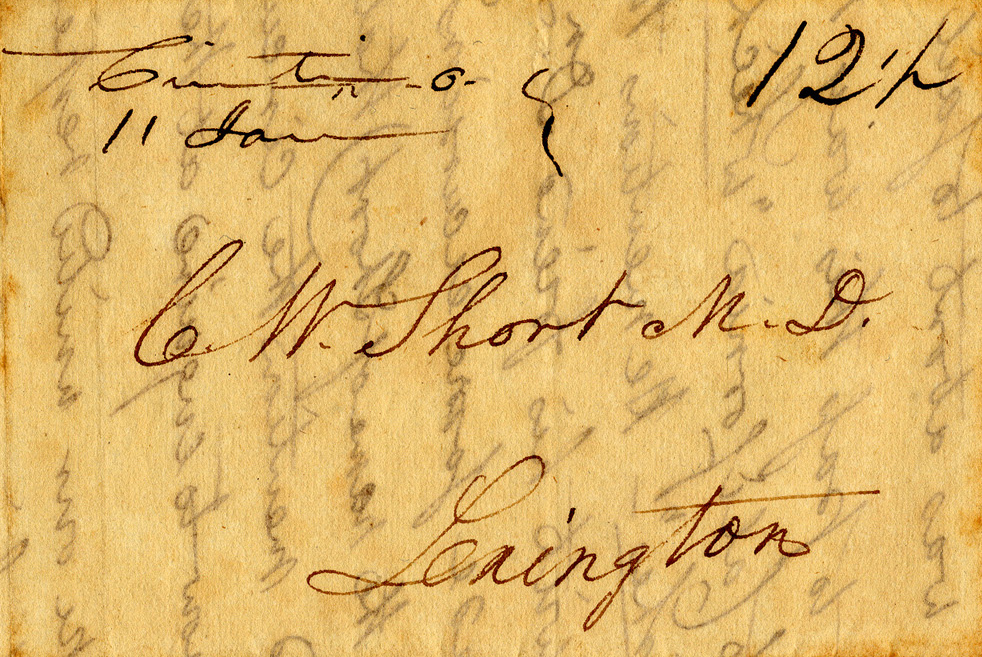} &
\includegraphics[width=1\columnwidth]{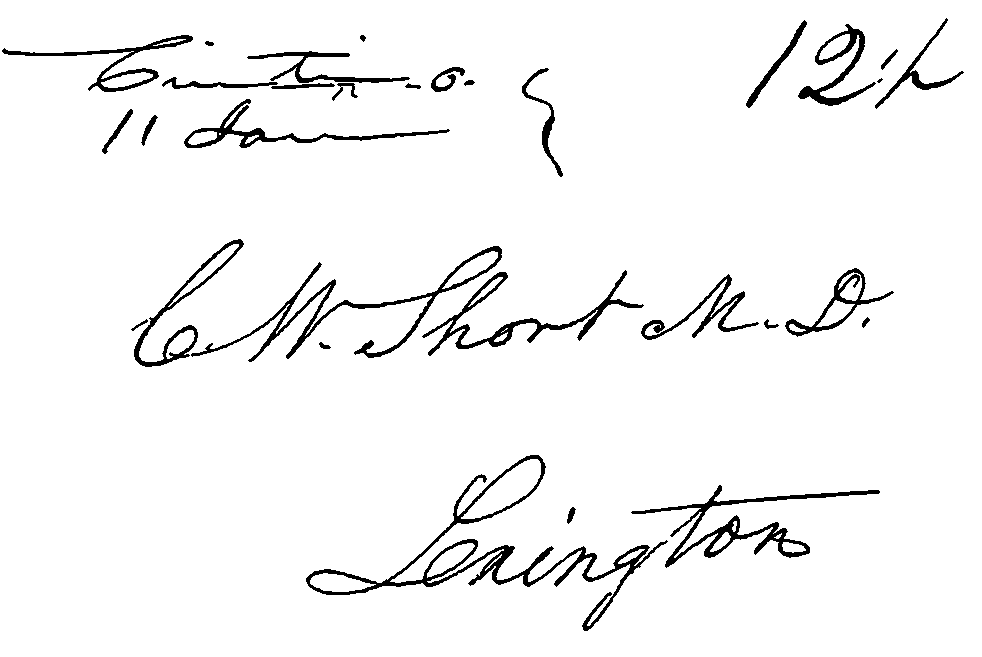} &
\includegraphics[width=1\columnwidth]{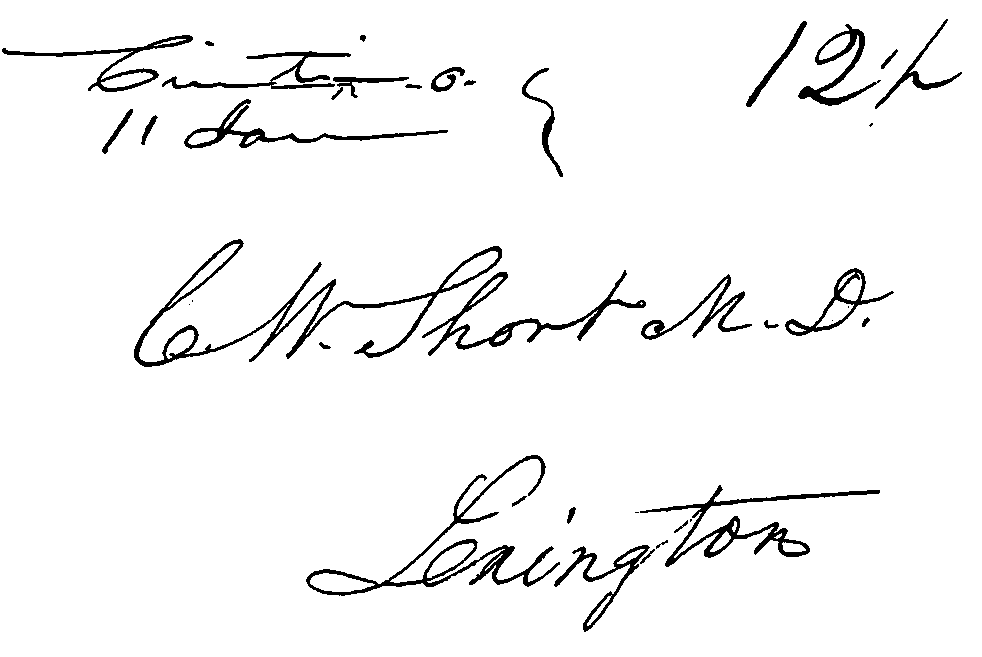} \\ \\
\includegraphics[width=1\columnwidth]{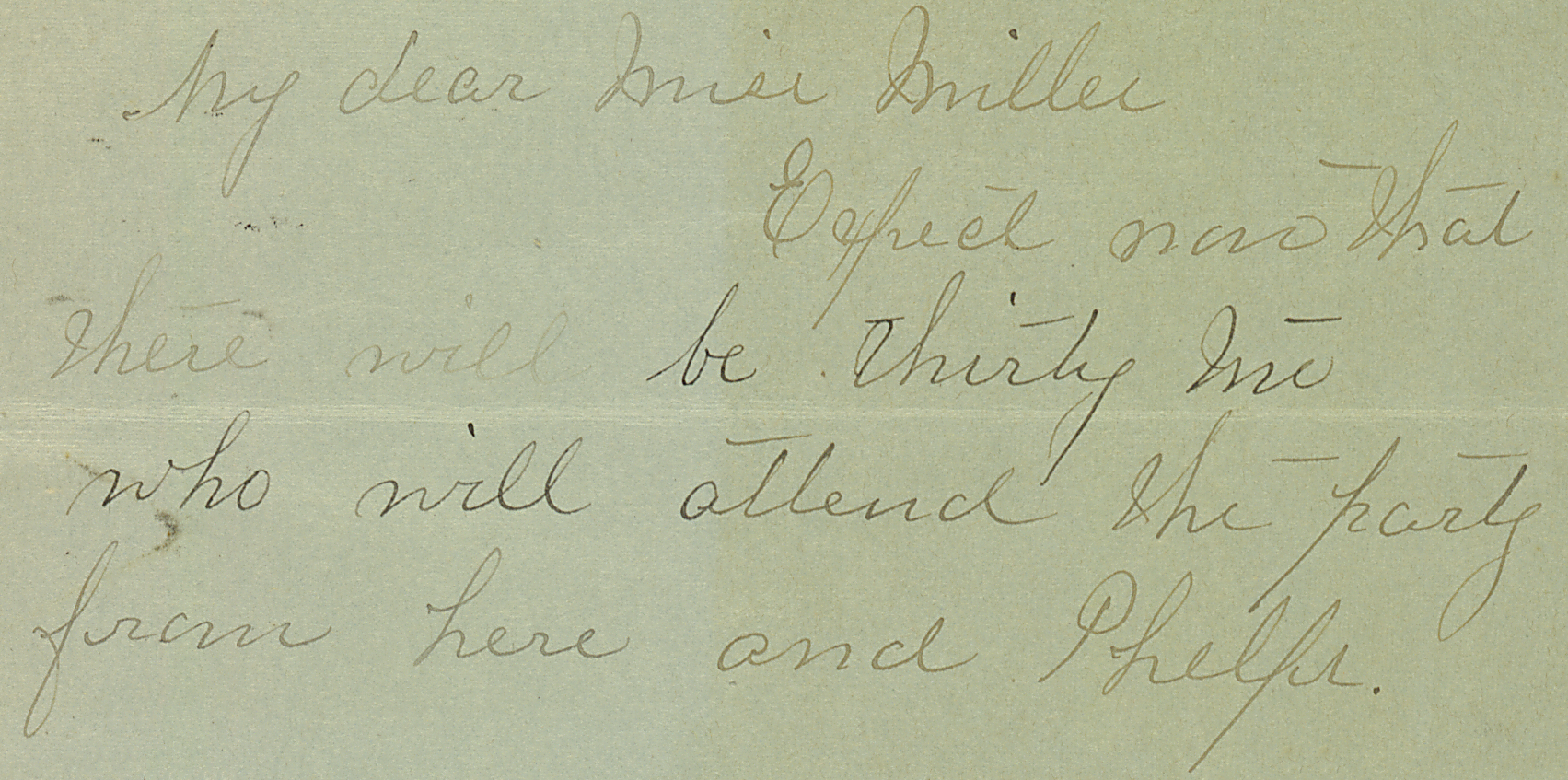} &
\includegraphics[width=1\columnwidth]{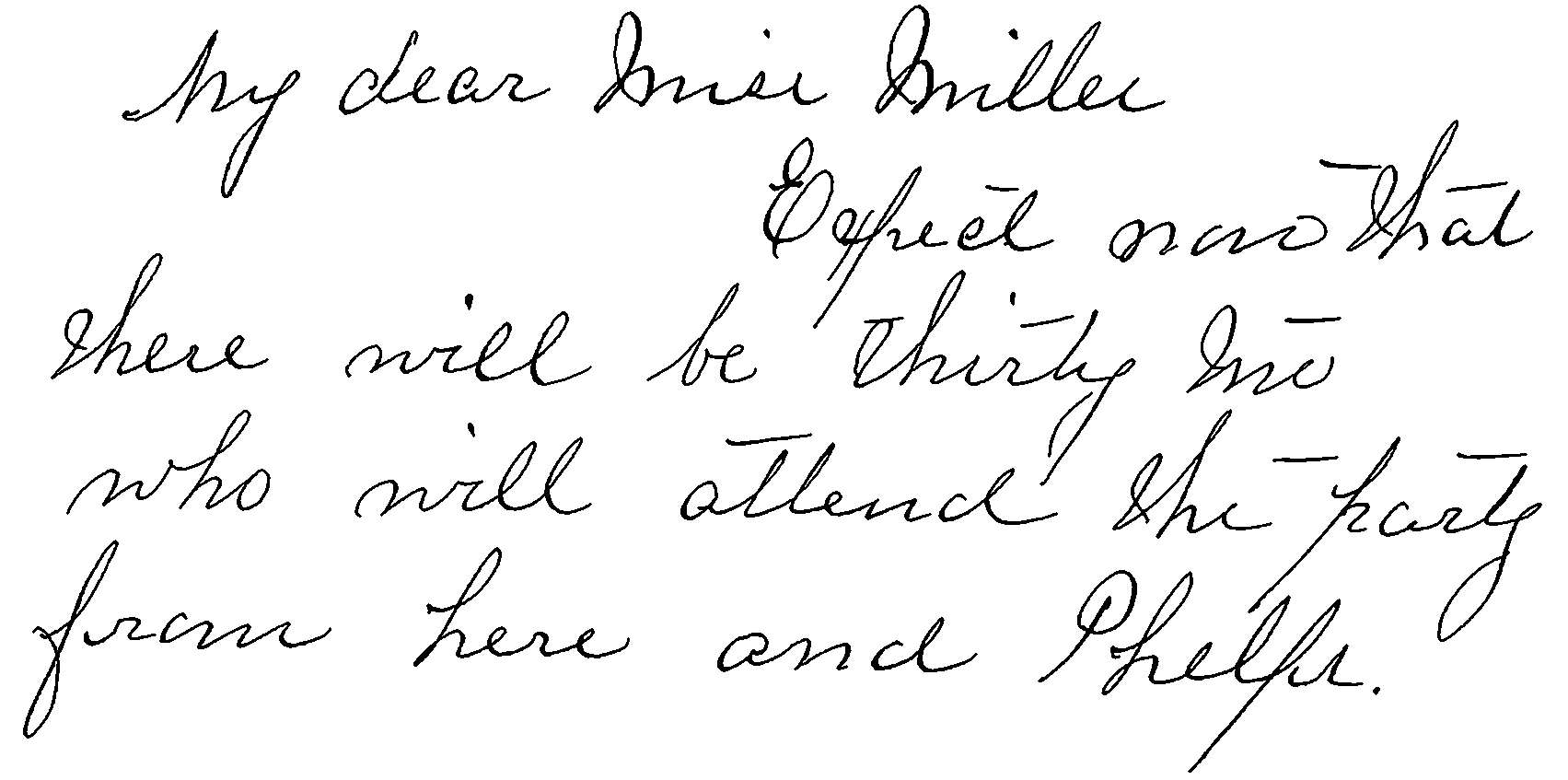} &
\includegraphics[width=1\columnwidth]{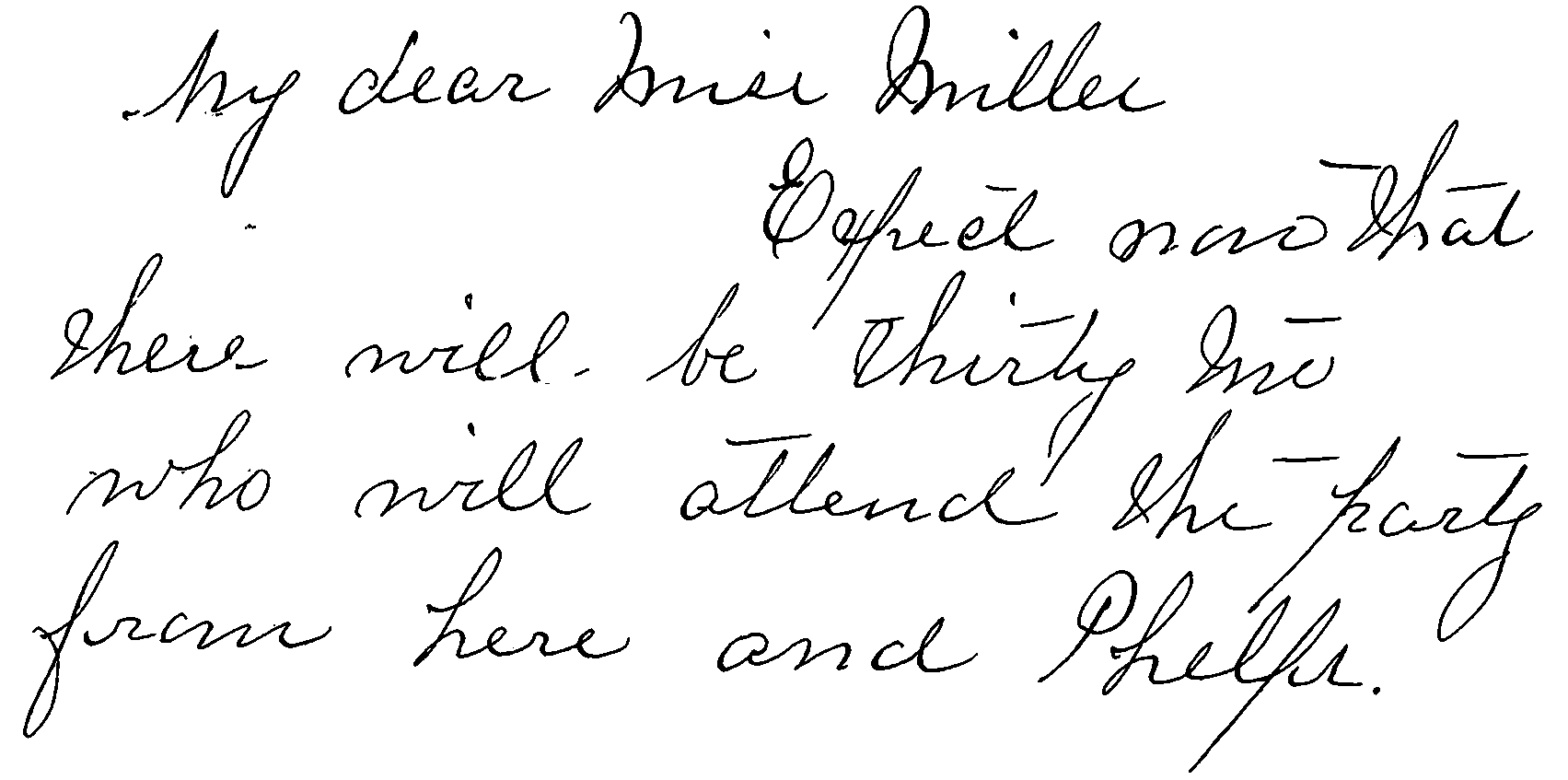} \\ \\
\includegraphics[width=1\columnwidth]{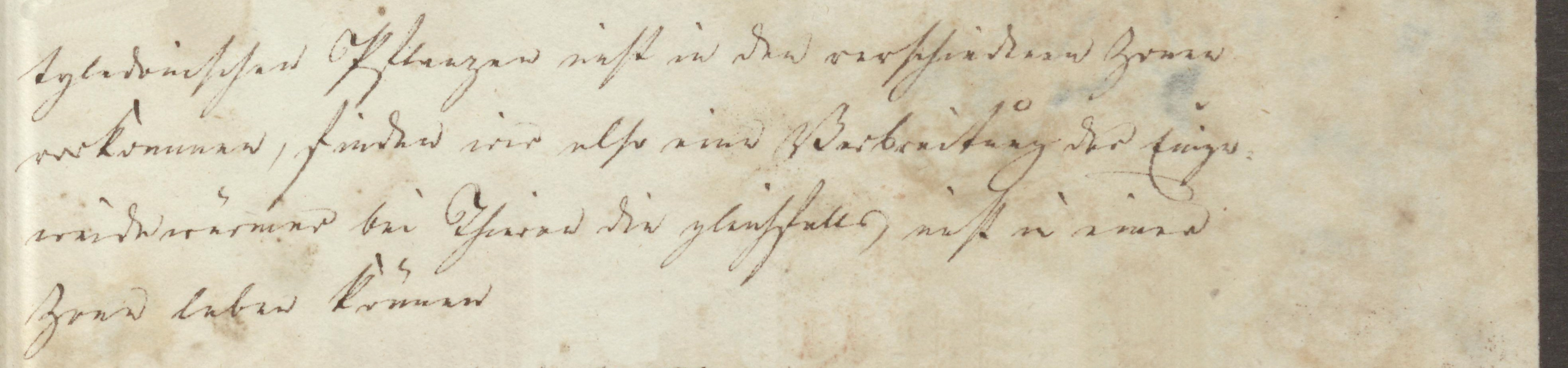} &
\includegraphics[width=1\columnwidth]{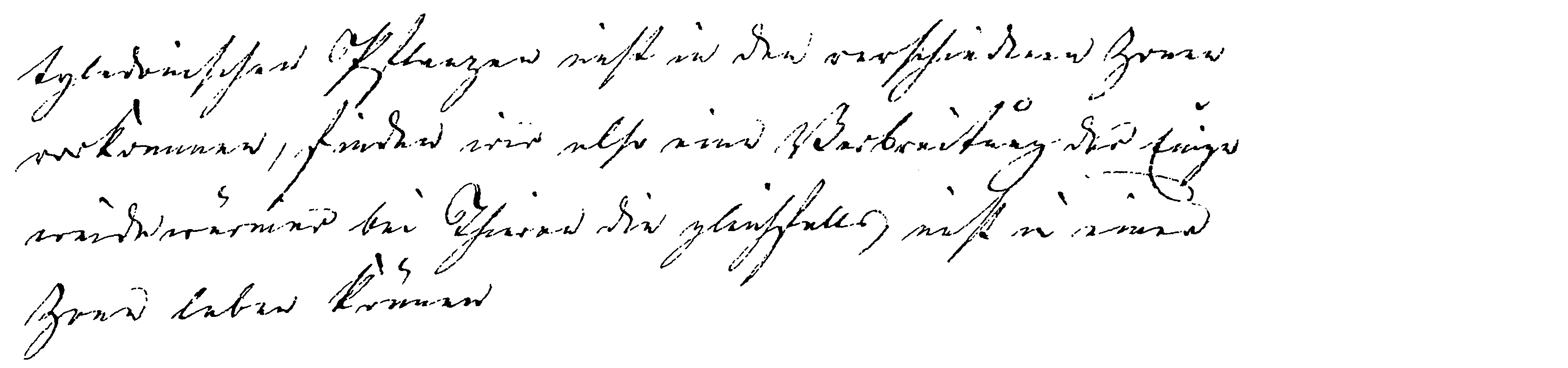} &
\includegraphics[width=1\columnwidth]{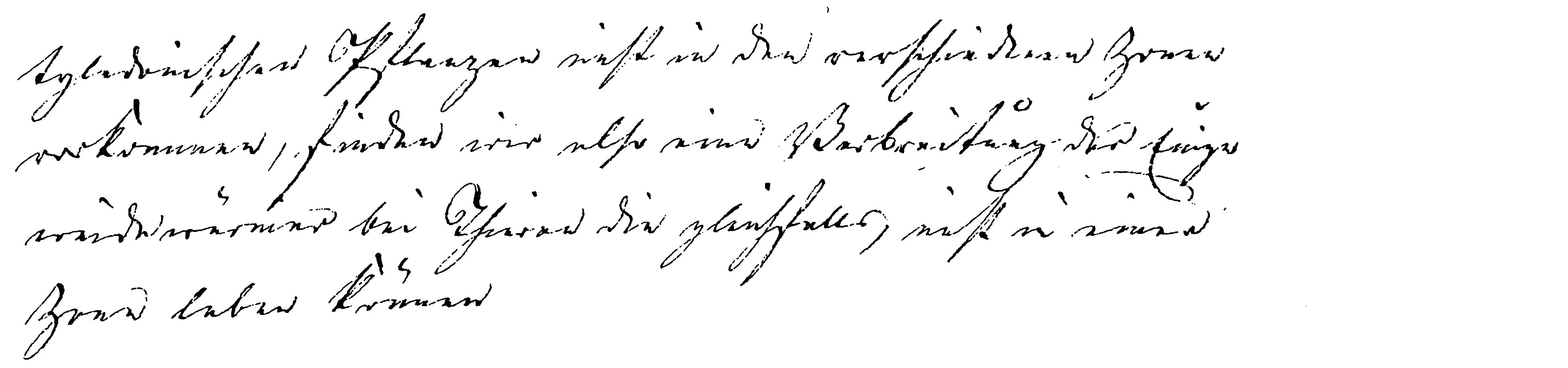} \\ \\
\end{tabular}}
\caption{Qualitative analysis on output generated by \textbf{T2T-BinFormer} on various \textbf{machine-typed} DIBCO and H-DIBCO Images. Images from Left to Right are: Original, Ground Truth, and Binarized output.}
\label{fig:Output_Handwritten}
\end{center}
\end{figure}

\begin{figure}[htbp]
\begin{center}
\captionsetup{justification=centering}
\resizebox{9cm}{!}{
\begin{tabular}{cccc}
\includegraphics[width=1\columnwidth]{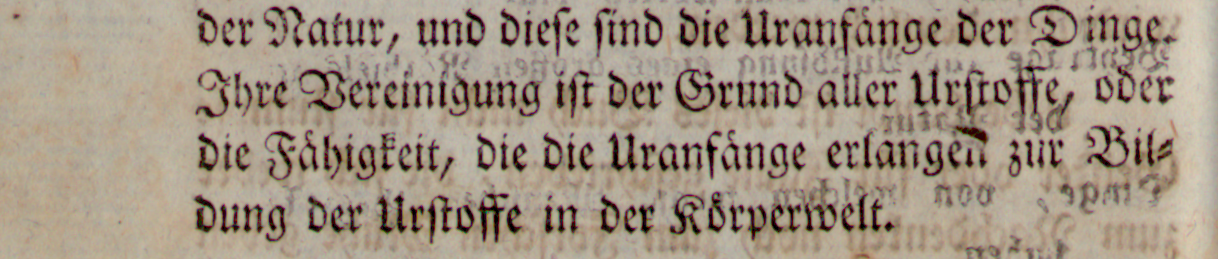} &
\includegraphics[width=1\columnwidth]{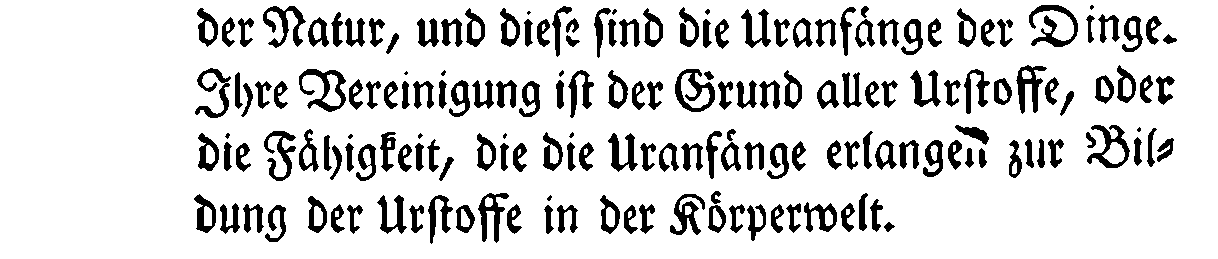} &
\includegraphics[width=1\columnwidth]{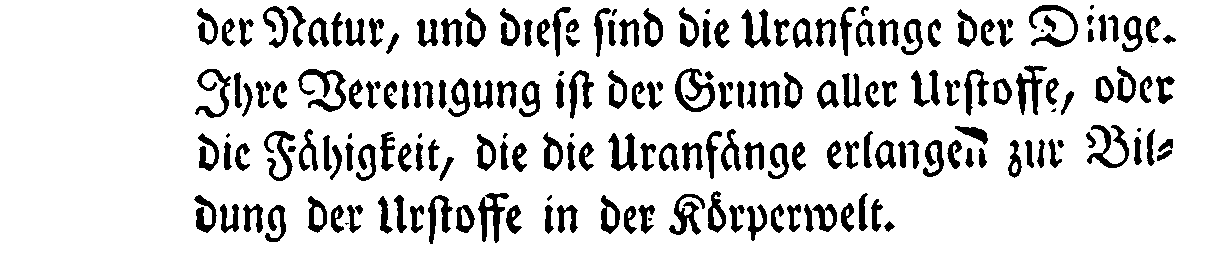} \\ \\
\includegraphics[width=1\columnwidth]{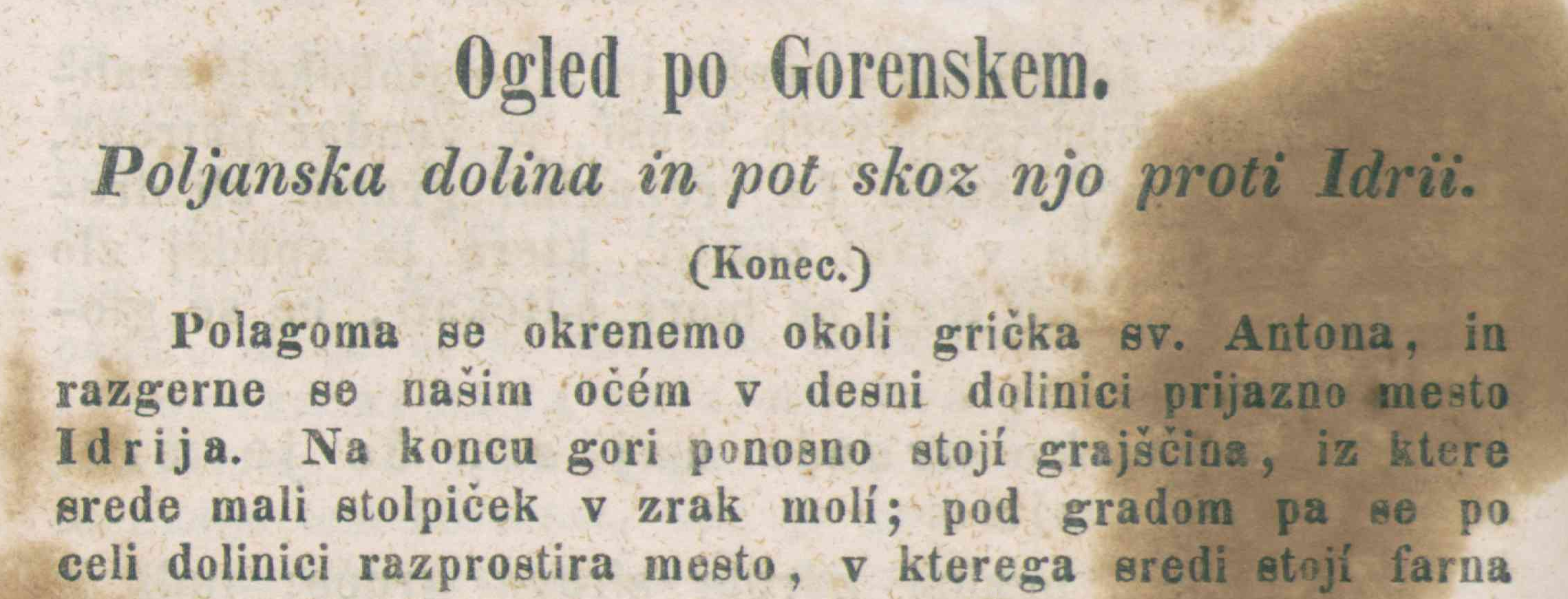} &
\includegraphics[width=1\columnwidth]{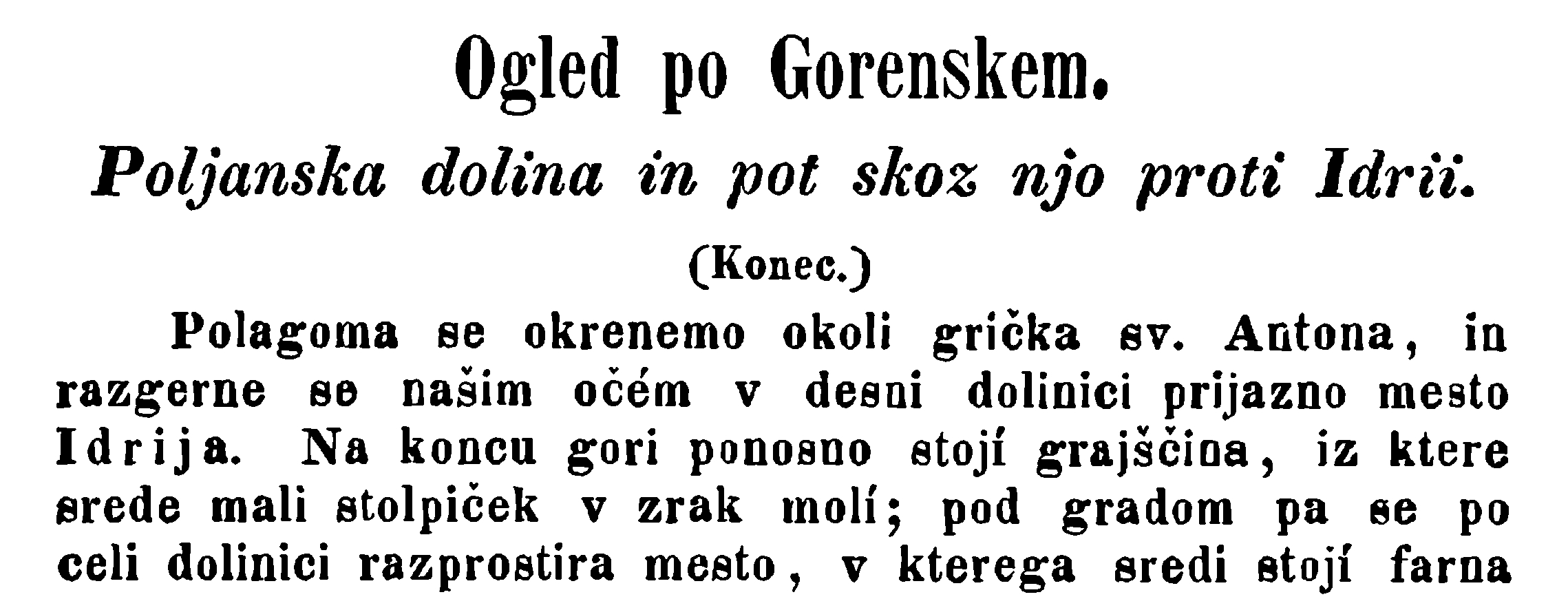} &
\includegraphics[width=1\columnwidth]{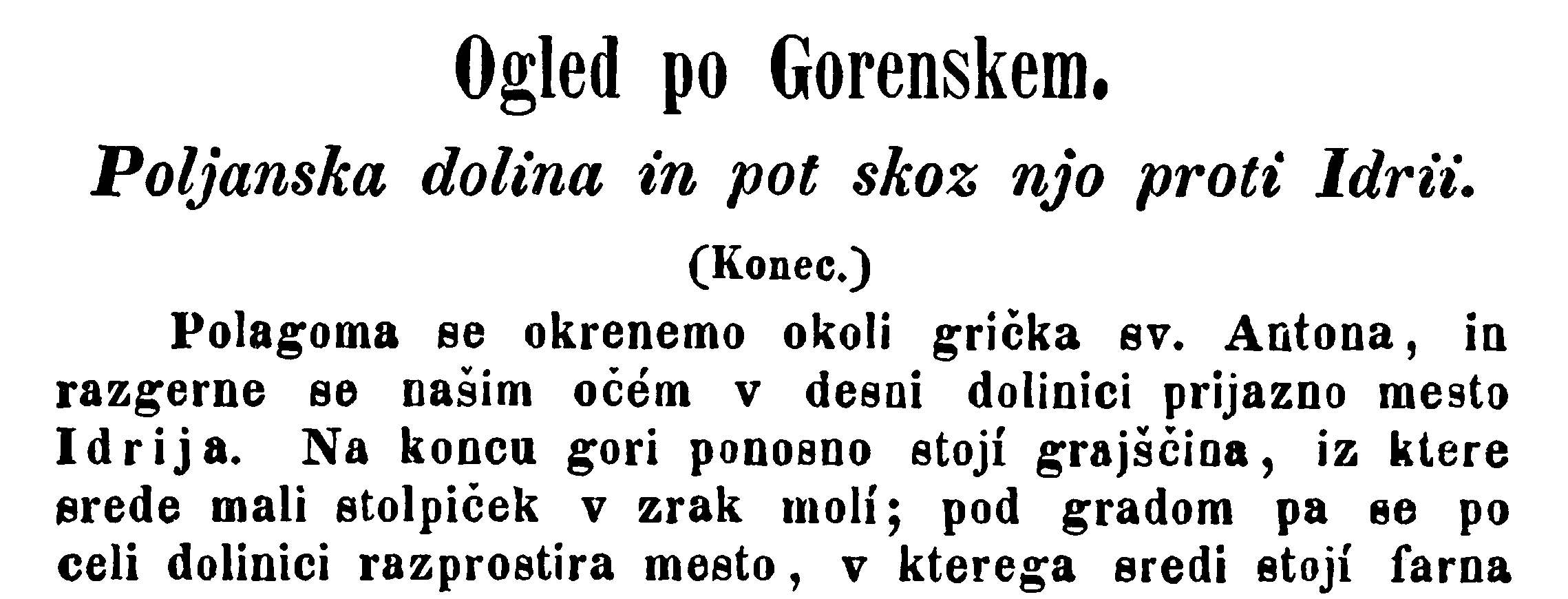} \\ \\
\includegraphics[width=1\columnwidth]{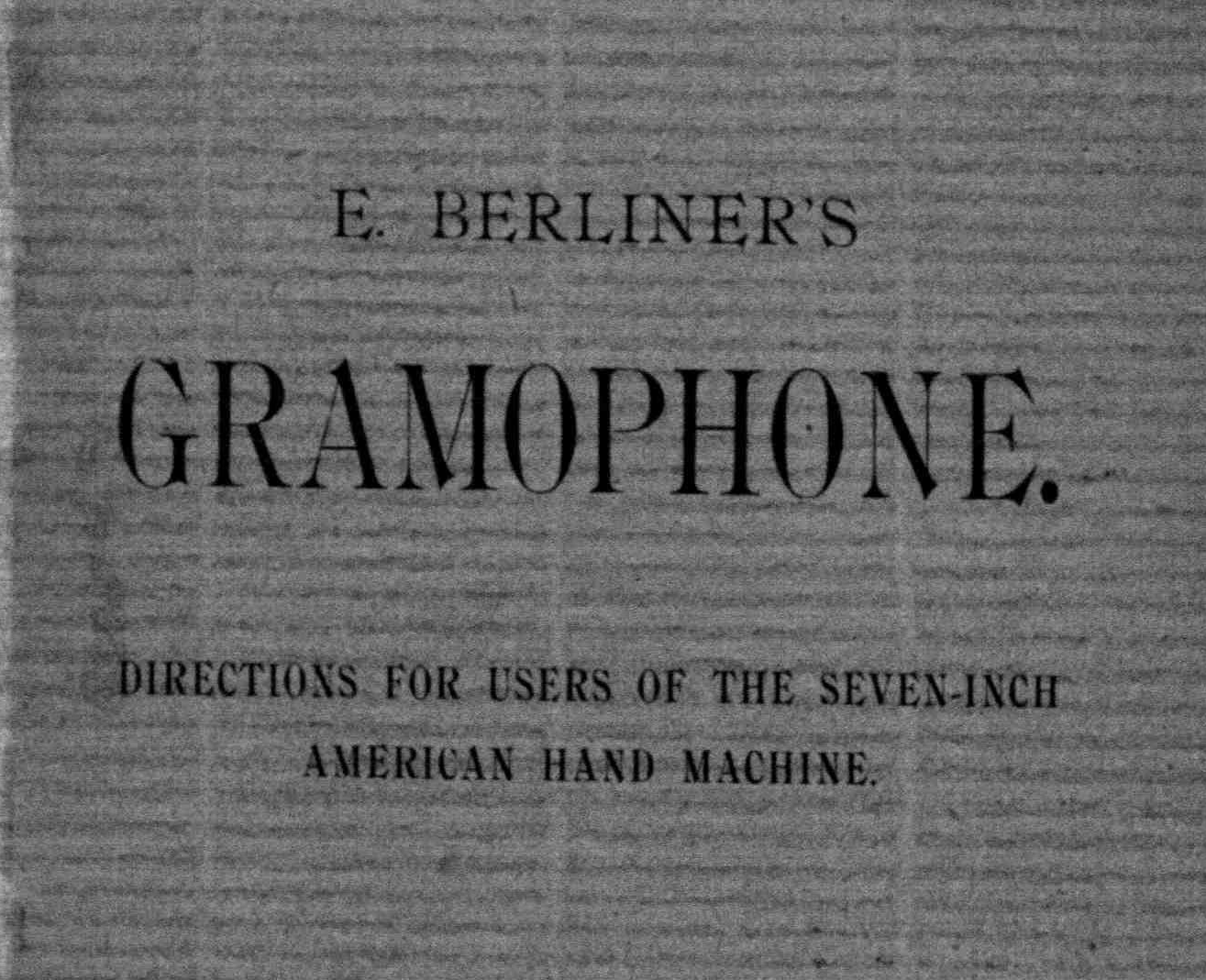} &
\includegraphics[width=1\columnwidth]{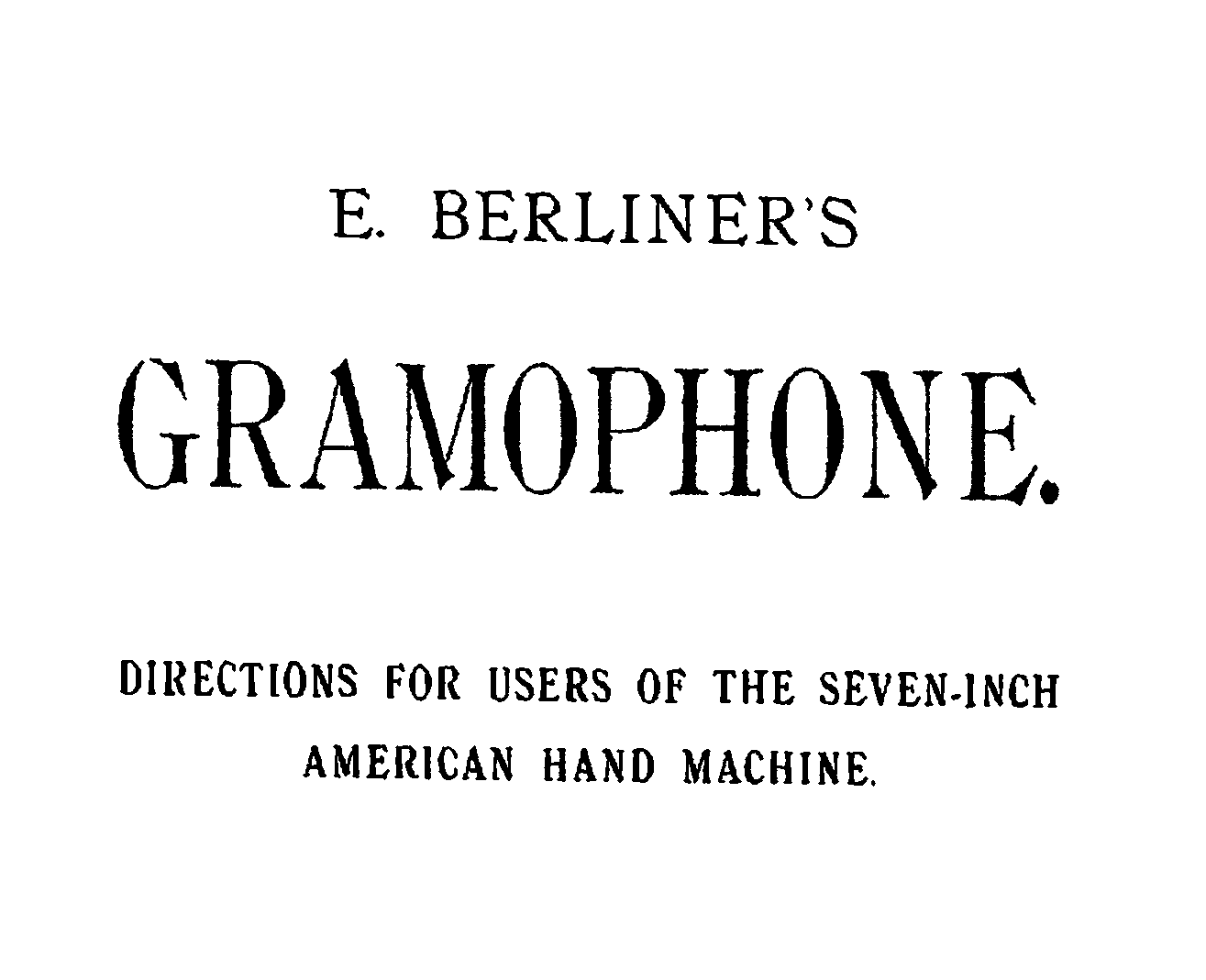} &
\includegraphics[width=0.9\columnwidth]{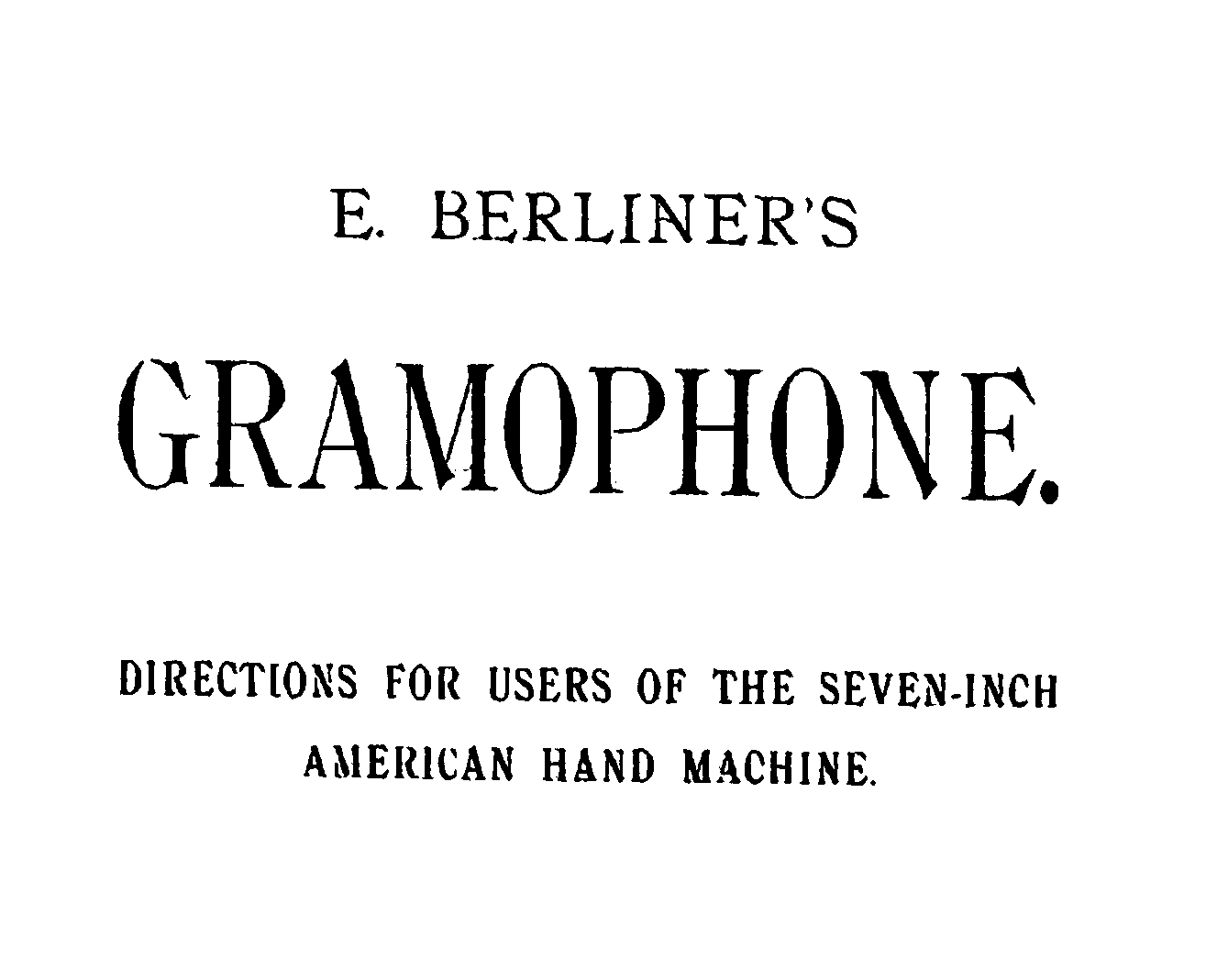} \\
\includegraphics[width=1\columnwidth]{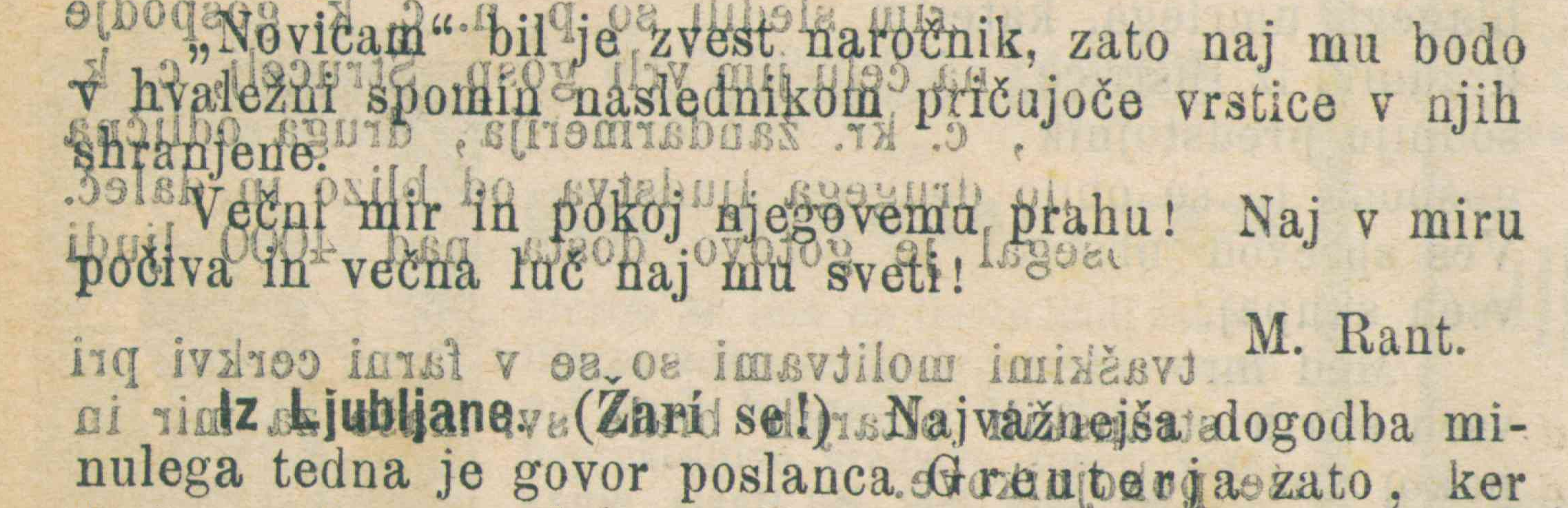} &
\includegraphics[width=1\columnwidth]{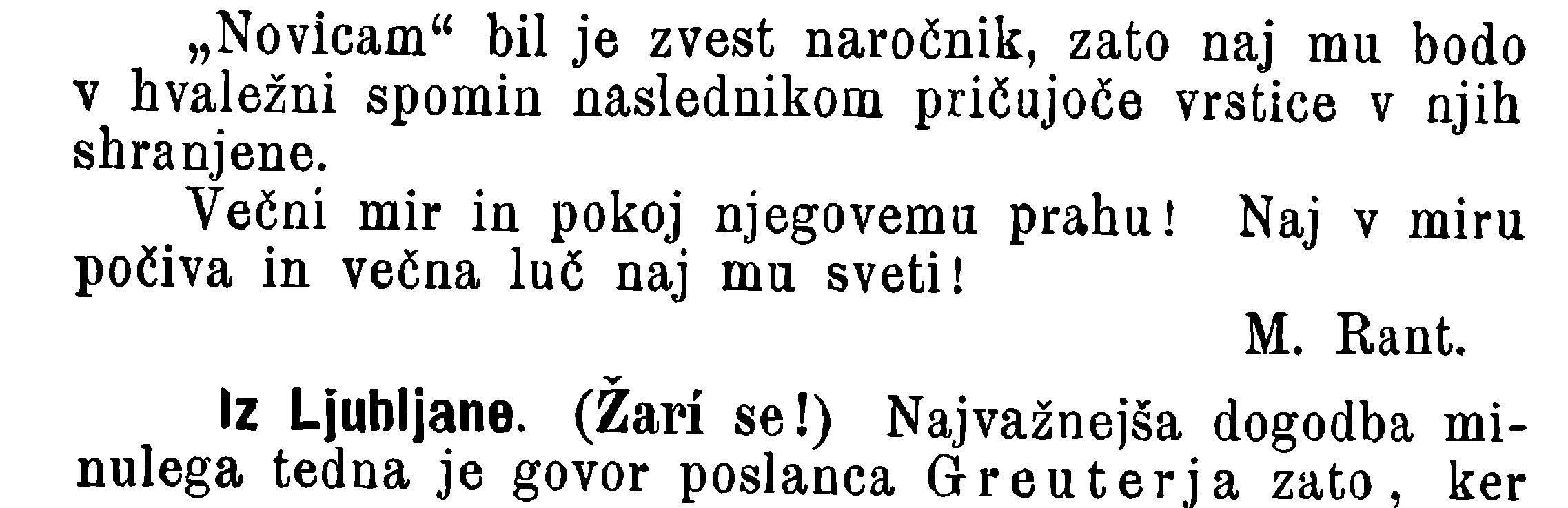} &
\includegraphics[width=1\columnwidth]{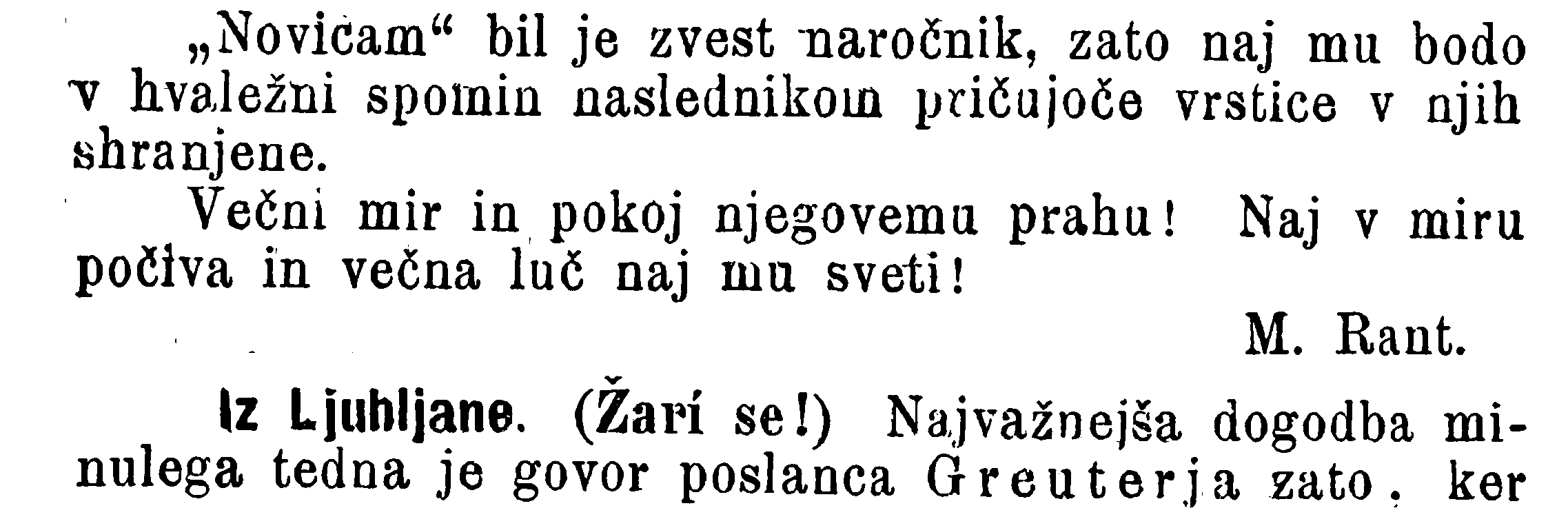} \\ \\
\includegraphics[width=1\columnwidth]{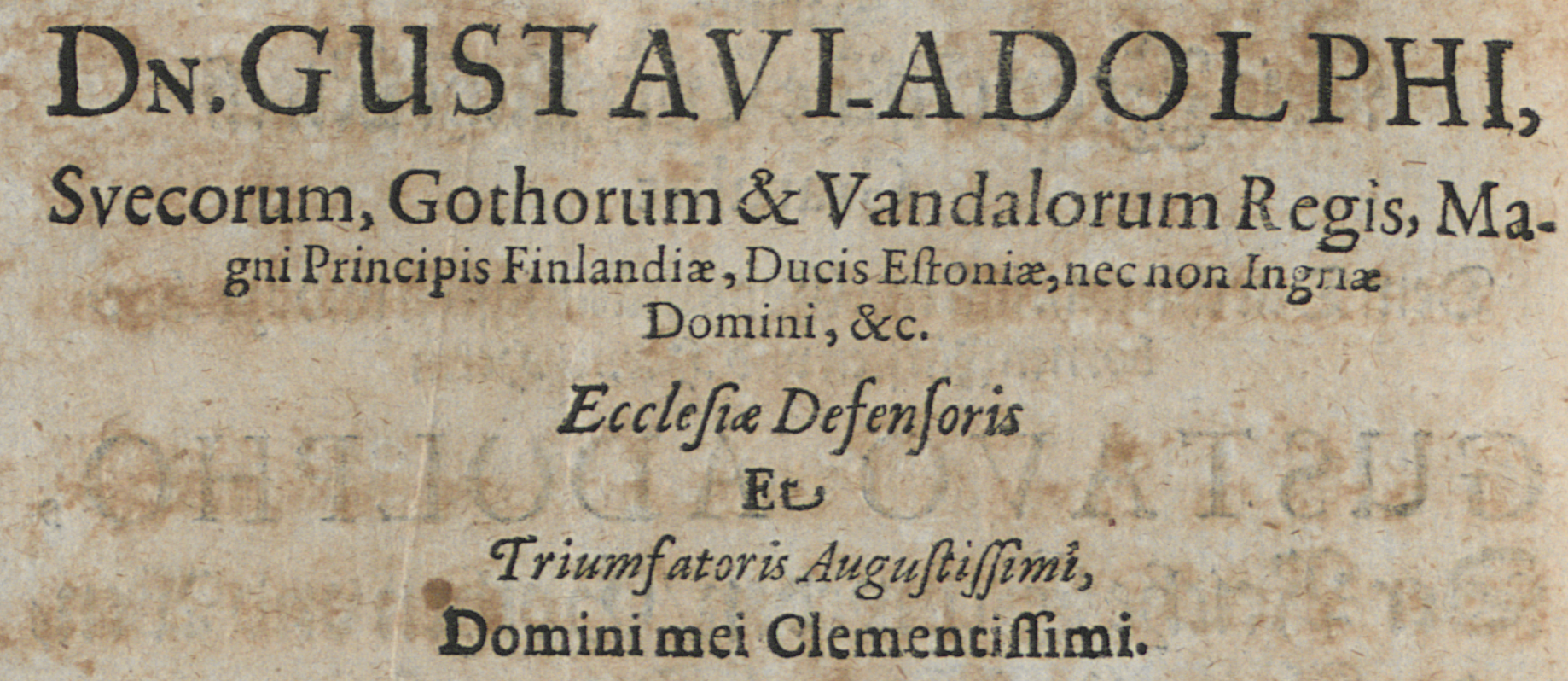} &
\includegraphics[width=1\columnwidth]{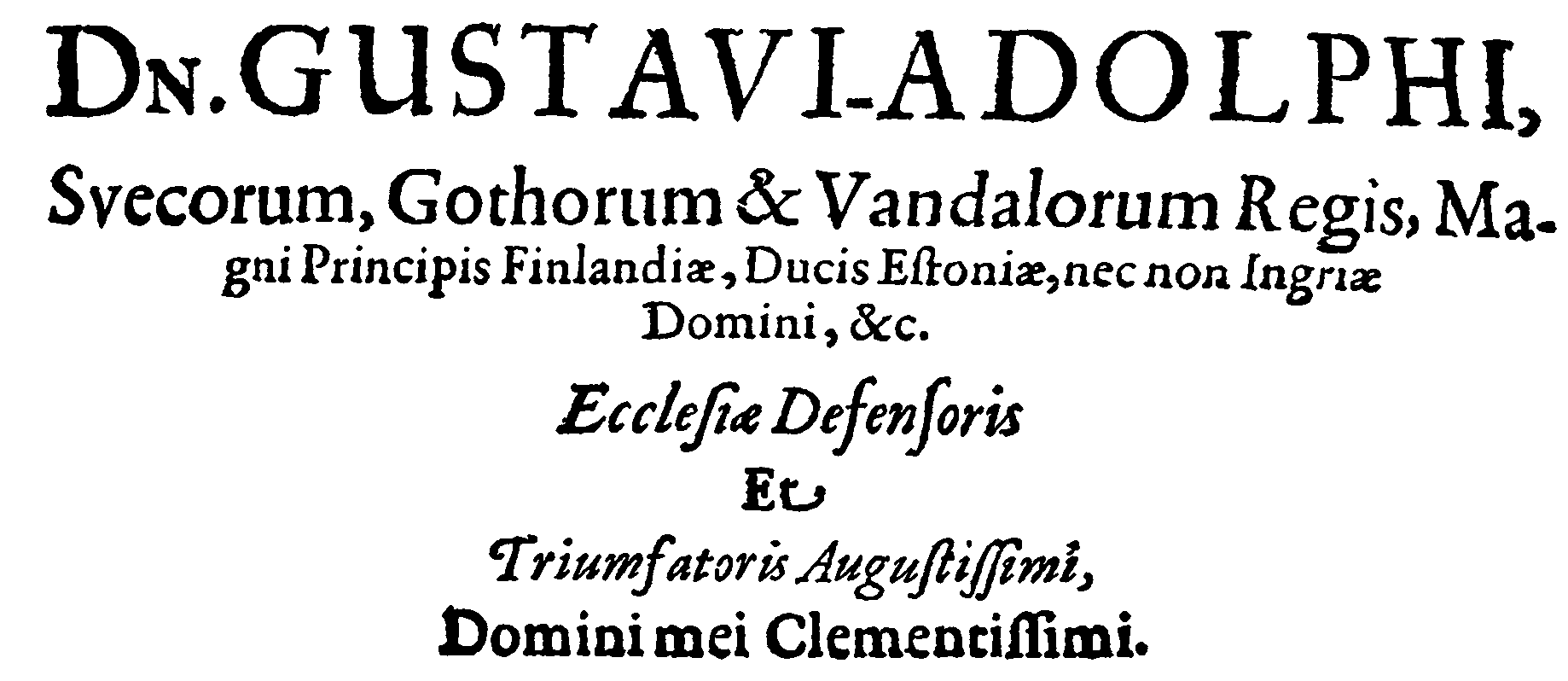} &
\includegraphics[width=1\columnwidth]{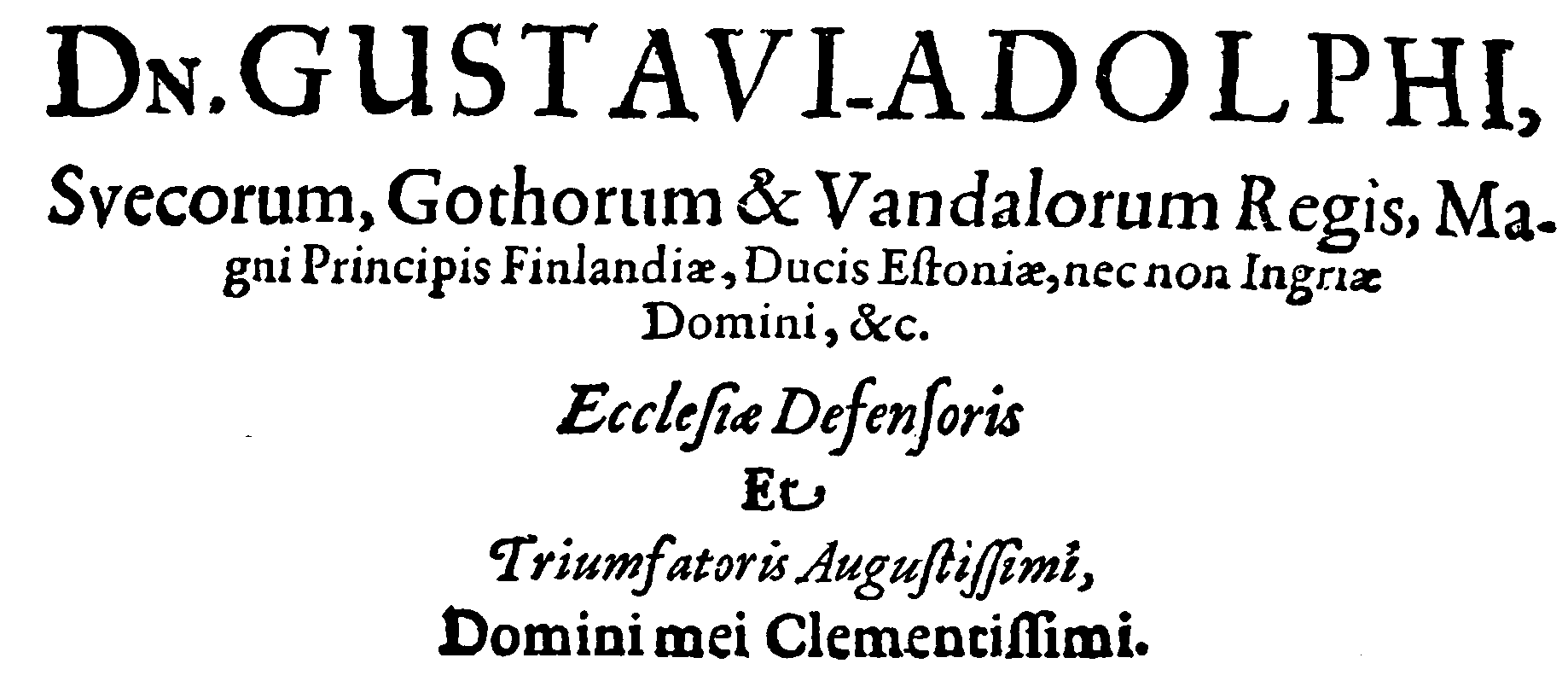} \\ \\
\includegraphics[width=1\columnwidth]{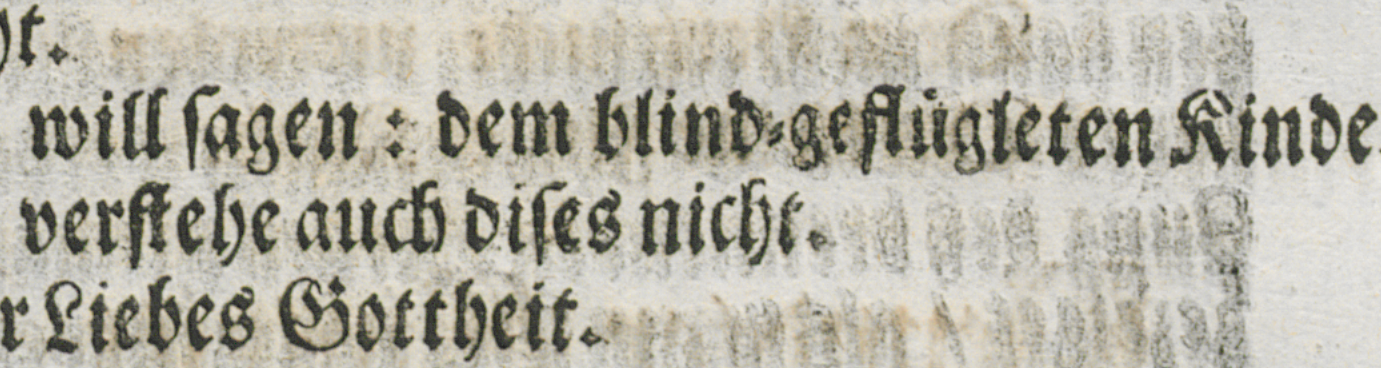} &
\includegraphics[width=1\columnwidth]{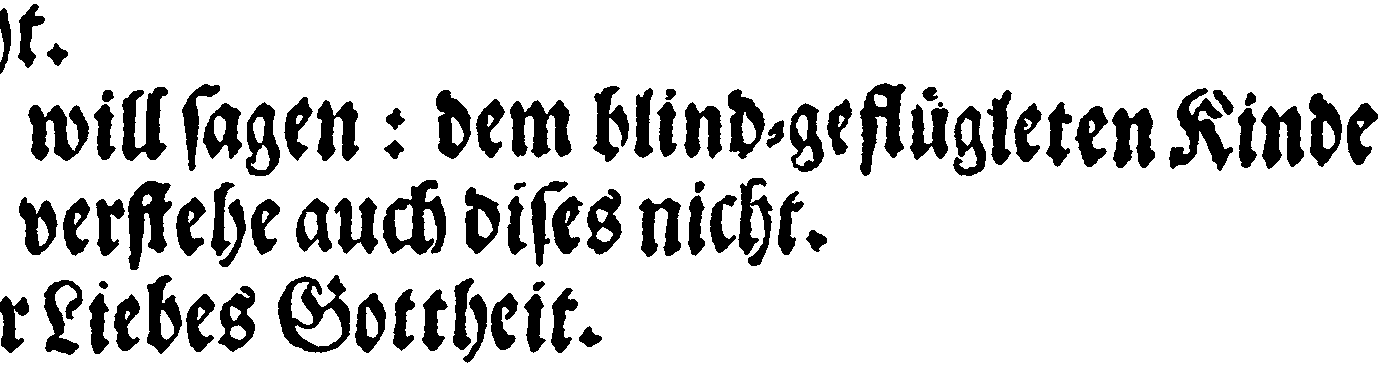} &
\includegraphics[width=1\columnwidth]{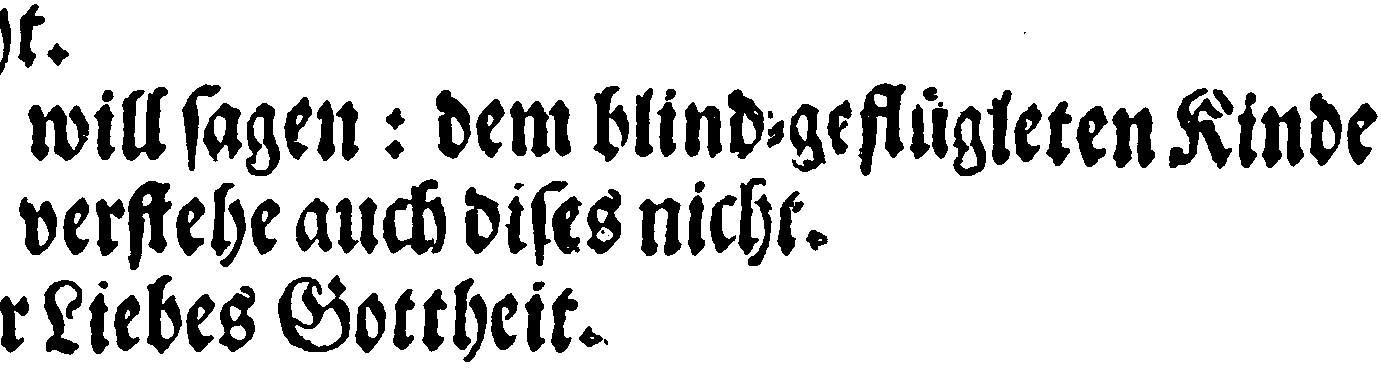} \\ \\
\end{tabular}}
\caption{Qualitative Analysis on Output generated by \textbf{T2T-BinFormer} on various \textbf{machine-typed} (H-)DIBCO Images. Images from Left to Right are Original, Ground Truth, and Binarized output.}
\label{fig:Output_Types}
\end{center}
\end{figure}

\begin{figure*}[ht]
\centering
\begin{tabular}{ccccc}
{\includegraphics[width = 0.45\columnwidth]{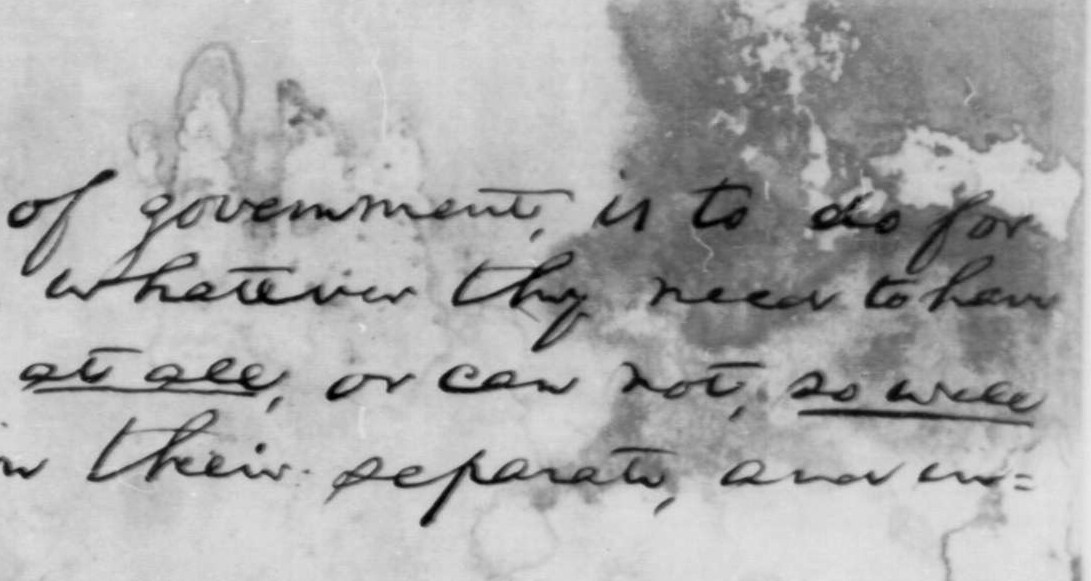}} &
{\includegraphics[width=0.45\columnwidth]{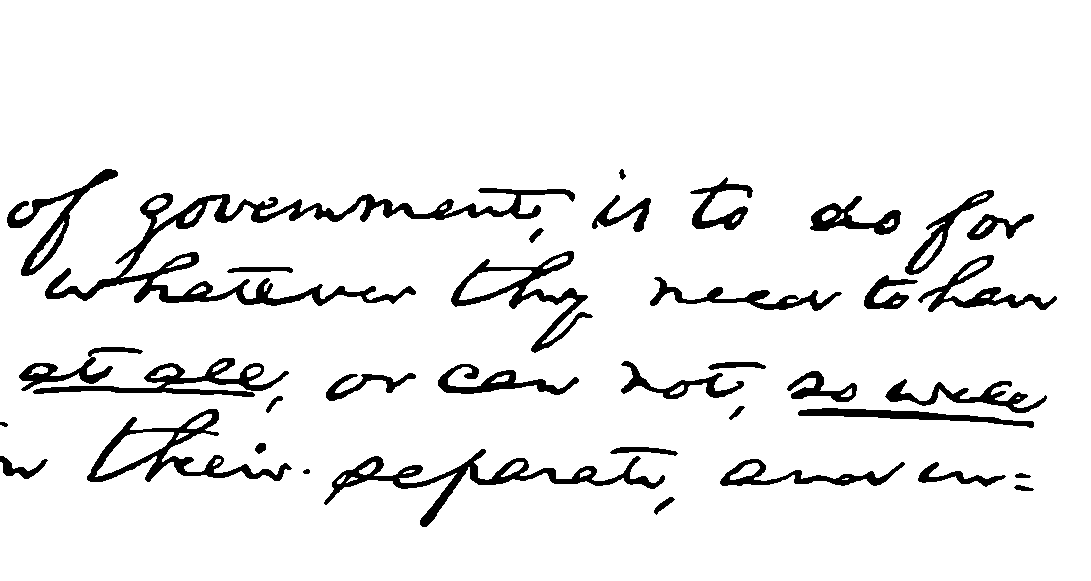}} &
{\includegraphics[width=0.45\columnwidth]{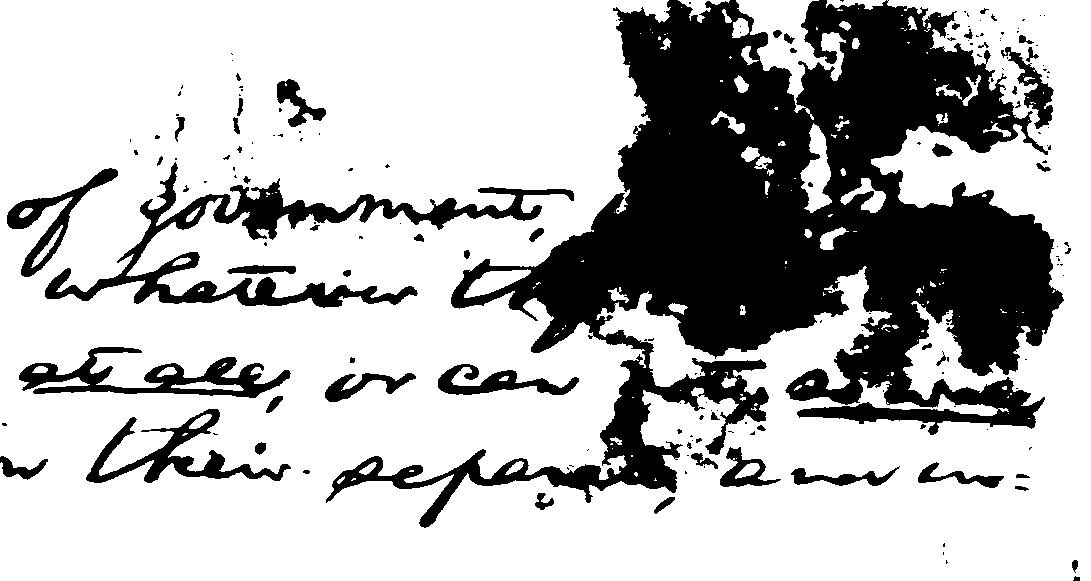}}&
{\includegraphics[width = 0.45\columnwidth]{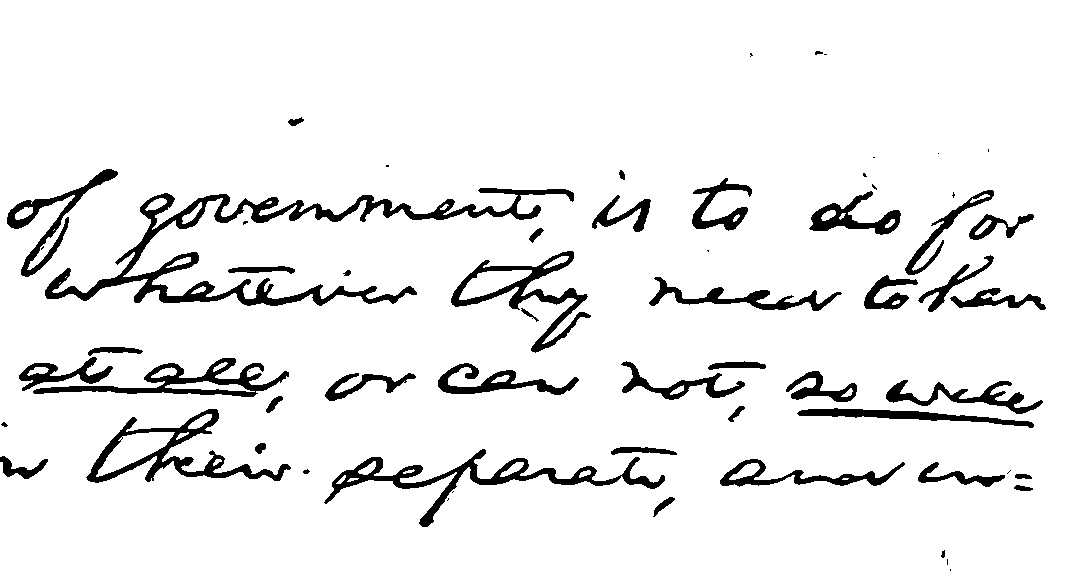}} & \\
\scriptsize a) Original &  \scriptsize b) GT  &  \scriptsize c) Otsu~\citep{otsu1979threshold}  &  \scriptsize d) SauvolaNet~\citep{li2021sauvolanet}\\ 
{\includegraphics[width=0.45\columnwidth]{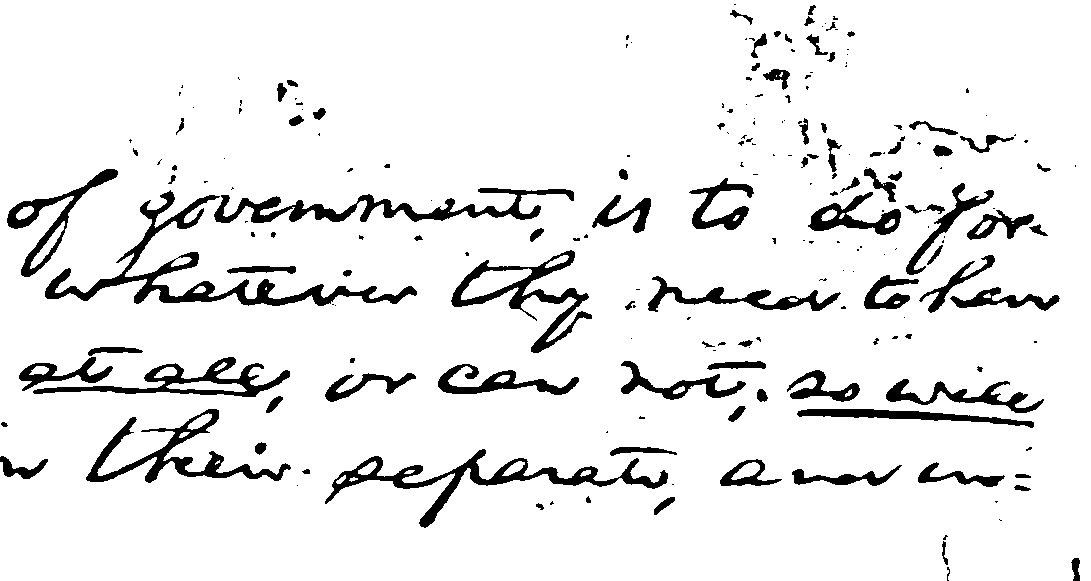}} &
{\includegraphics[width=0.45\columnwidth]{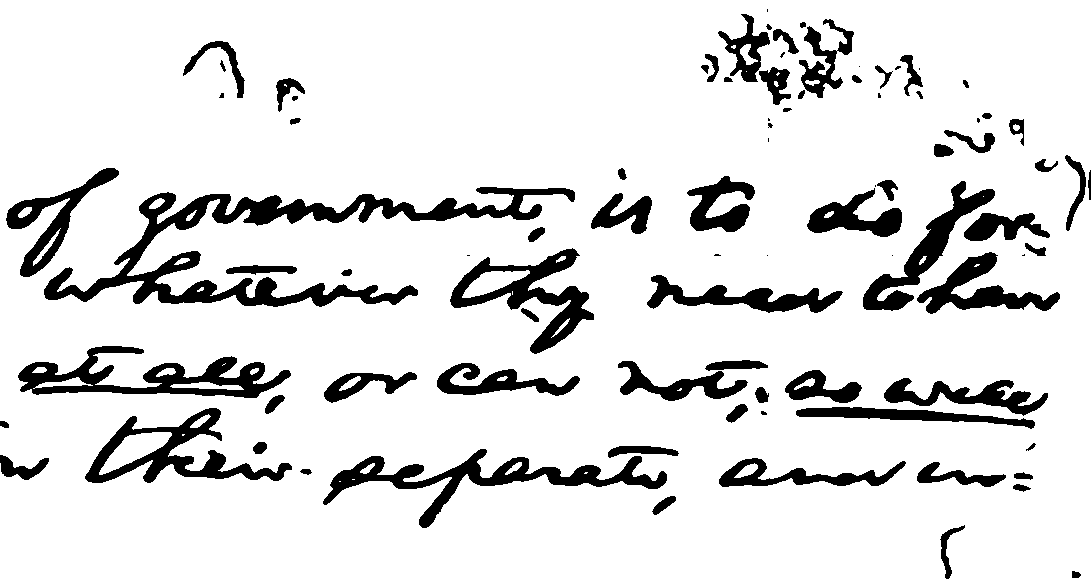}}&
{\includegraphics[width=0.45\columnwidth]{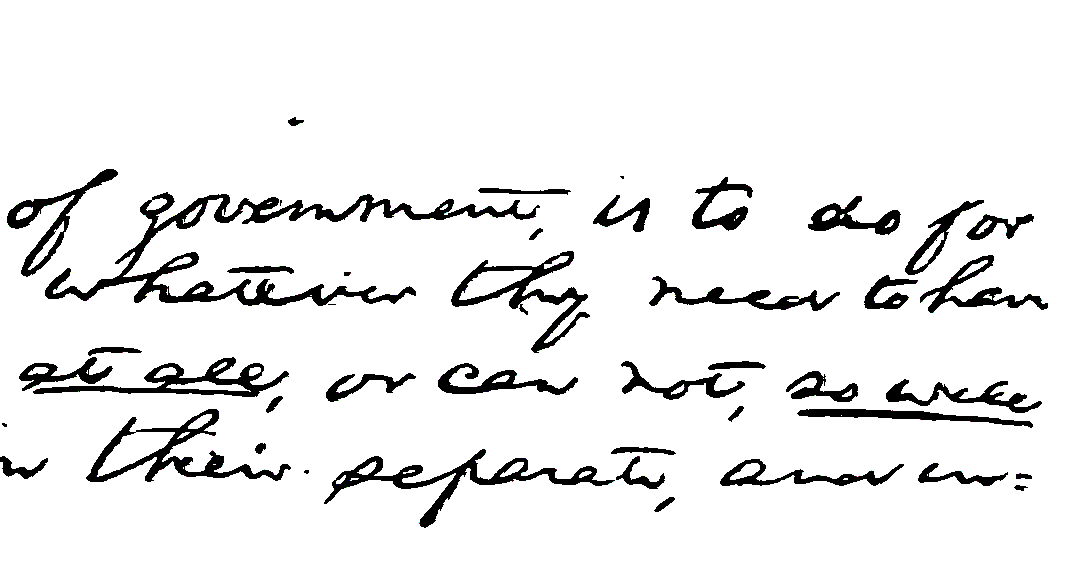}} &
{\includegraphics[width = 0.45\columnwidth]{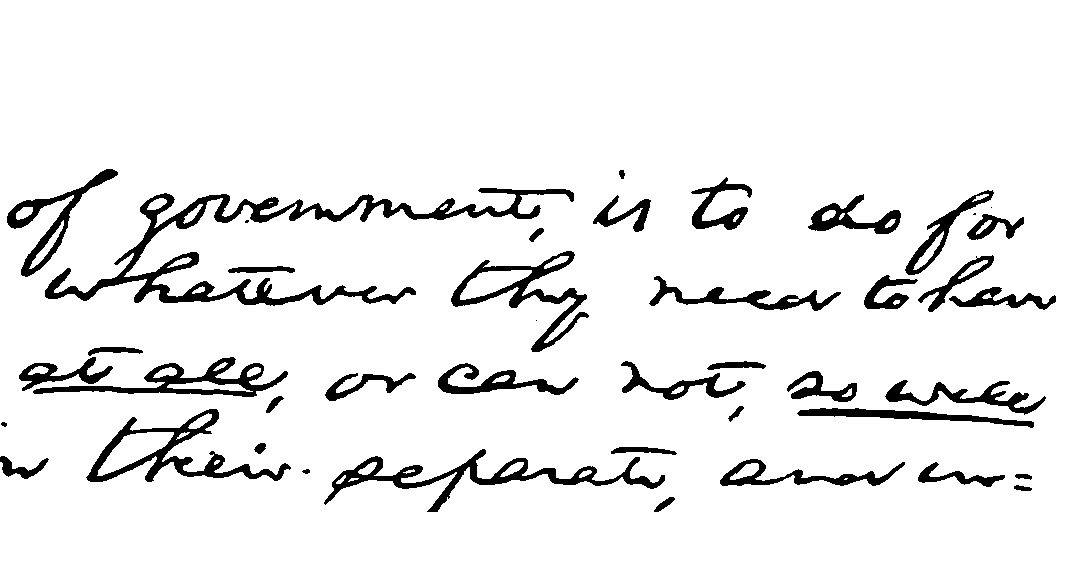}} &\\
\scriptsize e) Bradley~\citep{bradley2007adaptive} & \scriptsize f) DE-GAN~\citep{souibgui2020gan}  & \scriptsize g) DocEnTr~\citep{souibgui2022docentr}  &  \scriptsize h) \textbf{T2T-BinFormer}\\
\end{tabular}
\caption{Qualitative performance of the various binarization techniques on sample no. 4 from the DIBCO 2009 dataset.}
\label{fig:compDIBCO2009}
\end{figure*}

\begin{figure*}[ht]
\centering
\begin{tabular}{ccccc}
{\includegraphics[width = 0.45\columnwidth]{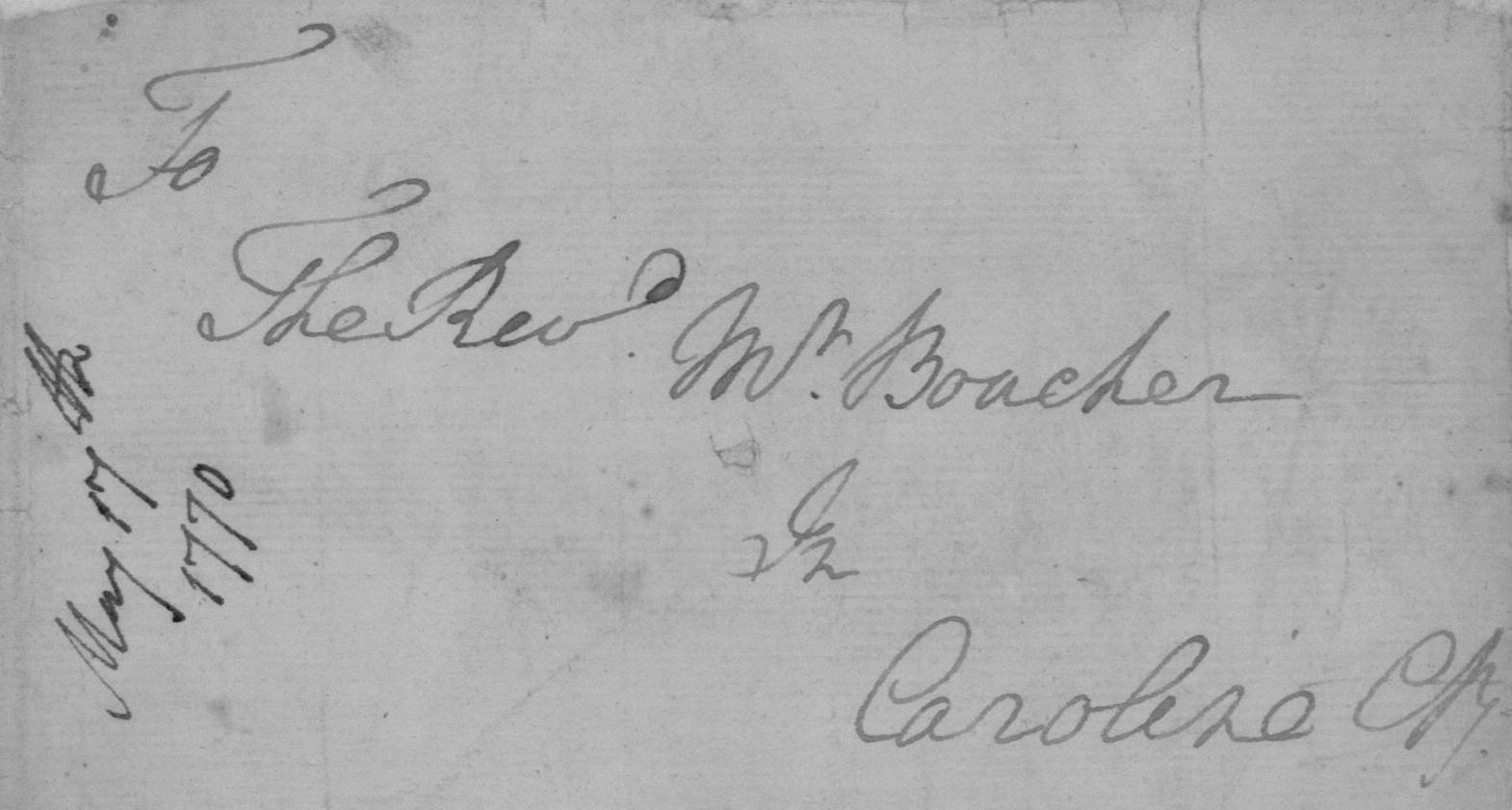}} &
{\includegraphics[width=0.45\columnwidth]{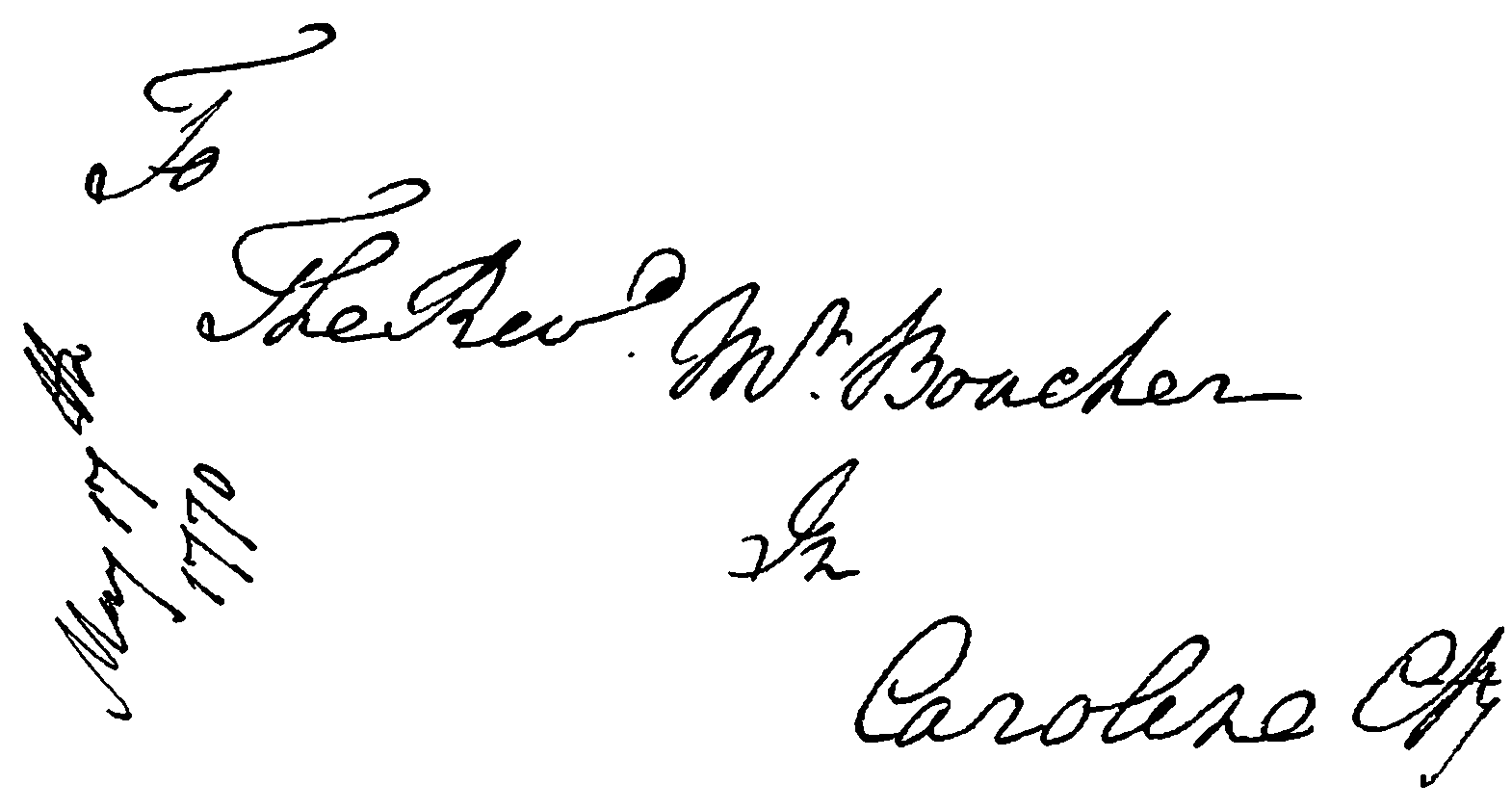}} &
{\includegraphics[width=0.45\columnwidth]{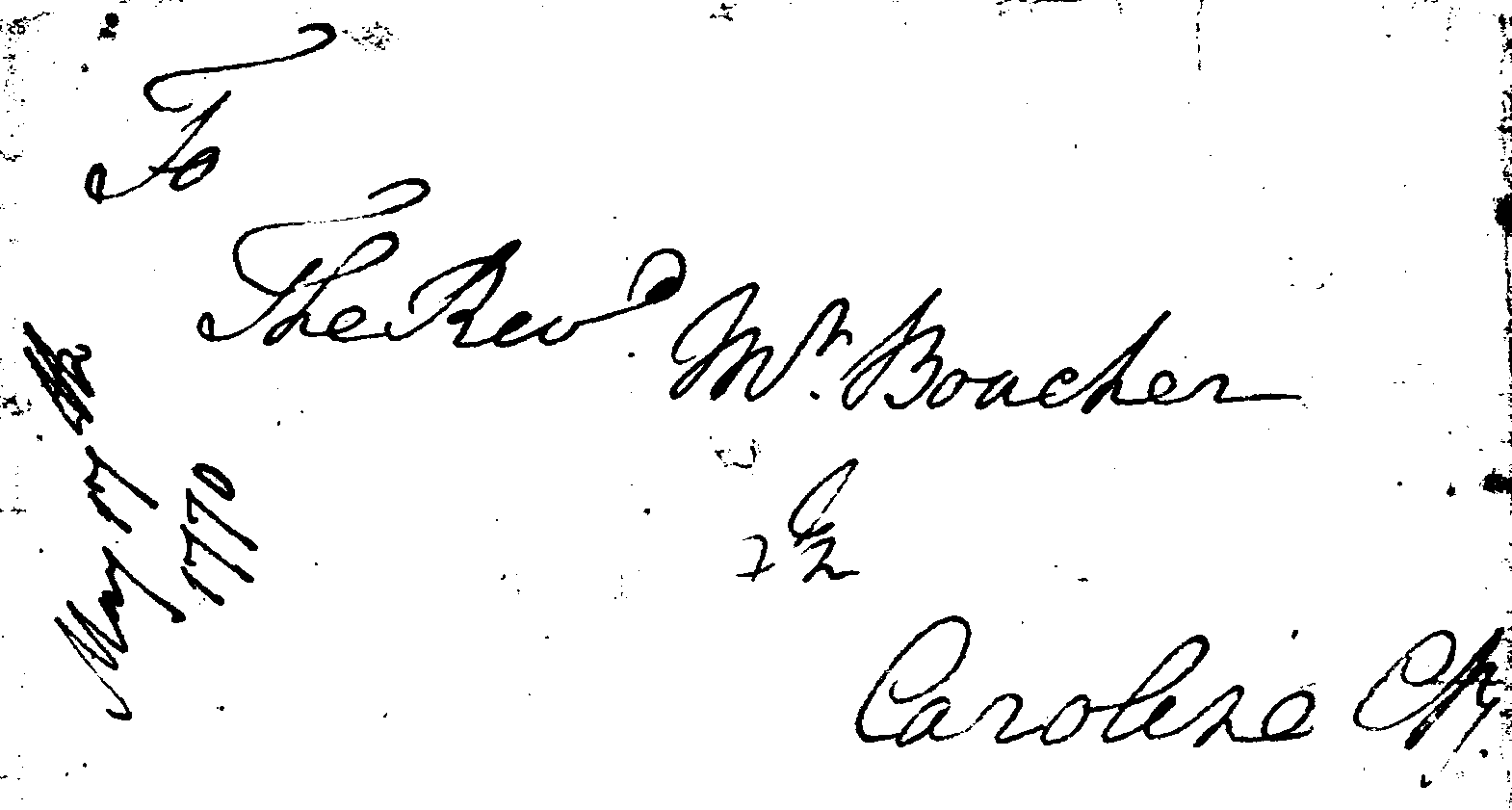}}&
{\includegraphics[width = 0.45\columnwidth]{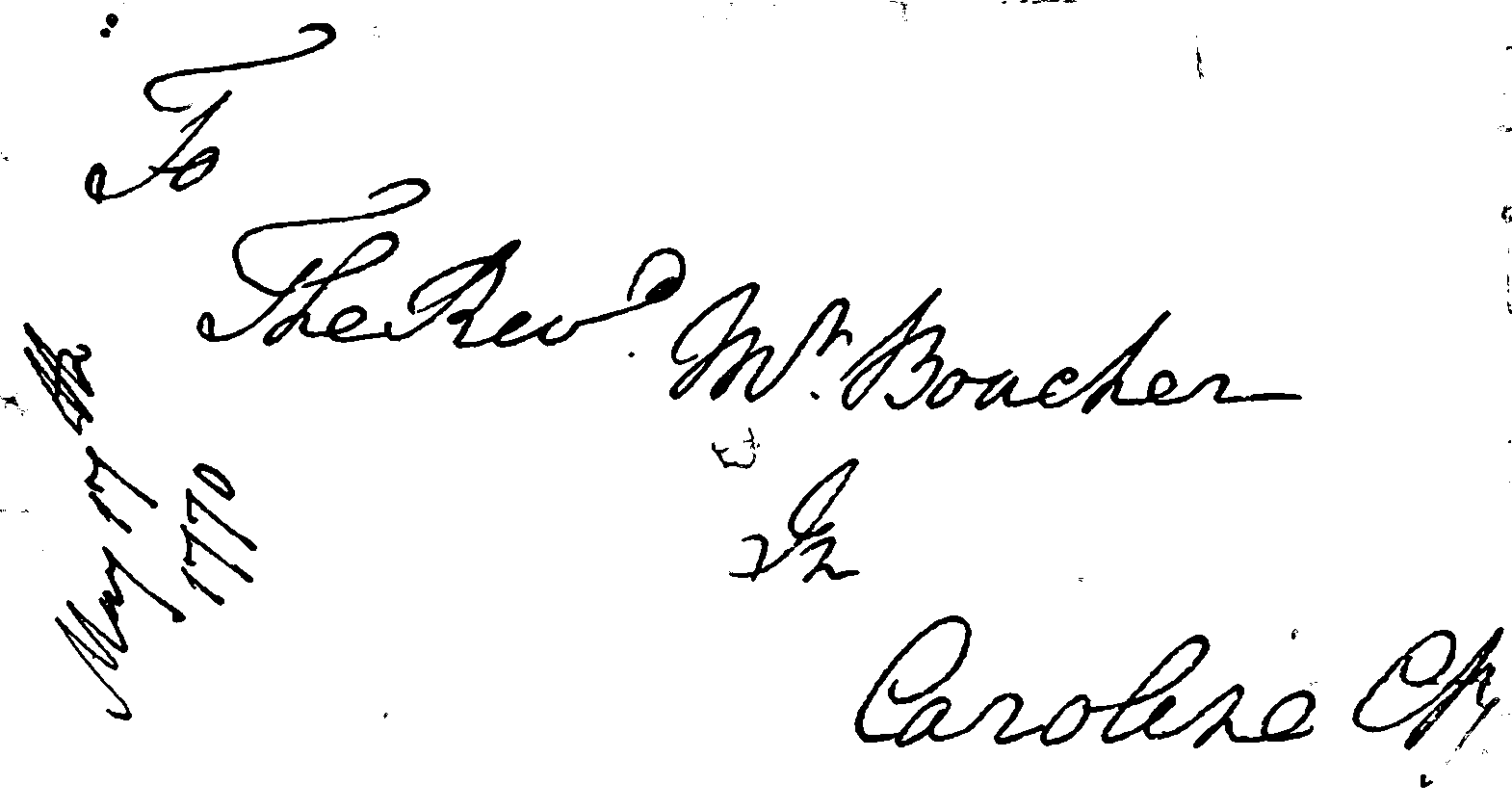}} & \\
\scriptsize a) Original &  \scriptsize b) GT  &  \scriptsize c) Otsu~\citep{otsu1979threshold}  &  \scriptsize d) SauvolaNet~\citep{li2021sauvolanet}\\ 
{\includegraphics[width=0.45\columnwidth]{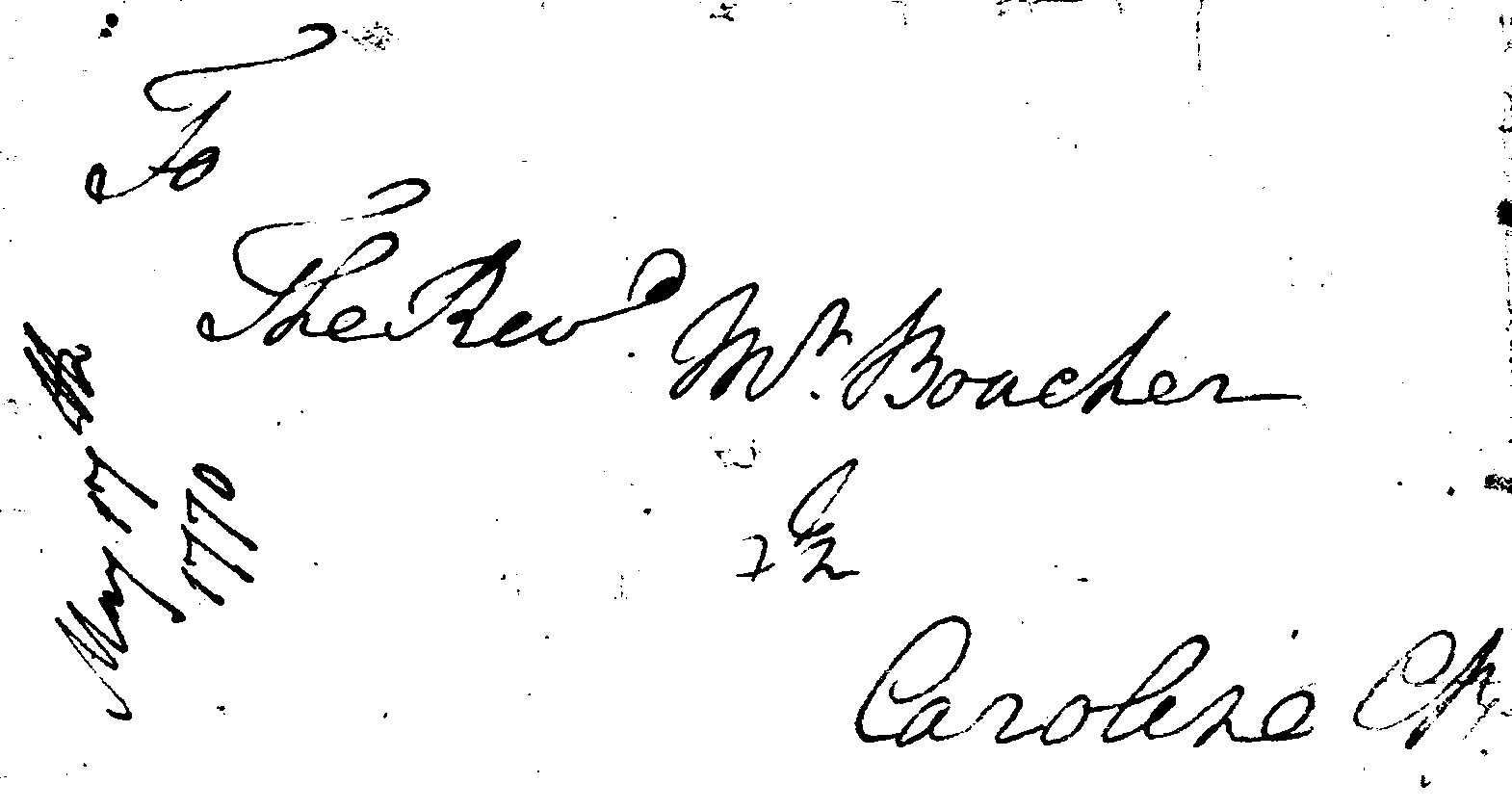}} &
{\includegraphics[width=0.45\columnwidth]{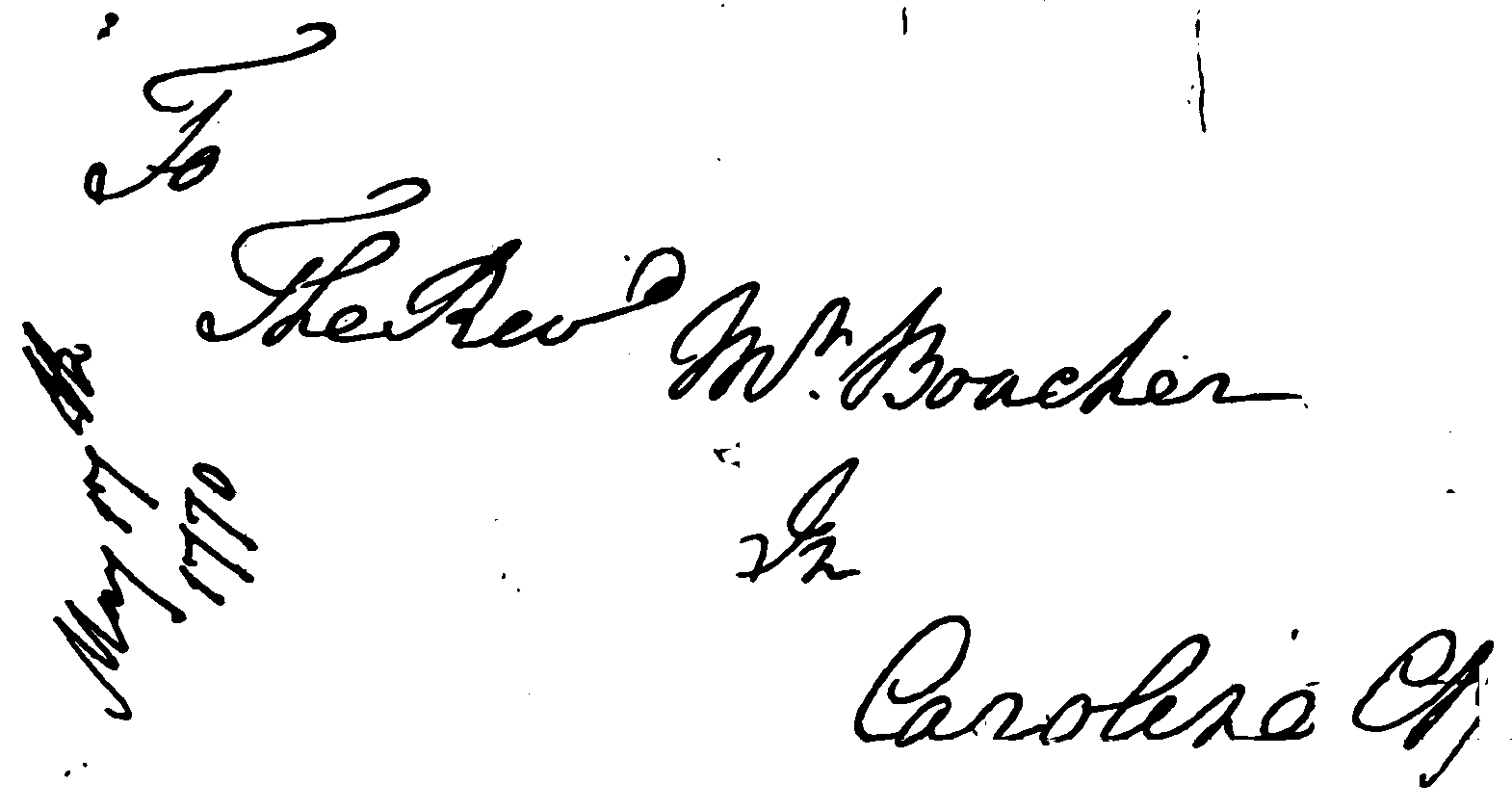}}&
{\includegraphics[width=0.45\columnwidth]{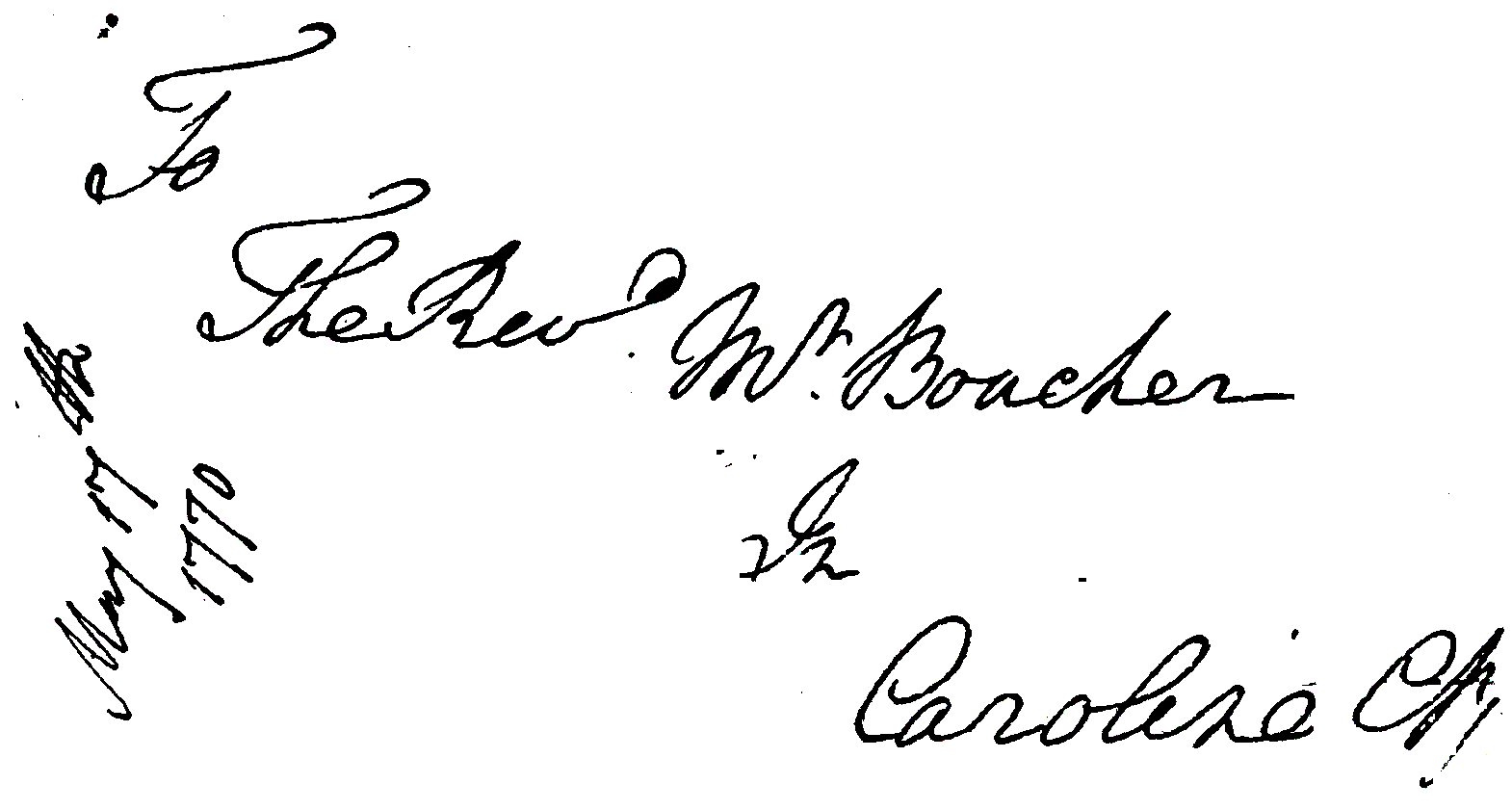}} &
{\includegraphics[width = 0.45\columnwidth]{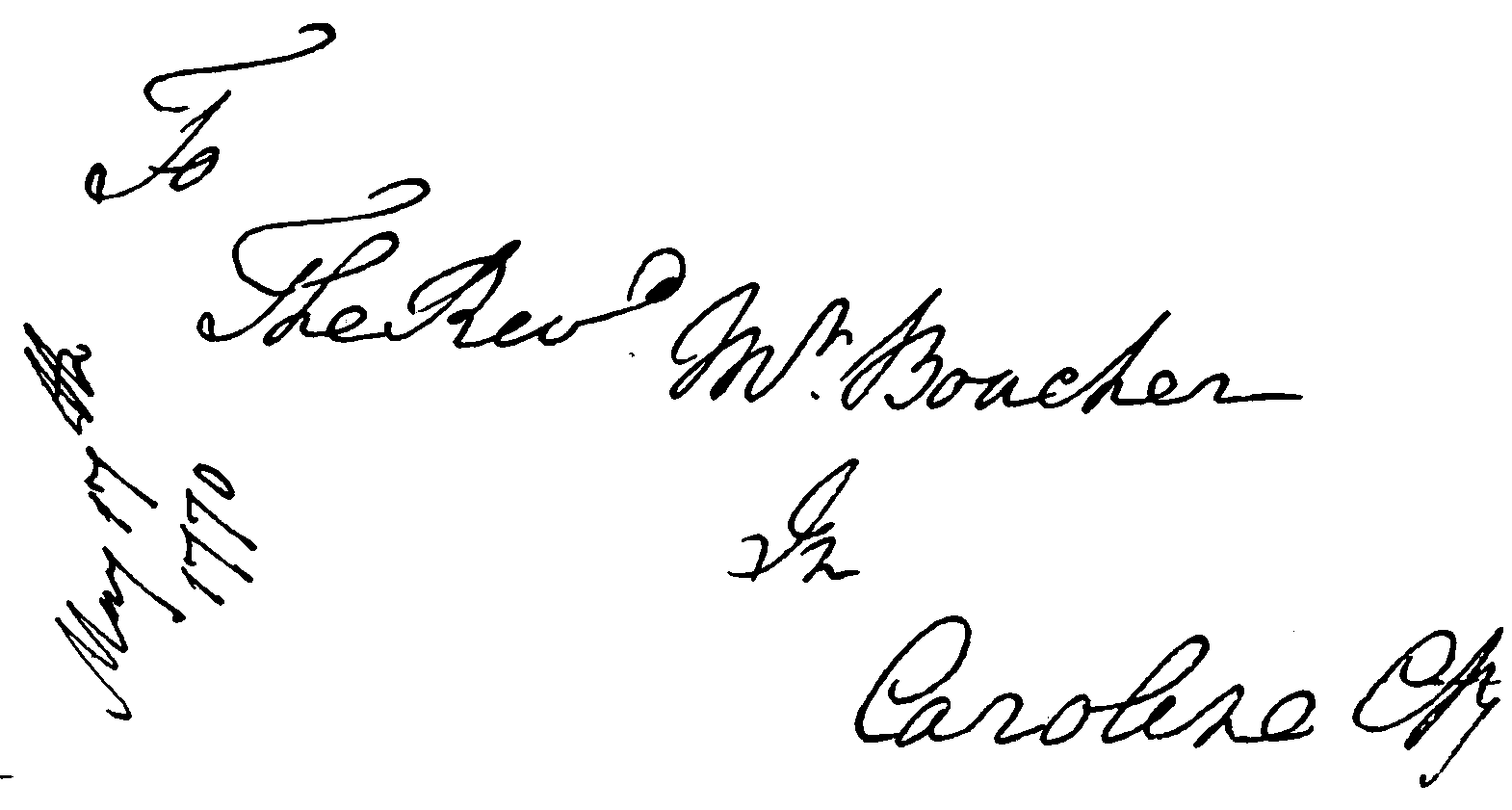}} &\\
\scriptsize e) Bradley~\citep{bradley2007adaptive} & \scriptsize f) DE-GAN~\citep{souibgui2020gan}  & \scriptsize g) DocEnTr~\citep{souibgui2022docentr}  &  \scriptsize h) \textbf{T2T-BinFormer}\\
\end{tabular}
\caption{Qualitative performance of the various binarization techniques on sample no. 2 from the DIBCO 2010 dataset.}
\label{fig:compDIBCO2010}
\end{figure*}

\begin{figure*}[ht!]
\centering
\begin{tabular}{ccccc}
{\includegraphics[width = 0.45\columnwidth]{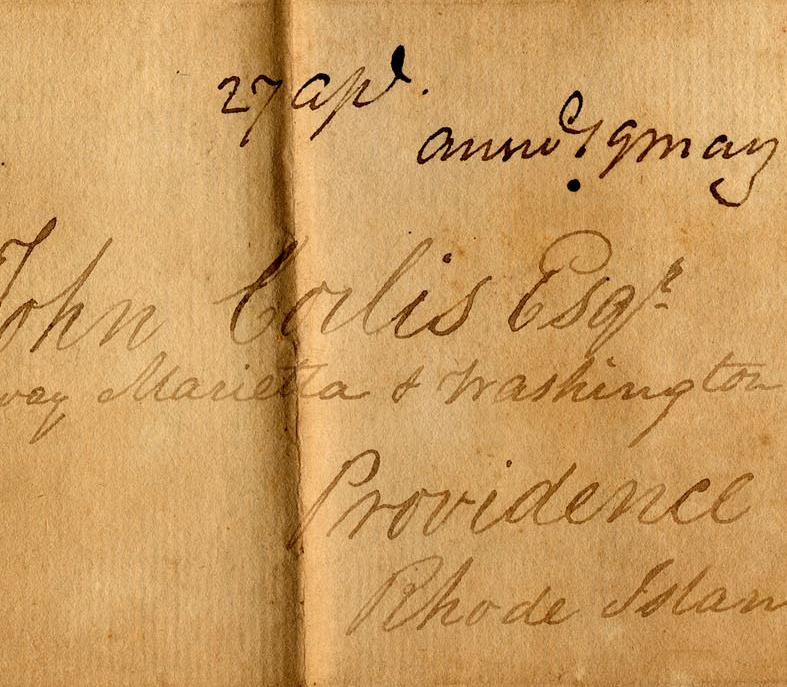}} &
{\includegraphics[width=0.45\columnwidth]{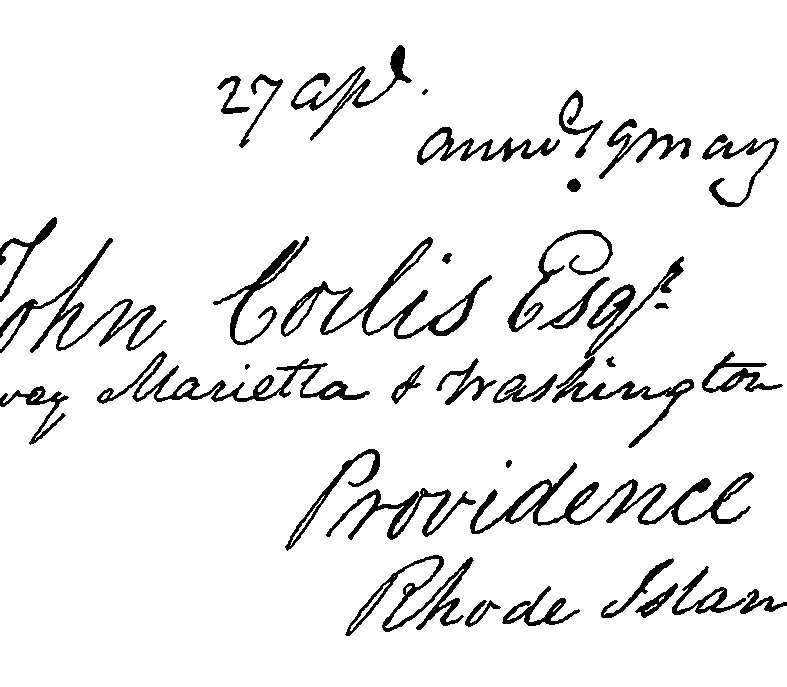}} &
{\includegraphics[width=0.45\columnwidth]{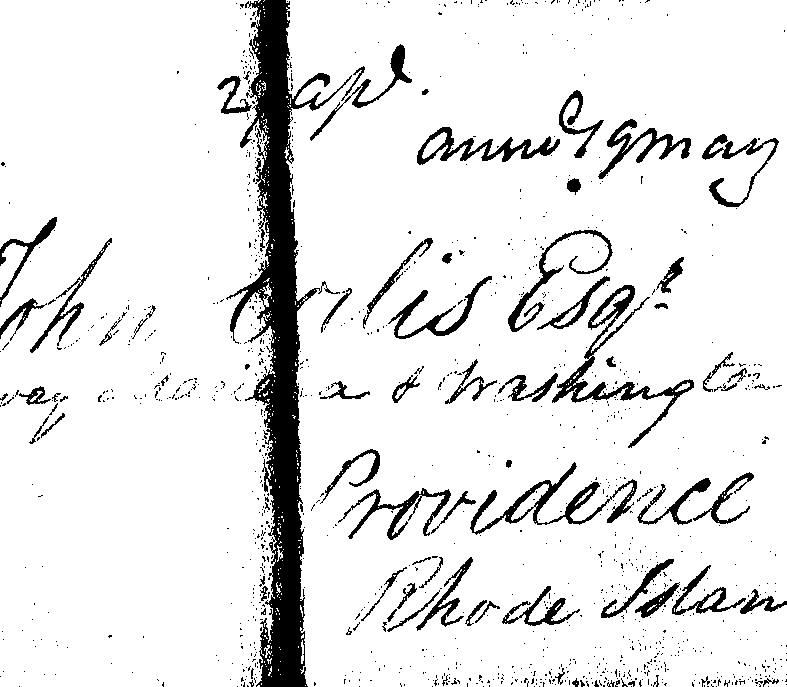}}&
{\includegraphics[width = 0.45\columnwidth]{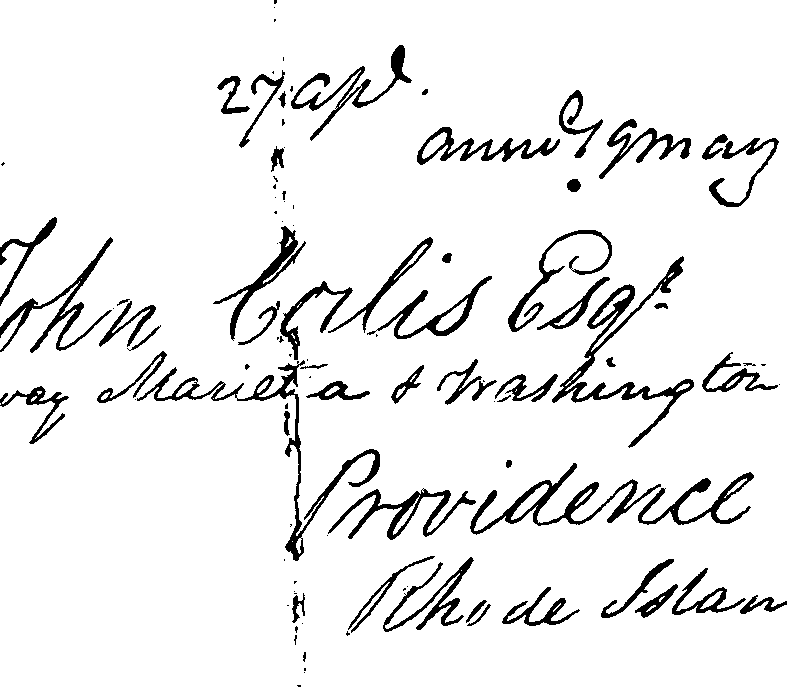}} & \\
\scriptsize a) Original &  \scriptsize b) GT  &  \scriptsize c) Otsu~\citep{otsu1979threshold}  &  \scriptsize d) SauvolaNet~\citep{li2021sauvolanet} \\ 
{\includegraphics[width=0.45\columnwidth]{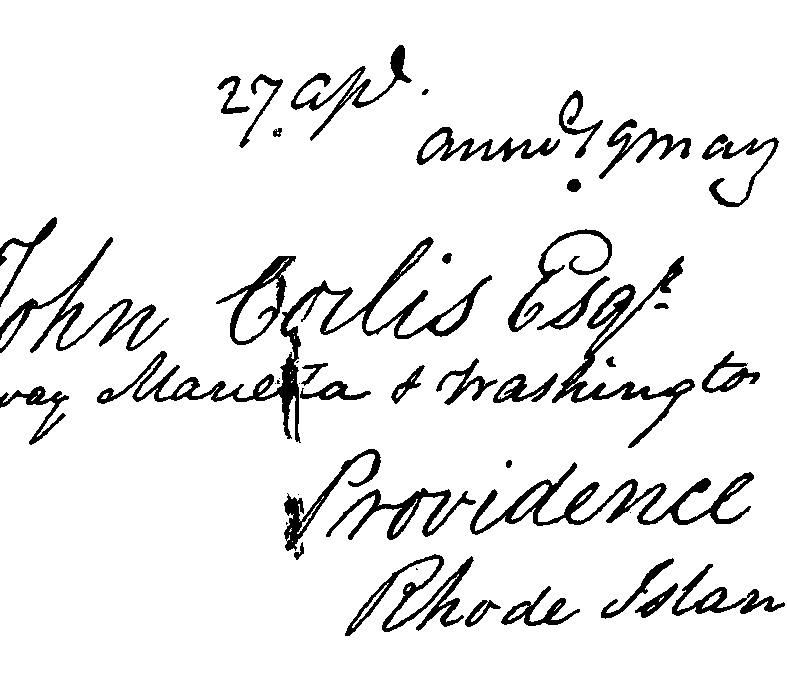}} &
{\includegraphics[width=0.45\columnwidth]{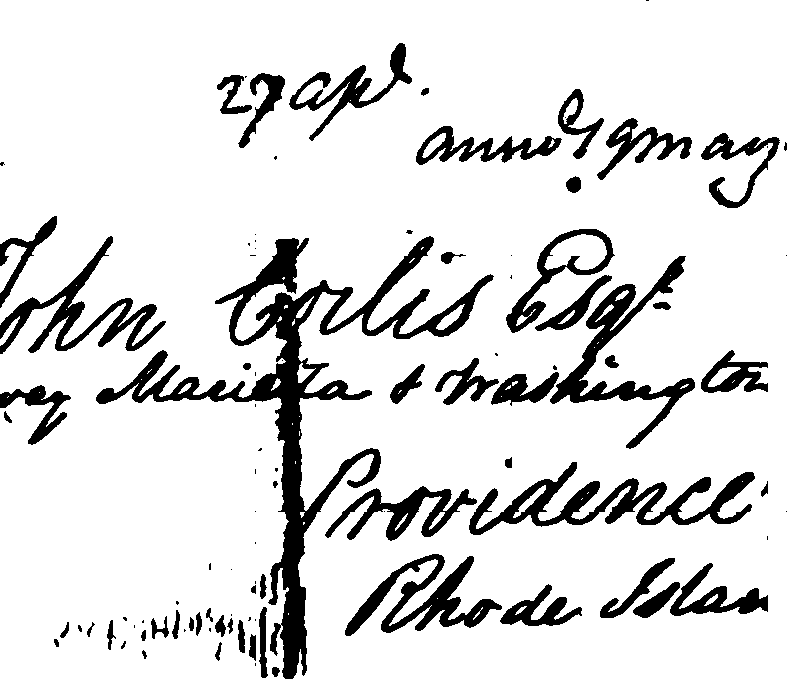}}&
{\includegraphics[width=0.45\columnwidth]{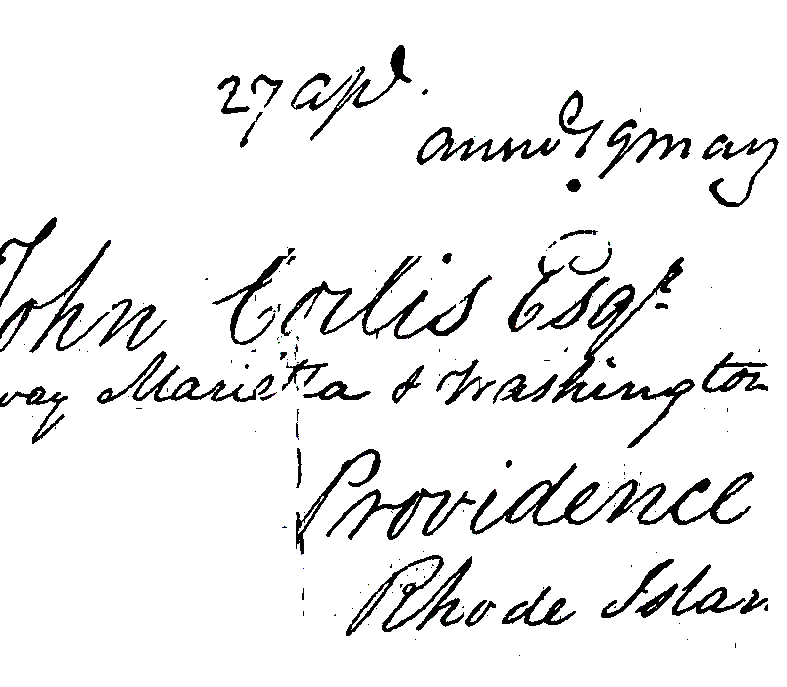}} &
{\includegraphics[width = 0.45\columnwidth]{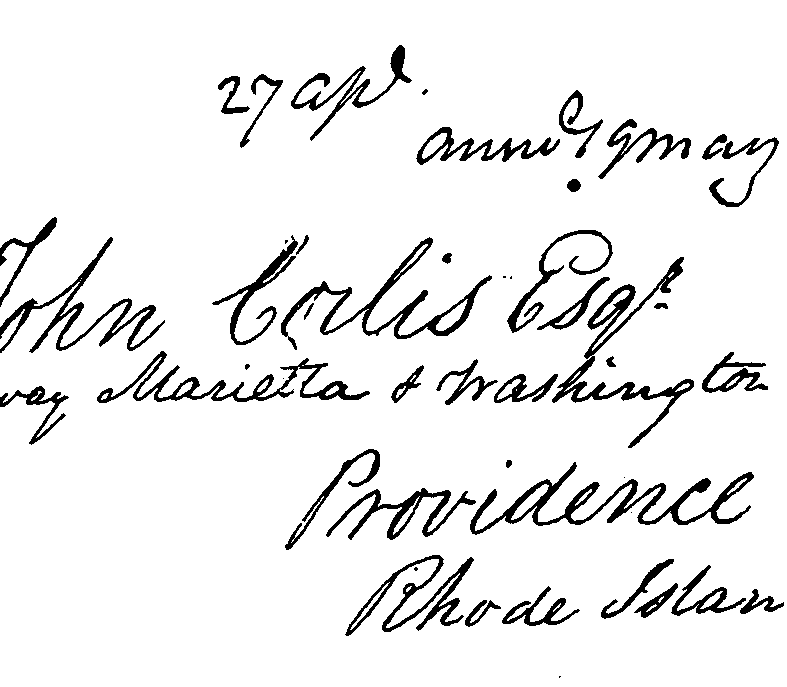}} &\\
\scriptsize e) Jemni et al.~\citep{jemni2022enhance} & \scriptsize f) DE-GAN~\citep{souibgui2020gan}  & \scriptsize g) DocEnTr~\citep{souibgui2022docentr}  &  \scriptsize h) \textbf{T2T-BinFormer}\\
\end{tabular}
\caption{Qualitative performance of the various binarization techniques on sample no. 6 from the DIBCO 2011 dataset.}
\label{fig:compDIBCO2011}
\end{figure*}

\begin{figure*}[ht!]
\centering
\begin{tabular}{ccccc}
{\includegraphics[width = 0.45\columnwidth]{images/DIBCO_Images/2012/13.png}} &
{\includegraphics[width=0.45\columnwidth]{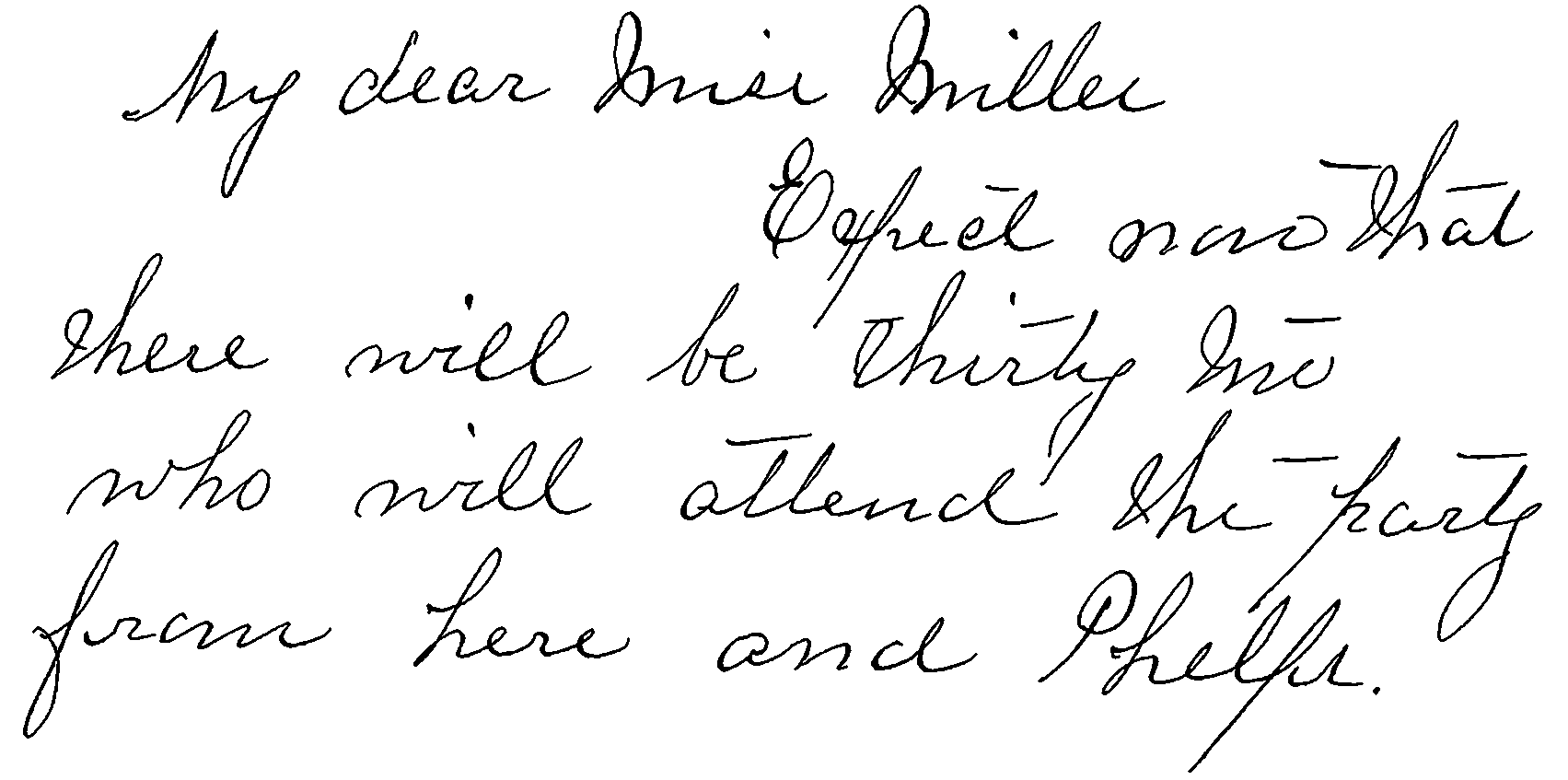}} &
{\includegraphics[width=0.45\columnwidth]{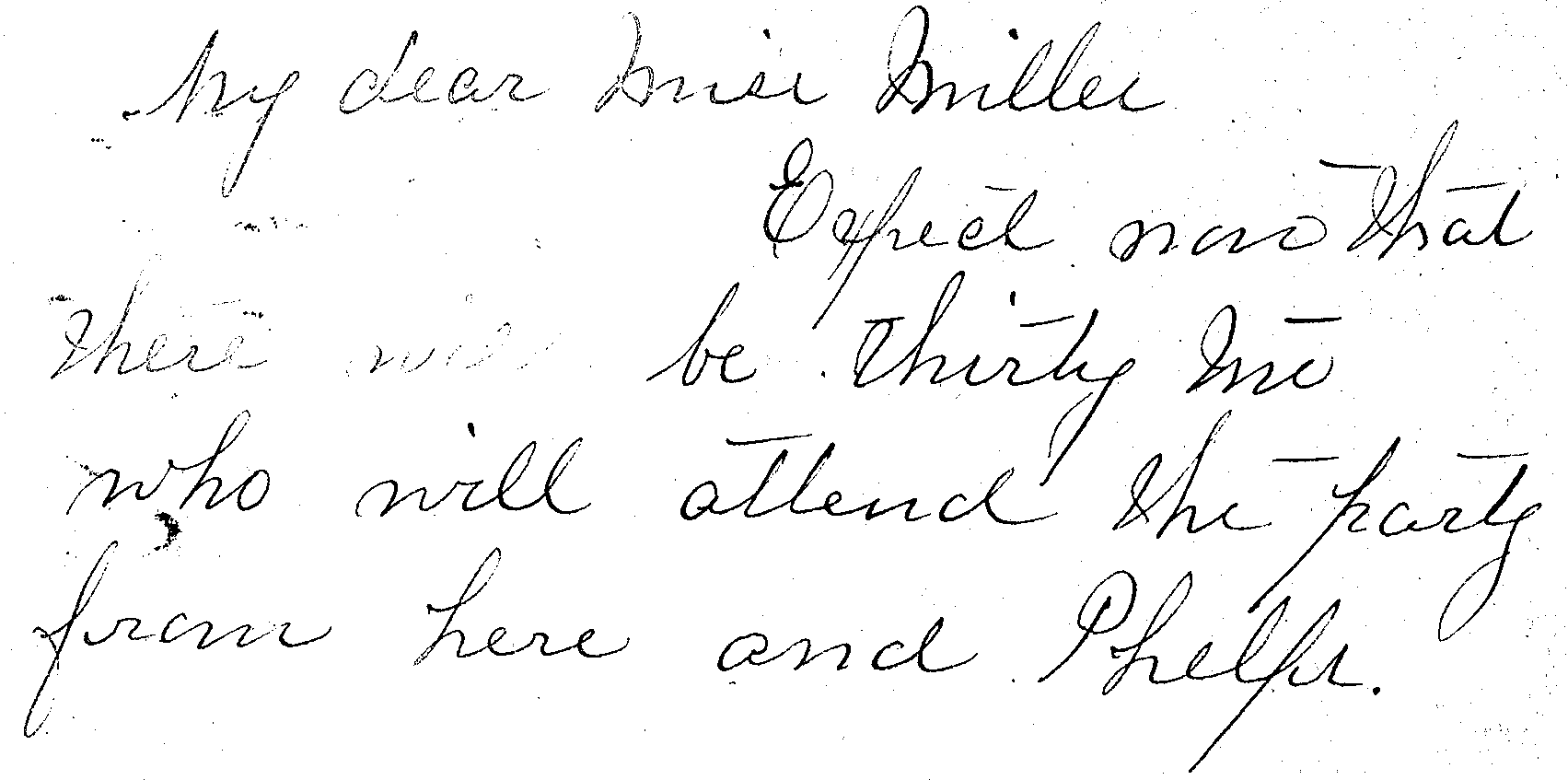}} &
{\includegraphics[width = 0.45\columnwidth]{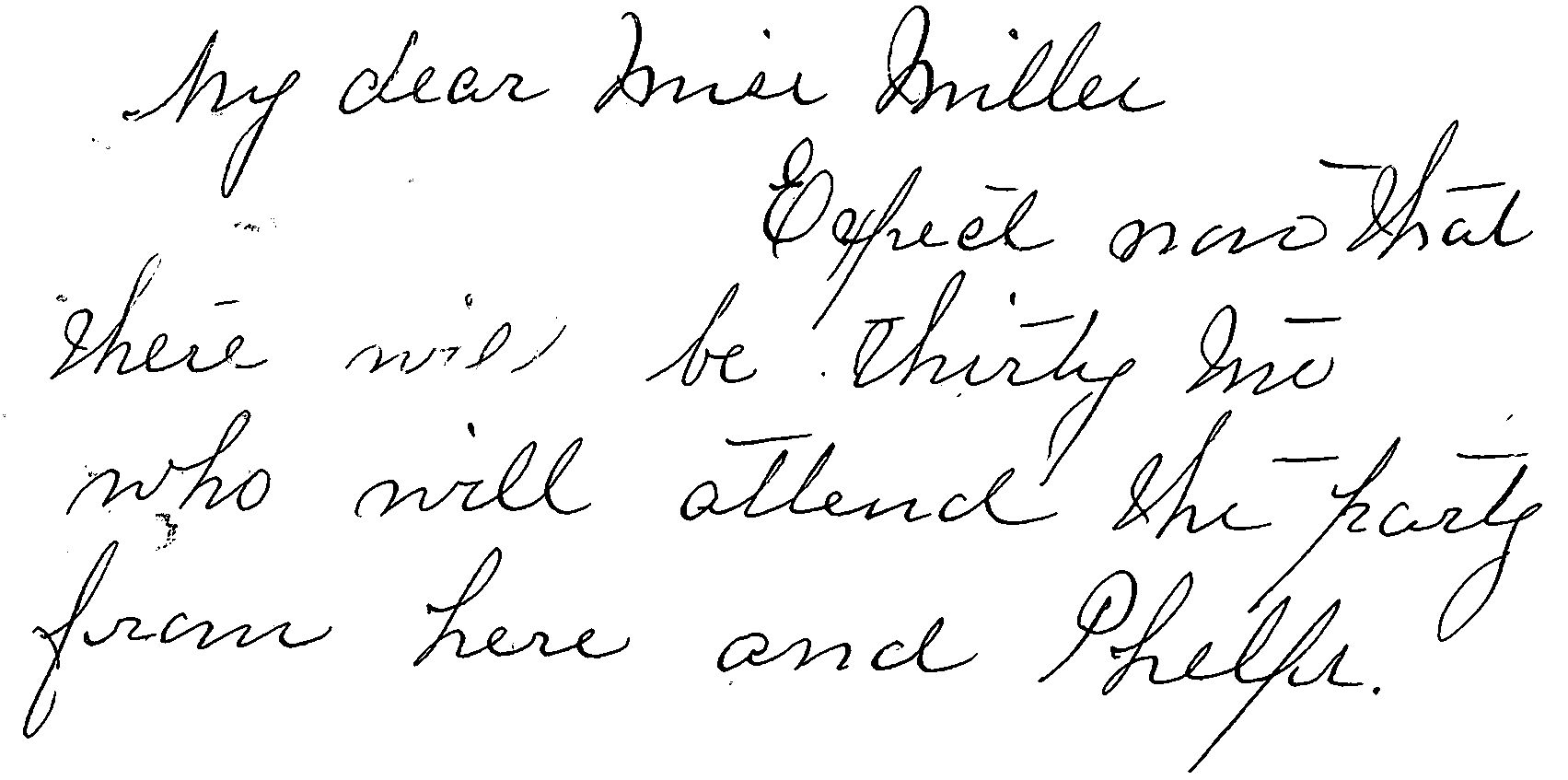}} & \\
\scriptsize a) Original &  \scriptsize b) GT  &  \scriptsize c) Otsu~\citep{otsu1979threshold}  &  \scriptsize d) SauvolaNet~\citep{li2021sauvolanet} \\ 
{\includegraphics[width=0.45\columnwidth]{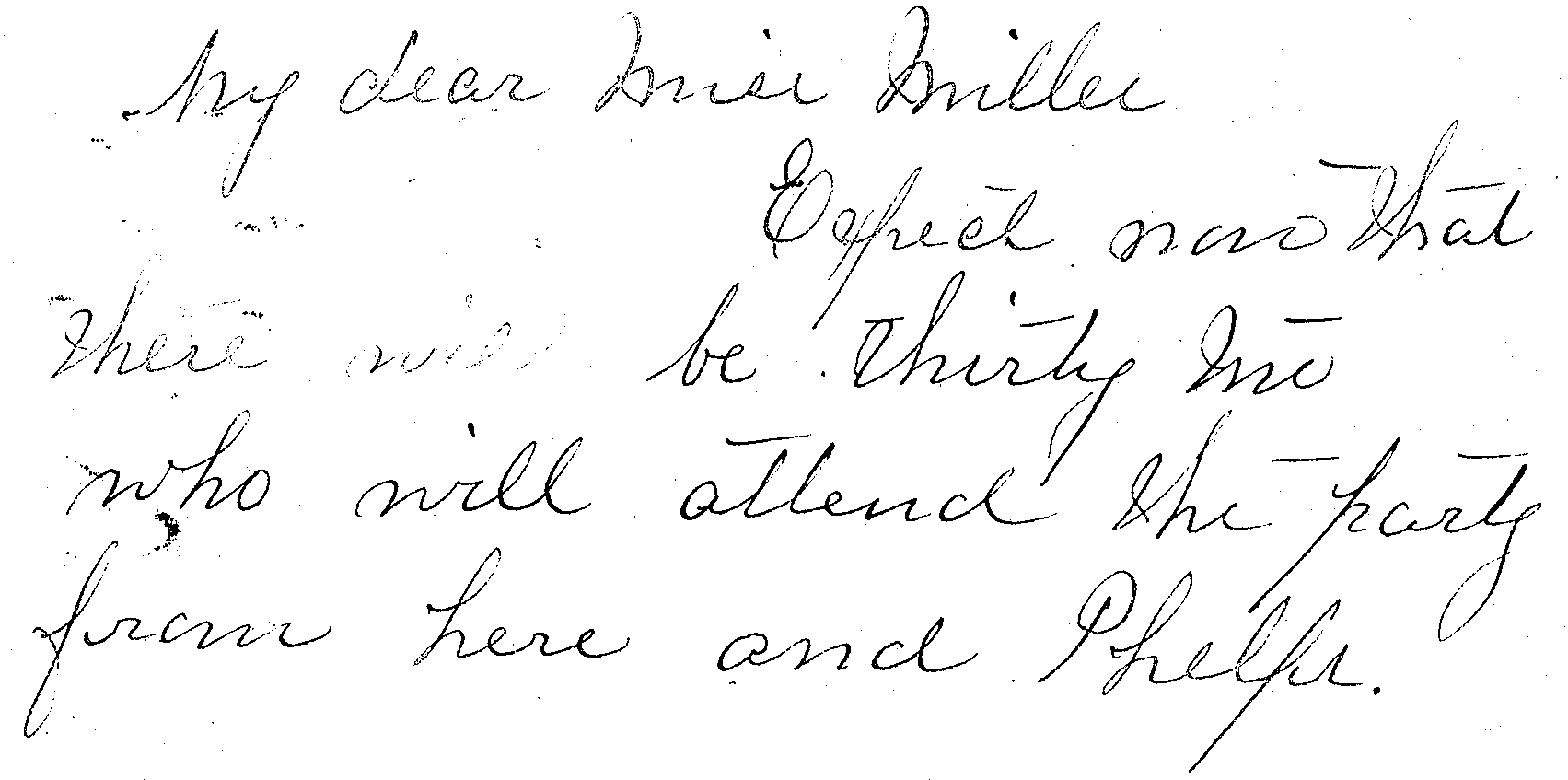}} &
{\includegraphics[width=0.45\columnwidth]{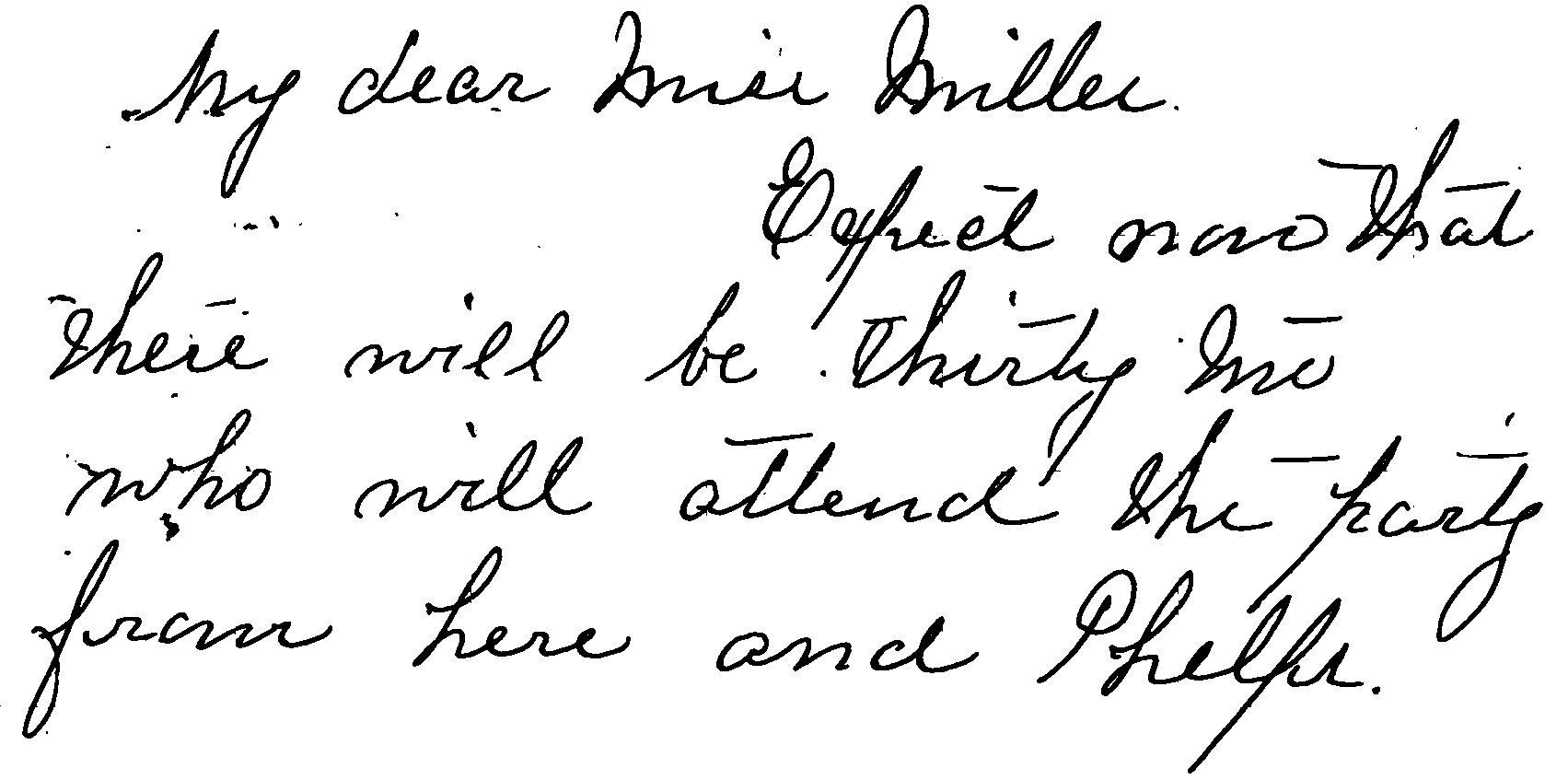}}&
{\includegraphics[width=0.45\columnwidth]{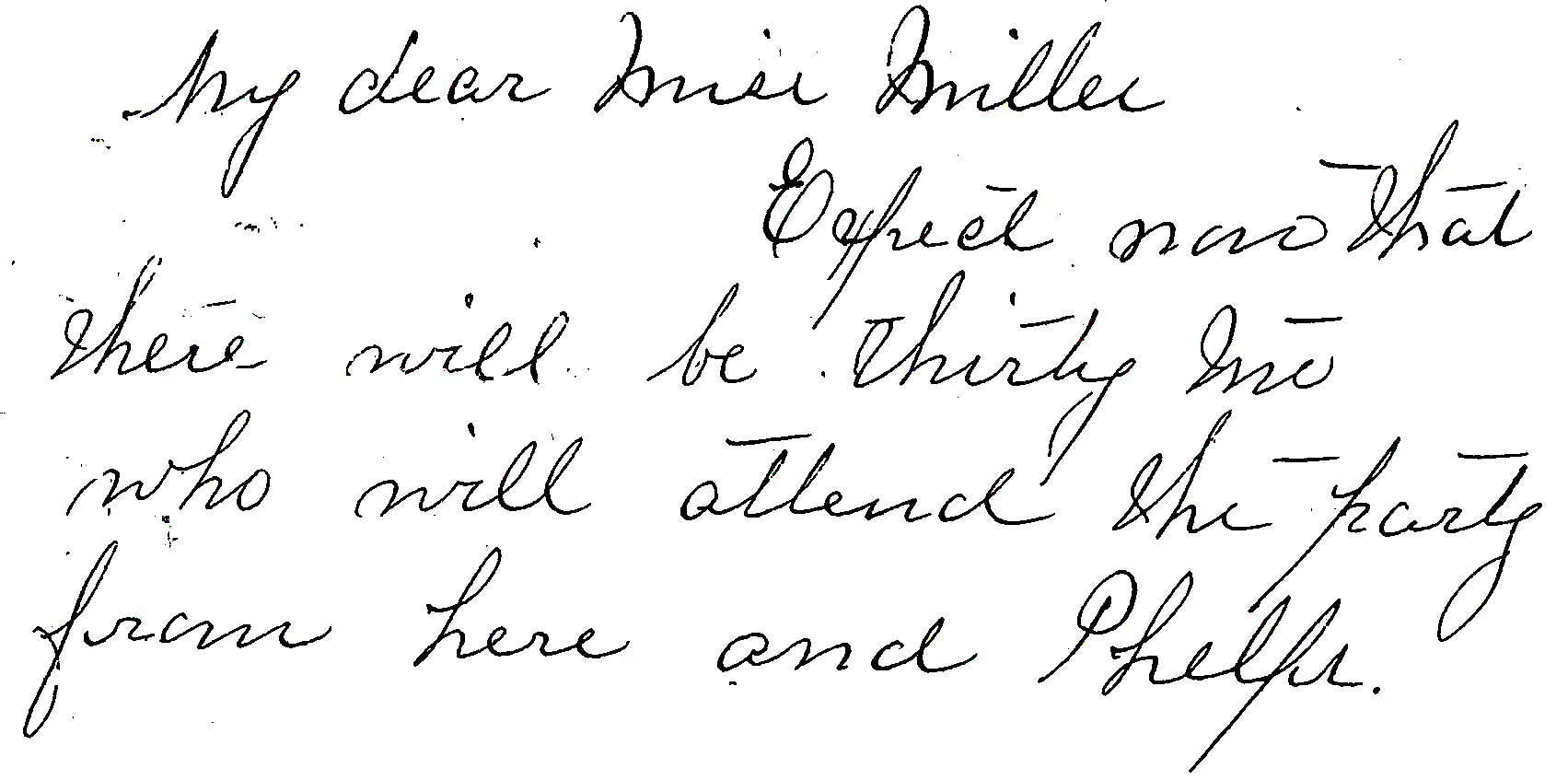}} &
{\includegraphics[width = 0.45\columnwidth]{images/T2T_Output/2012/13_pred.png}} & \\
\scriptsize e) Bradley~\citep{bradley2007adaptive} & \scriptsize f) DE-GAN~\citep{souibgui2020gan}  & \scriptsize g) DocEnTr~\citep{souibgui2022docentr}  &  \scriptsize h) \textbf{T2T-BinFormer} \\
\end{tabular}
\caption{Qualitative performance of the various binarization techniques on sample no. 13 from the DIBCO 2012 dataset.}
\label{fig:compDIBCO2012}
\end{figure*}

\begin{figure*}[ht]
\centering
\begin{tabular}{ccccc}
{\includegraphics[width = 0.45\columnwidth]{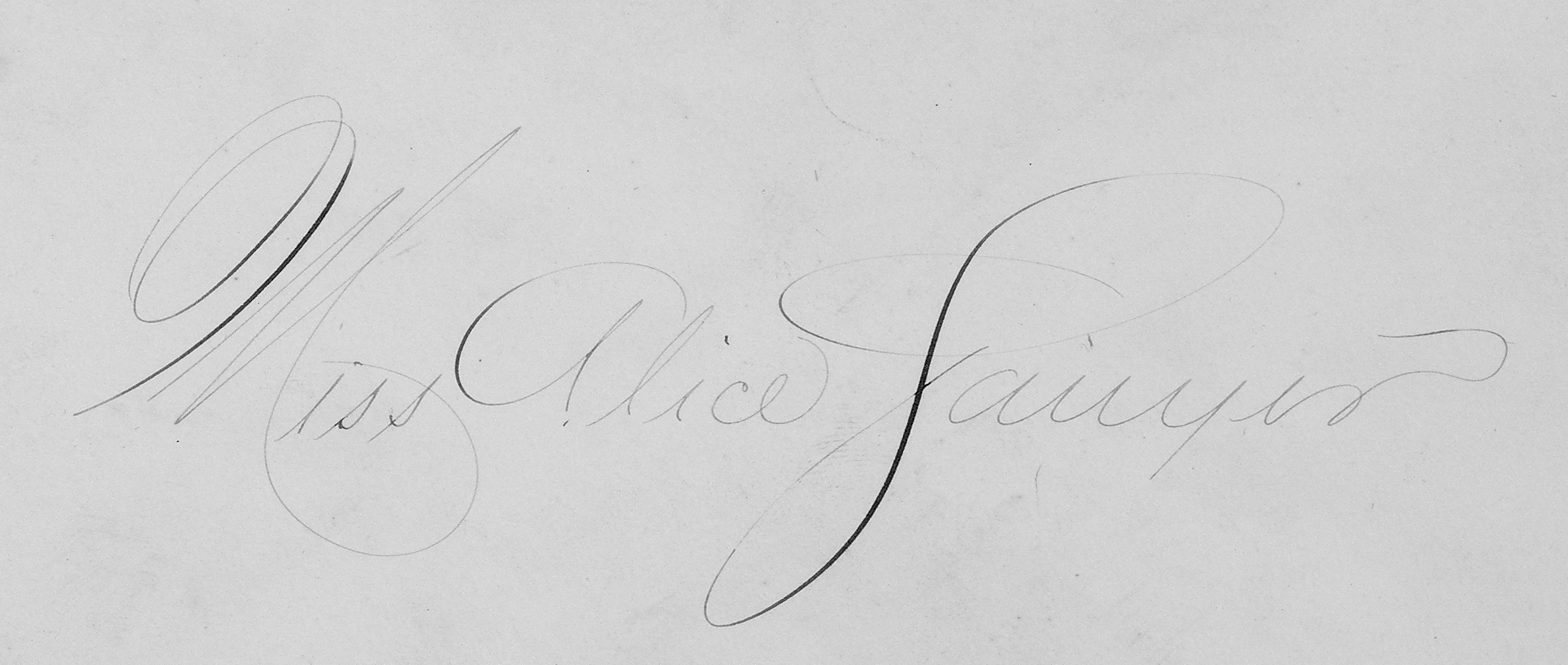}} &
{\includegraphics[width=0.45\columnwidth]{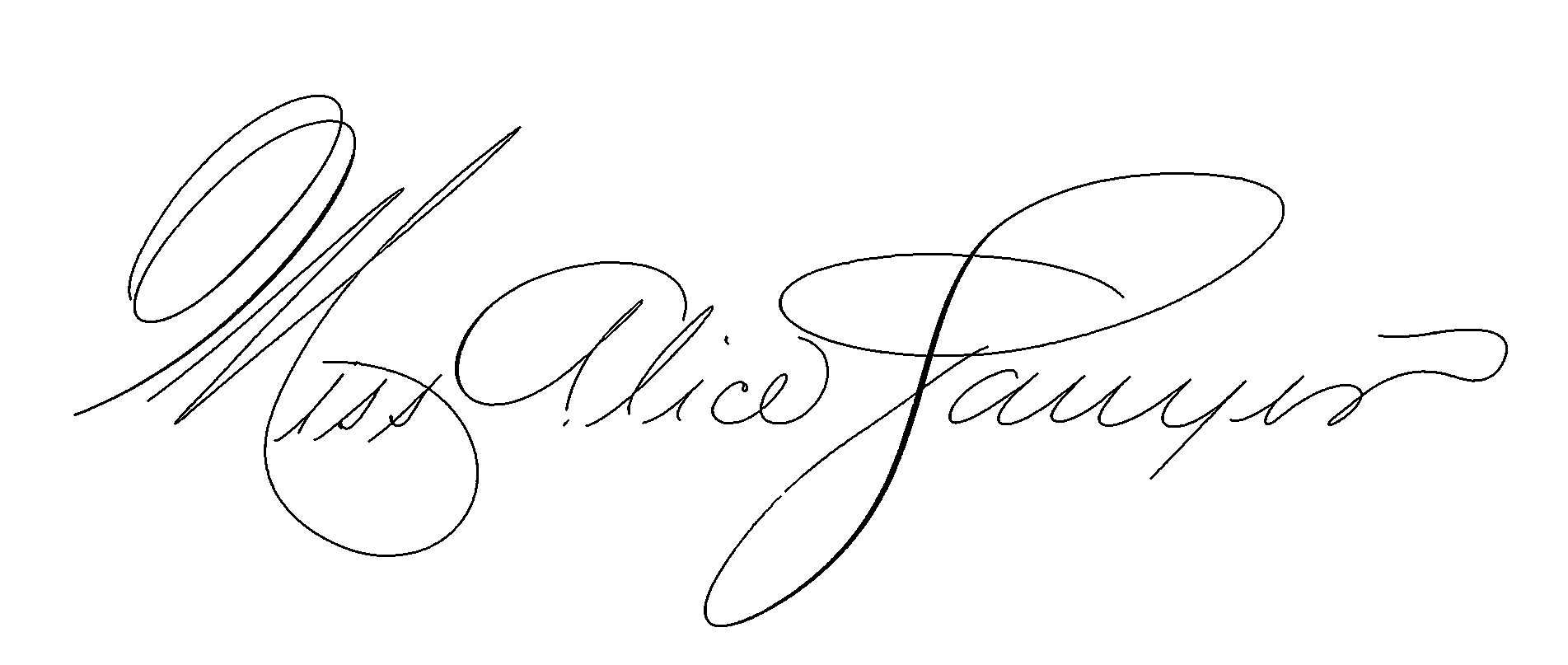}} &
{\includegraphics[width=0.45\columnwidth]{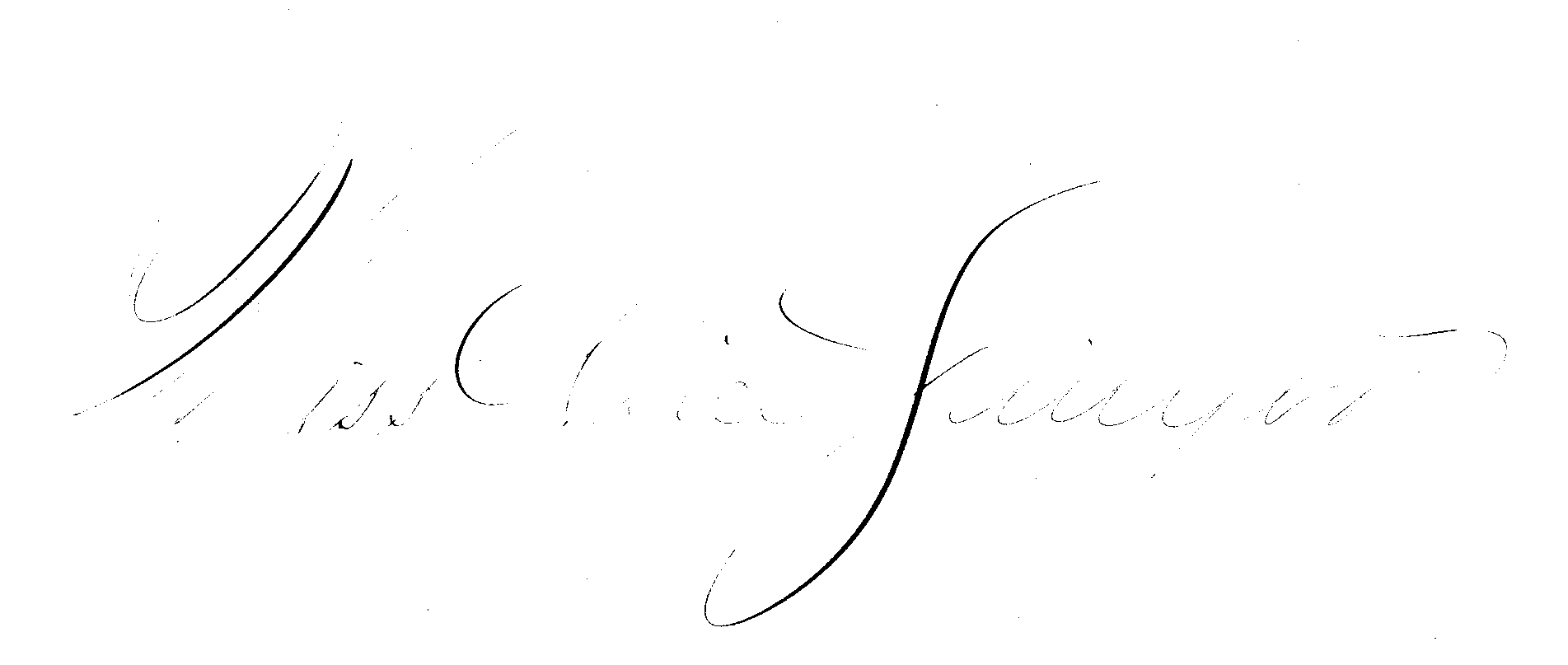}} &
{\includegraphics[width = 0.45\columnwidth]{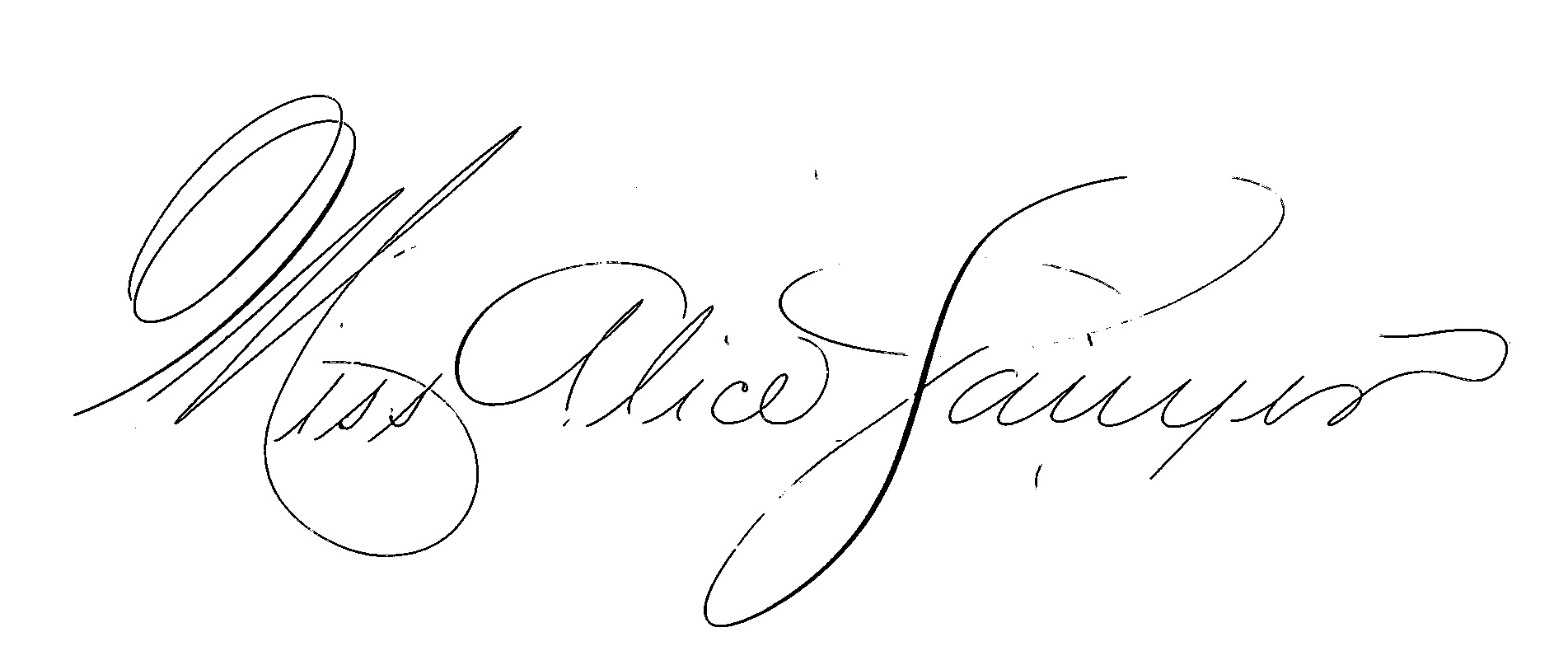}} & \\
\scriptsize a) Original &  \scriptsize b) GT  &  \scriptsize c) Otsu~\citep{otsu1979threshold}  &  \scriptsize d) SauvolaNet~\citep{li2021sauvolanet} \\ 
{\includegraphics[width=0.45\columnwidth]{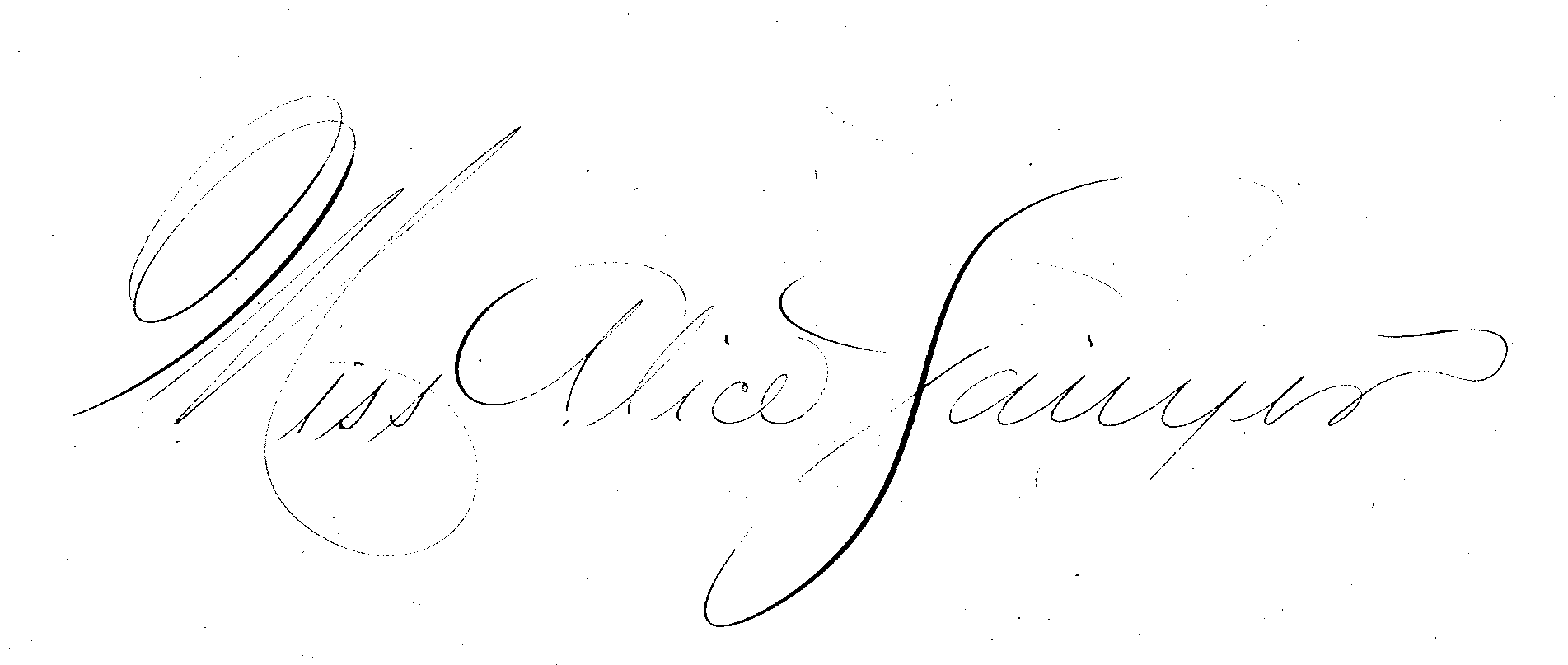}} &
{\includegraphics[width=0.45\columnwidth]{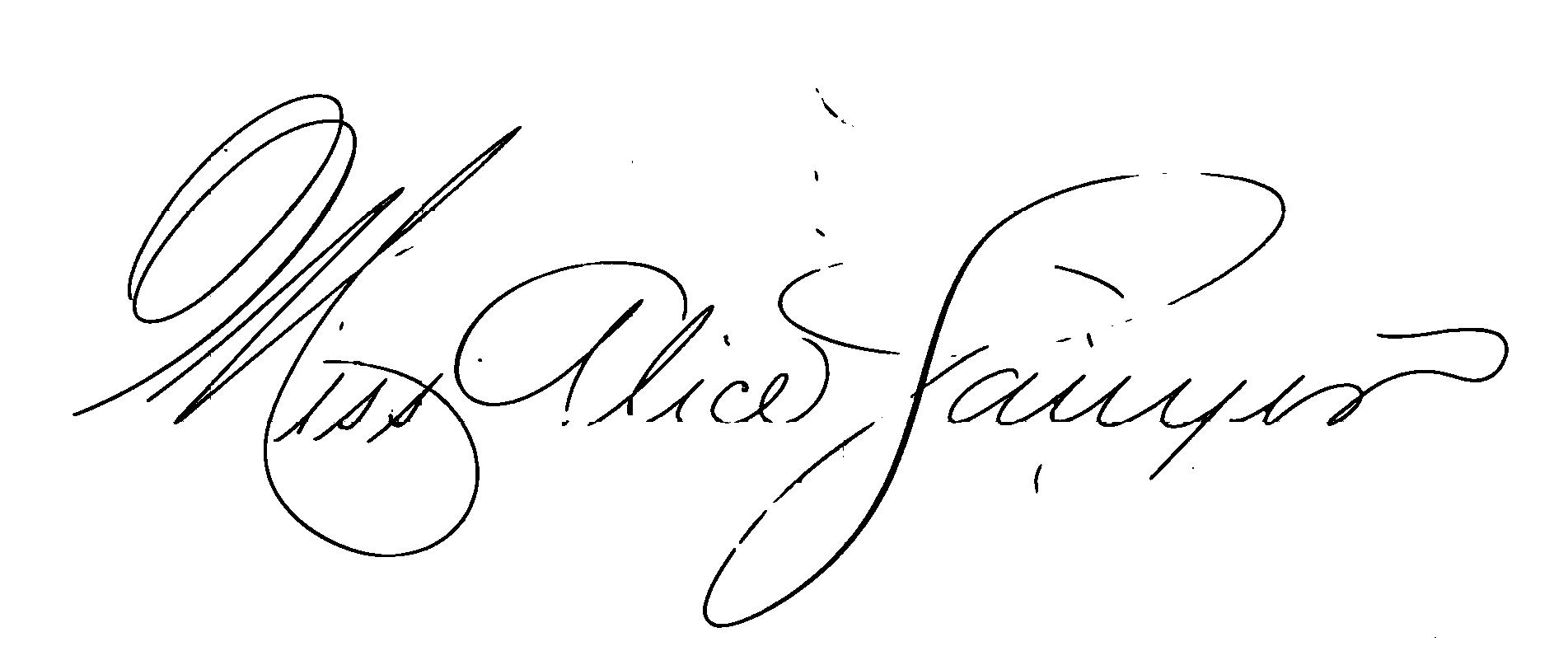}}&
{\includegraphics[width=0.45\columnwidth]{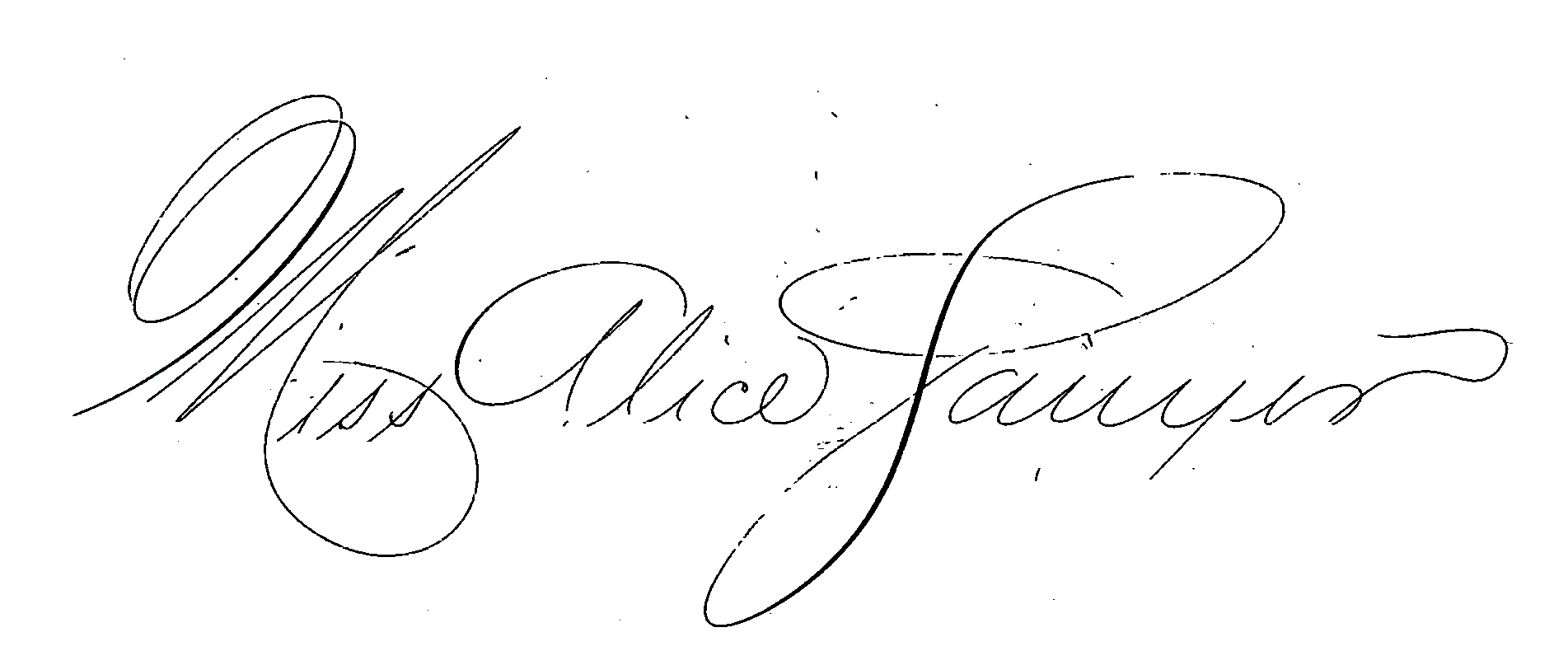}} &
{\includegraphics[width = 0.45\columnwidth]{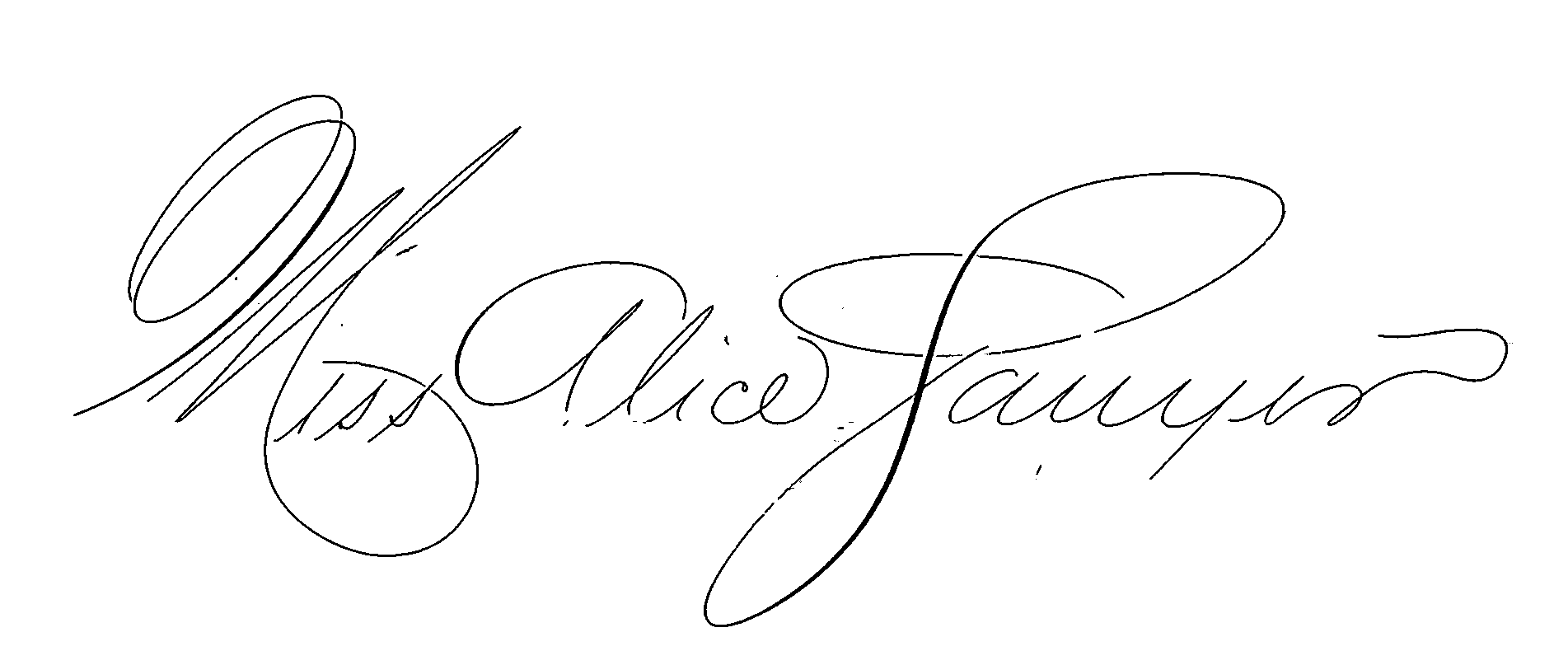}} & \\
\scriptsize e) Bradley~\citep{bradley2007adaptive} & \scriptsize f) DE-GAN~\citep{souibgui2020gan}  & \scriptsize g) DocEnTr~\citep{souibgui2022docentr}  &  \scriptsize h) \textbf{T2T-BinFormer} \\
\end{tabular}
\caption{Qualitative performance of the various binarization techniques on sample no. 7 from the DIBCO 2013 dataset.}
\label{fig:compDIBCO2013}
\end{figure*}

\begin{figure*}[ht!]
\centering
\begin{tabular}{ccccc}
{\includegraphics[width = 0.45\columnwidth]{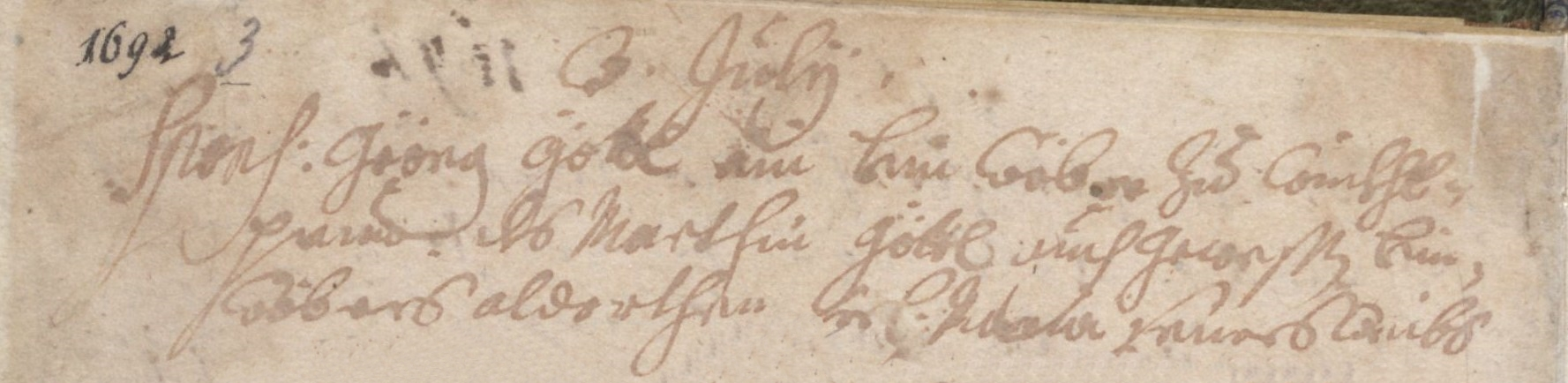}} &
{\includegraphics[width=0.45\columnwidth]{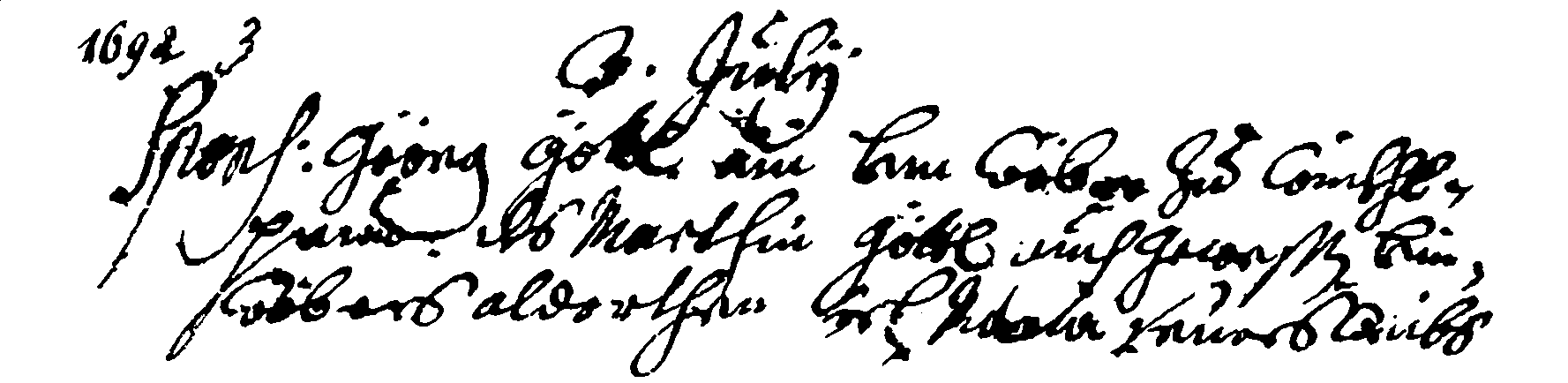}} &
{\includegraphics[width=0.45\columnwidth]{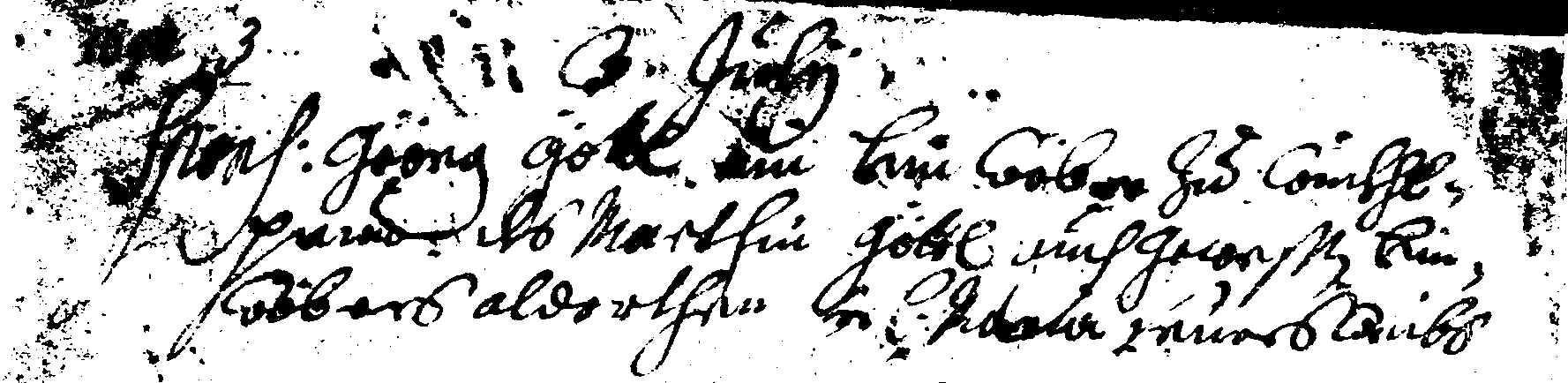}} &
{\includegraphics[width = 0.45\columnwidth]{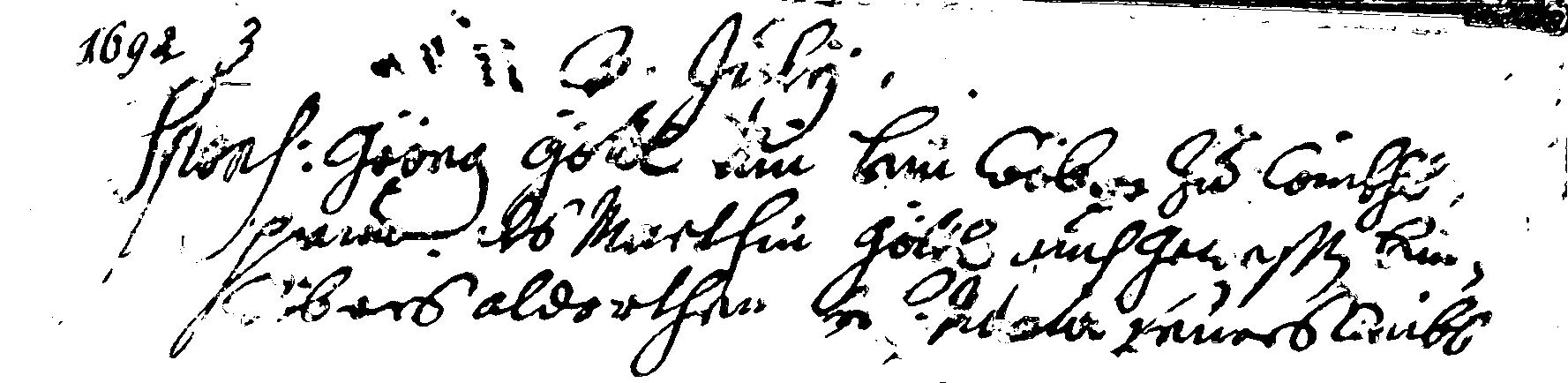}} & \\
\scriptsize a) Original &  \scriptsize b) GT  &  \scriptsize c) Otsu~\citep{otsu1979threshold}  &  \scriptsize d) SauvolaNet~\citep{li2021sauvolanet} \\ 
{\includegraphics[width=0.45\columnwidth]{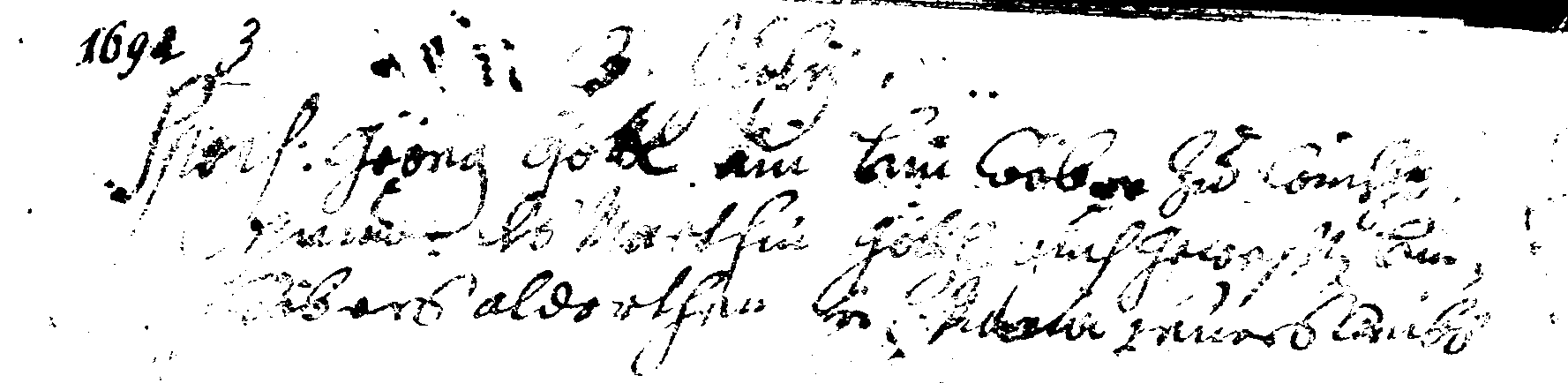}} &
{\includegraphics[width=0.45\columnwidth]{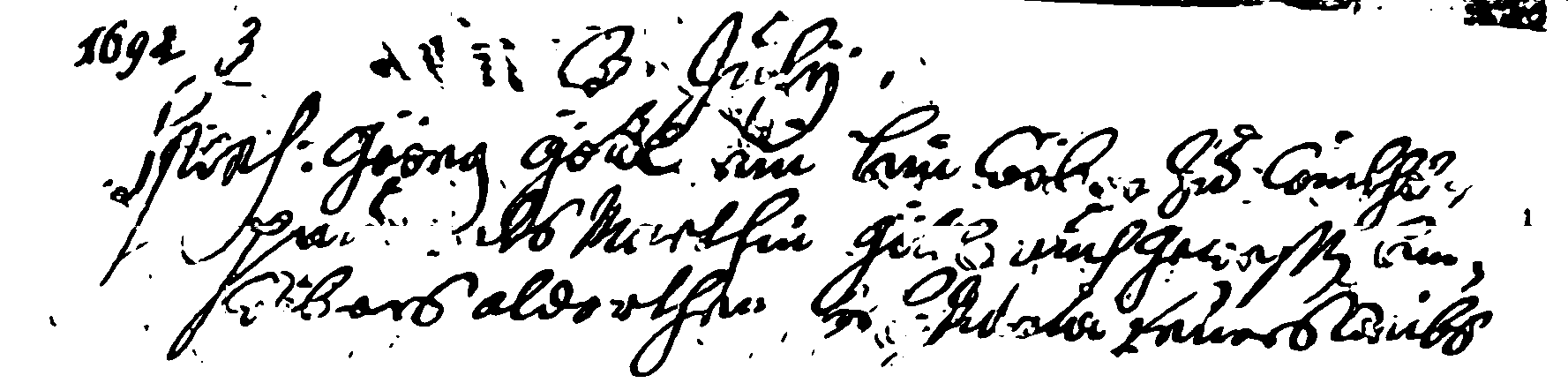}}&
{\includegraphics[width=0.45\columnwidth]{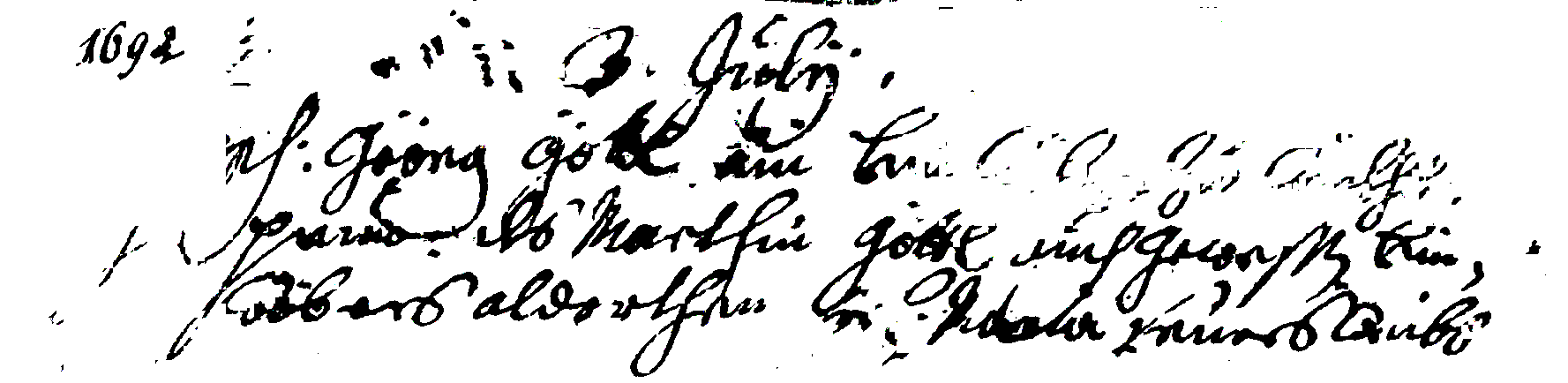}} &
{\includegraphics[width = 0.45\columnwidth]{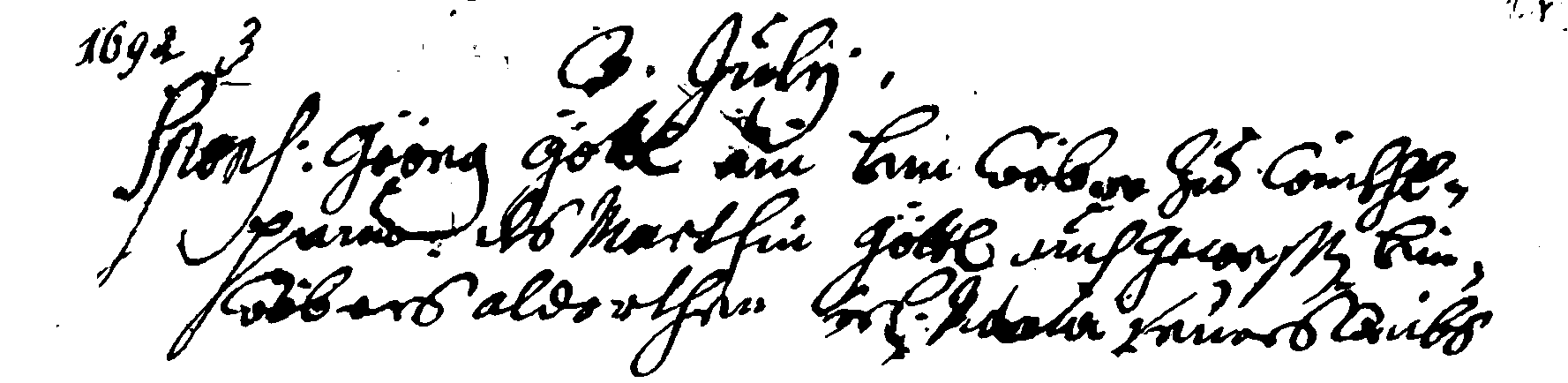}} & \\
\scriptsize e) Bradley~\citep{bradley2007adaptive} & \scriptsize f) DE-GAN~\citep{souibgui2020gan}  & \scriptsize g) DocEnTr~\citep{souibgui2022docentr}  &  \scriptsize h) \textbf{T2T-BinFormer} \\
\end{tabular}
\caption{Qualitative performance of the various binarization techniques on sample no. 1 from the DIBCO 2018 dataset.}
\label{fig:compDIBCO2018_1}
\end{figure*}

\begin{figure*}[ht!]
\centering
\begin{tabular}{ccccc}
{\includegraphics[width = 0.45\columnwidth]{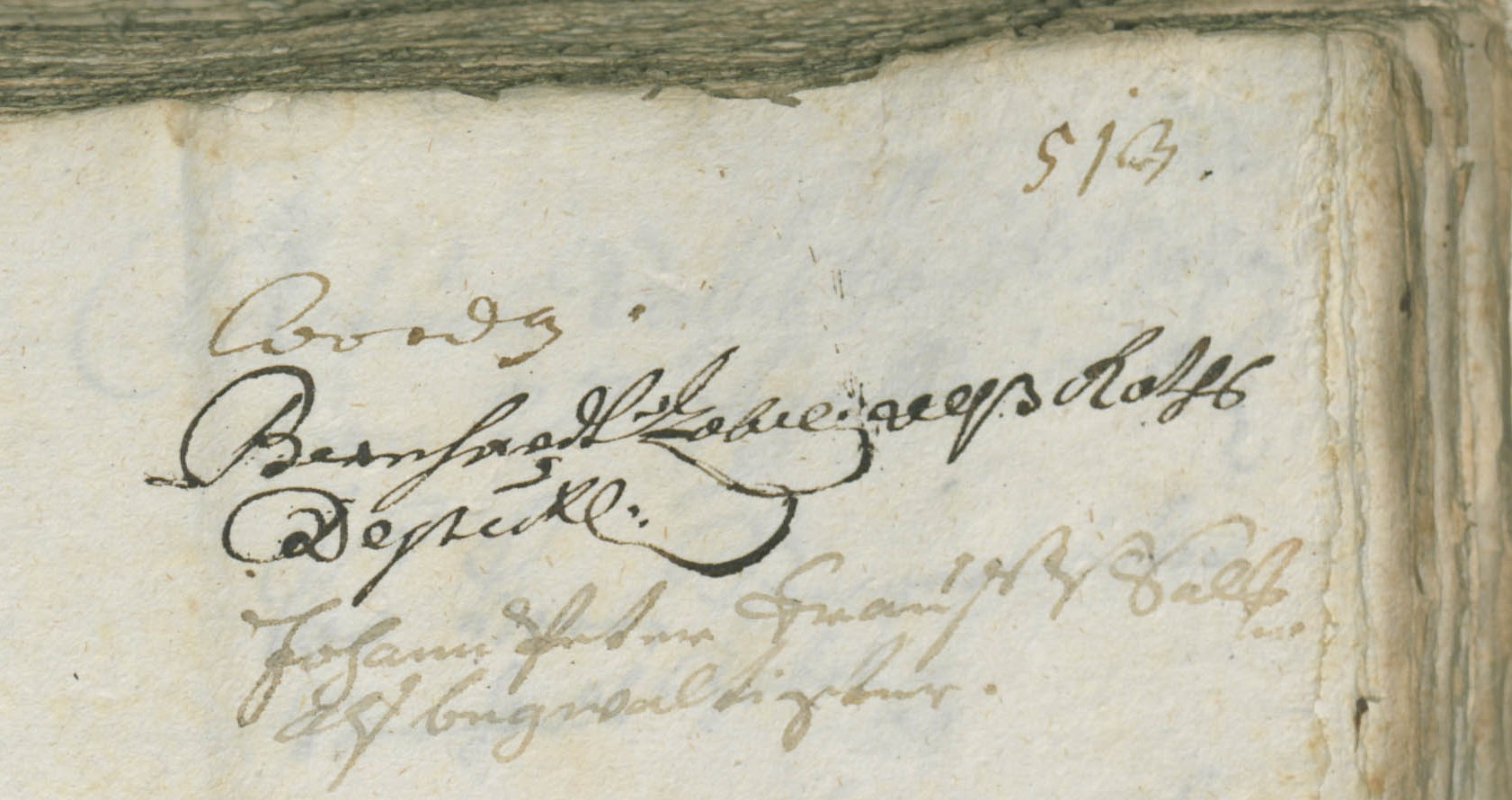}} &
{\includegraphics[width=0.45\columnwidth]{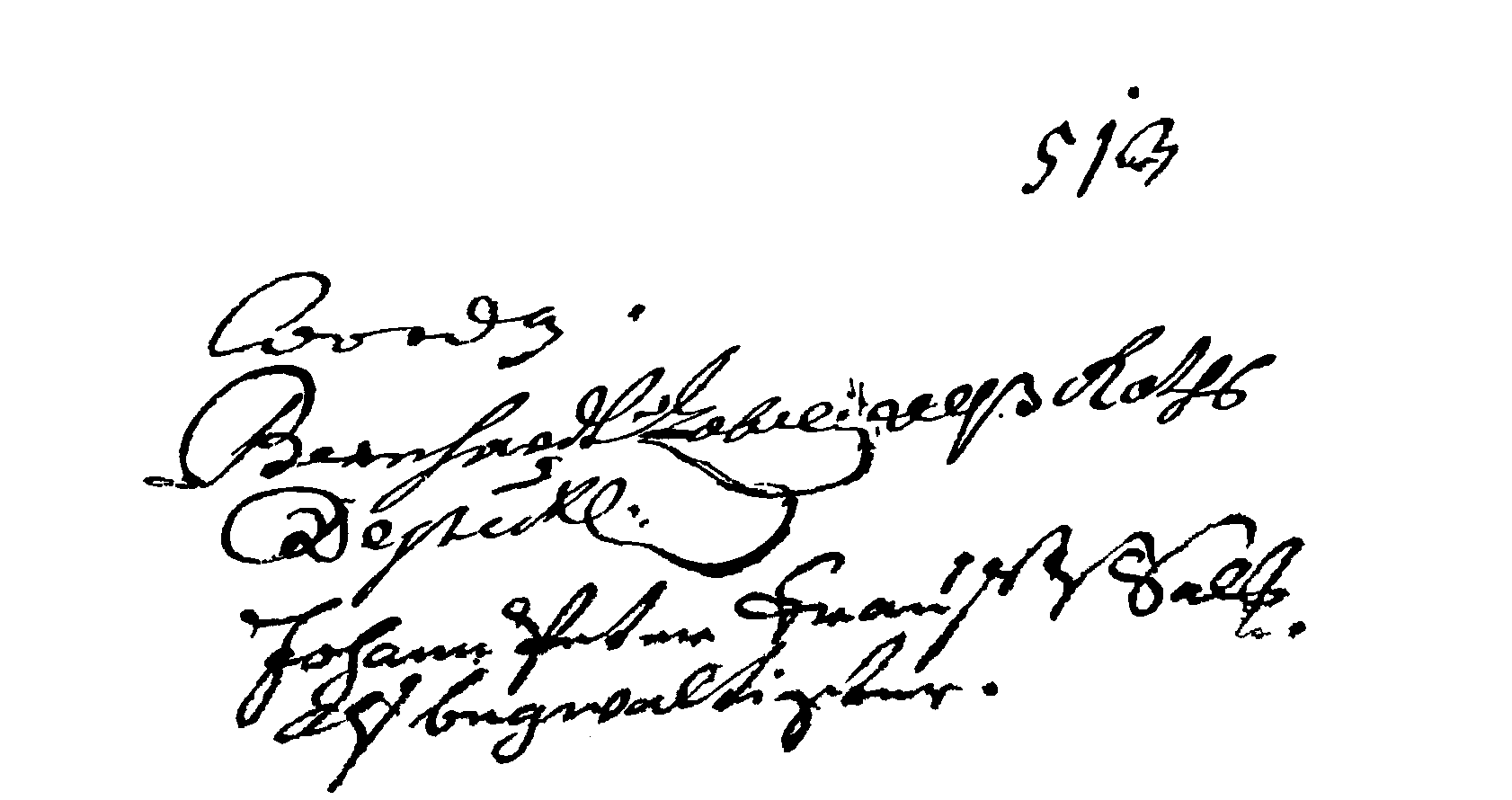}} &
{\includegraphics[width=0.45\columnwidth]{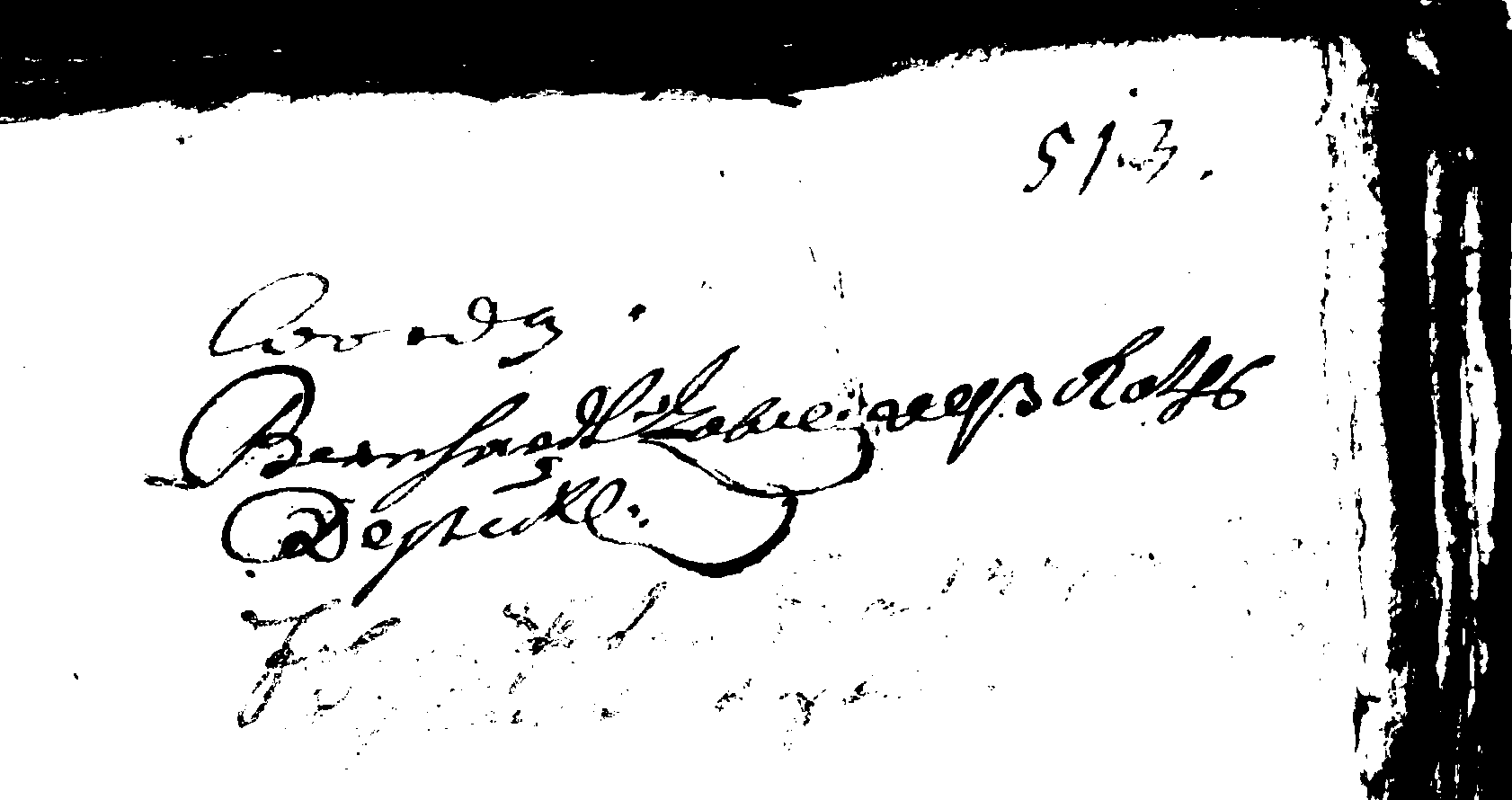}} &
{\includegraphics[width = 0.45\columnwidth]{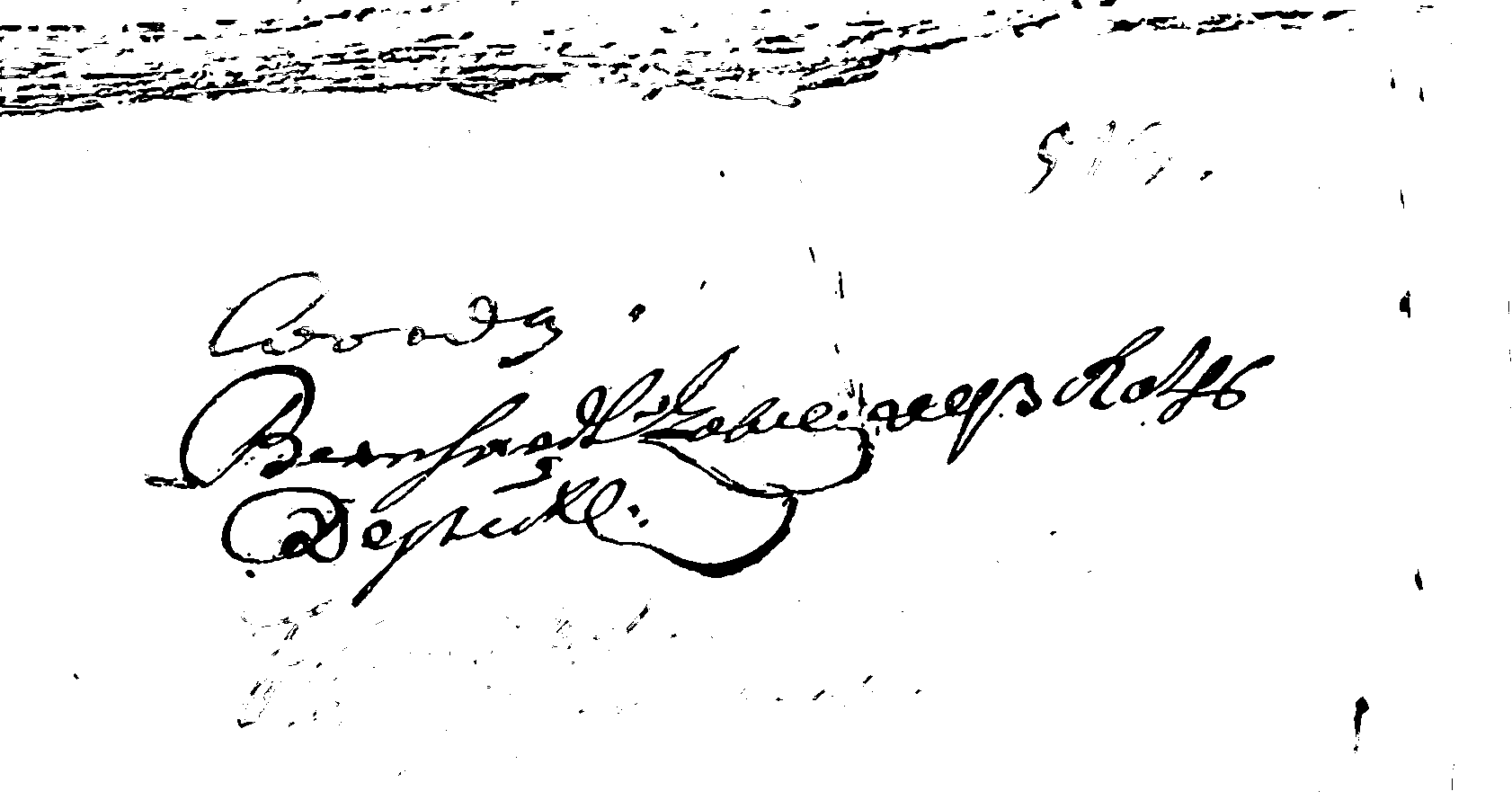}} & \\
\scriptsize a) Original &  \scriptsize b) GT  &  \scriptsize c) Otsu~\citep{otsu1979threshold}  &  \scriptsize d) SauvolaNet~\citep{li2021sauvolanet} \\ 
{\includegraphics[width=0.45\columnwidth]{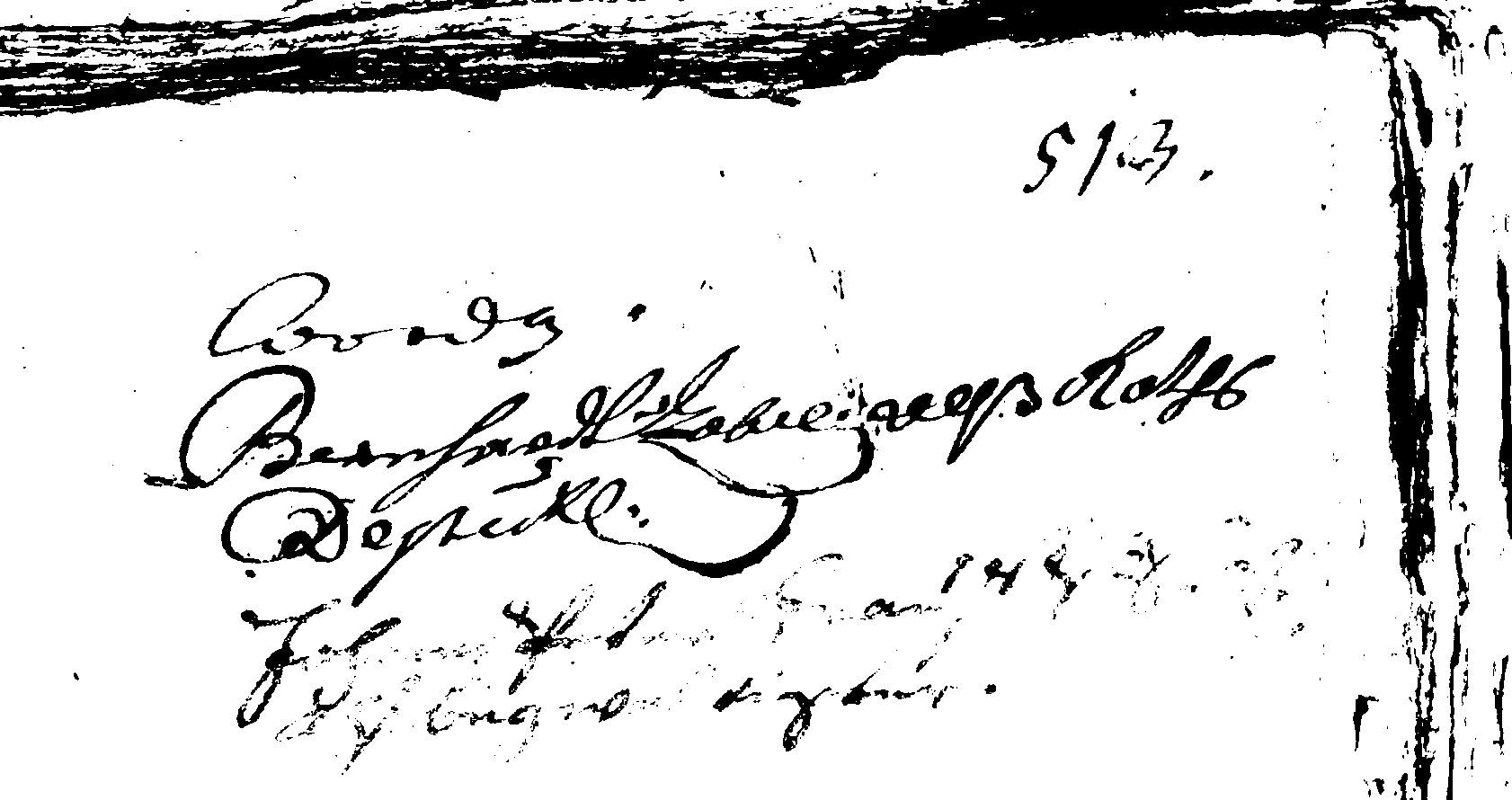}} &
{\includegraphics[width=0.45\columnwidth]{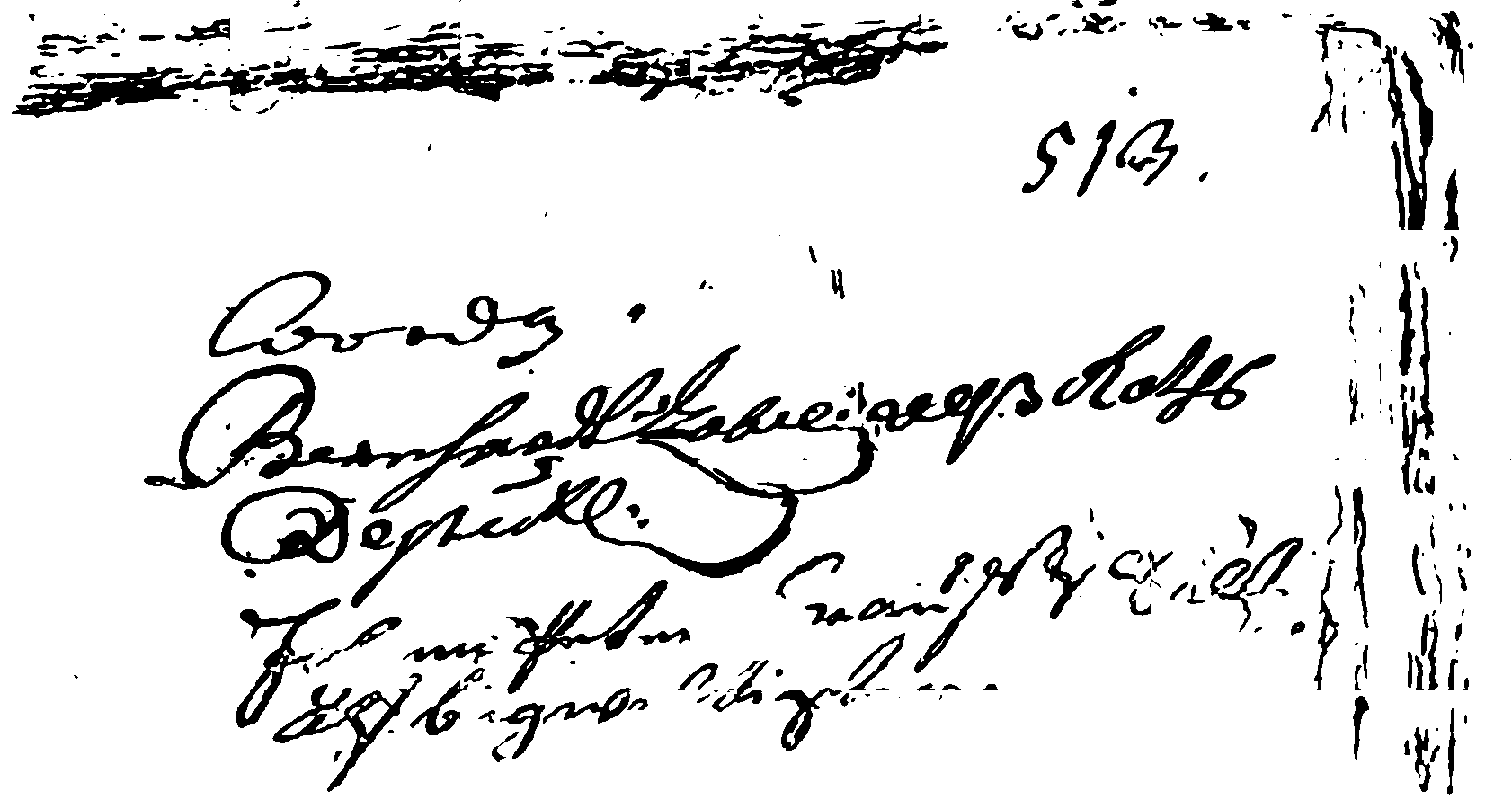}}&
{\includegraphics[width=0.45\columnwidth]{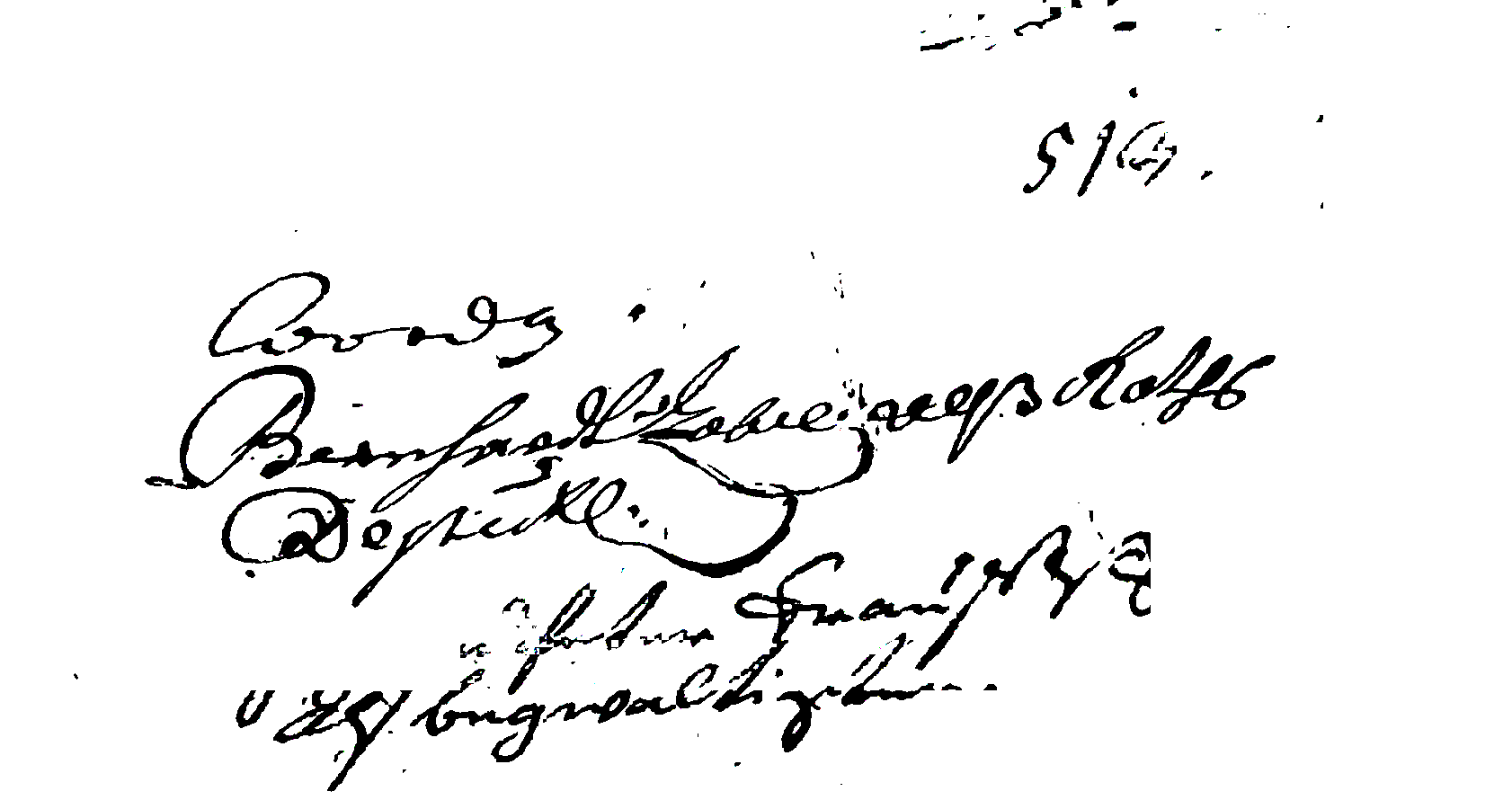}} &
{\includegraphics[width = 0.45\columnwidth]{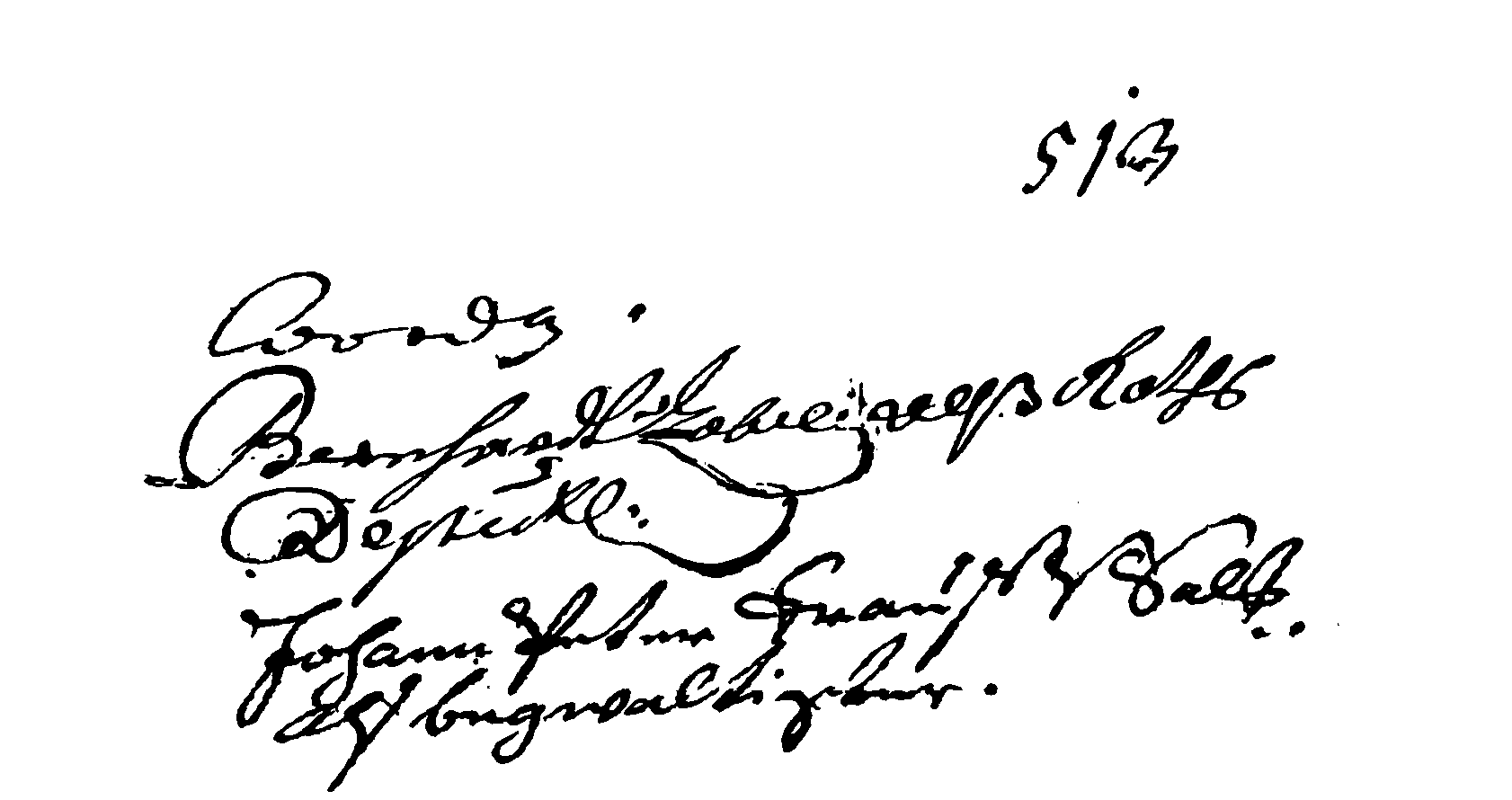}} & \\
\scriptsize e) Bradley~\citep{bradley2007adaptive} & \scriptsize f) DE-GAN~\citep{souibgui2020gan}  & \scriptsize g) DocEnTr~\citep{souibgui2022docentr}  &  \scriptsize h) \textbf{T2T-BinFormer} \\
\end{tabular}
\caption{Qualitative performance of the various binarization techniques on sample no. 6 from the DIBCO 2018 dataset.}
\label{fig:compDIBCO2018_2}
\end{figure*}

\begin{figure*}[ht!]
\centering
\captionsetup{justification=centering}
\begin{tabular}{cccc}
\includegraphics[width=0.45\columnwidth,scale=0.25]{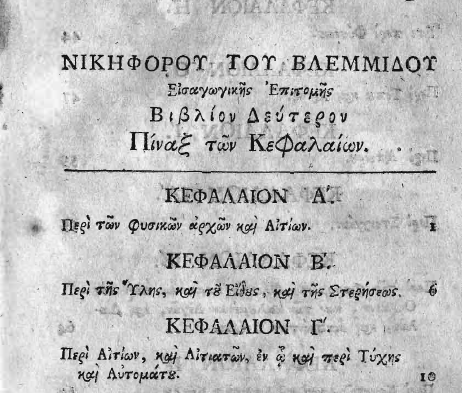} &
\includegraphics[width=0.45\columnwidth,scale=0.25]{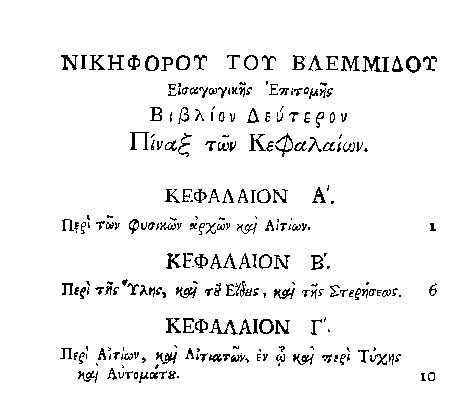} &
\includegraphics[width=0.45\columnwidth,scale=0.25]{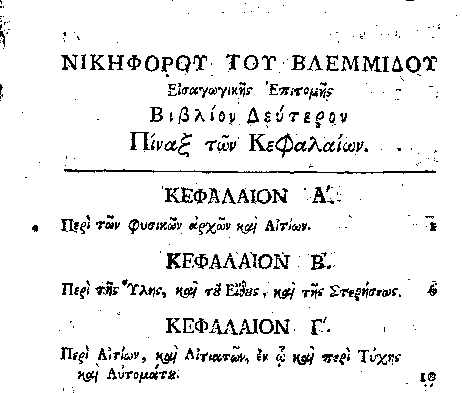} &
\includegraphics[width=0.45\columnwidth,scale=0.25]{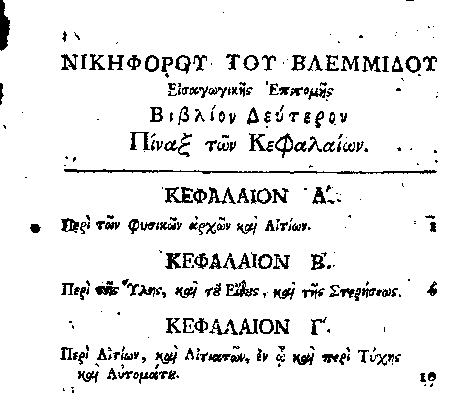} 
\\  \scriptsize a) Original &  \scriptsize b) GT  &  \scriptsize c) Otsu~\citep{otsu1979threshold}  &  \scriptsize d) Gatos~\citep{gatos2006adaptive}\\  
\includegraphics[width=0.45\columnwidth,scale=0.25]{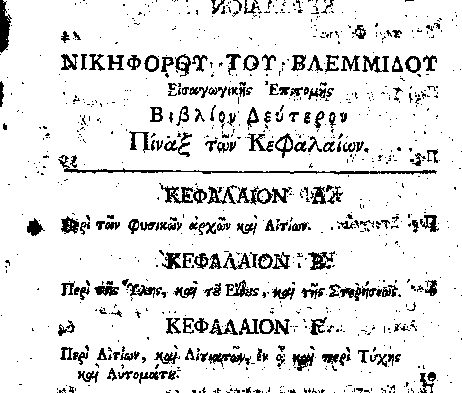} &
\includegraphics[width=0.45\columnwidth,scale=0.25]{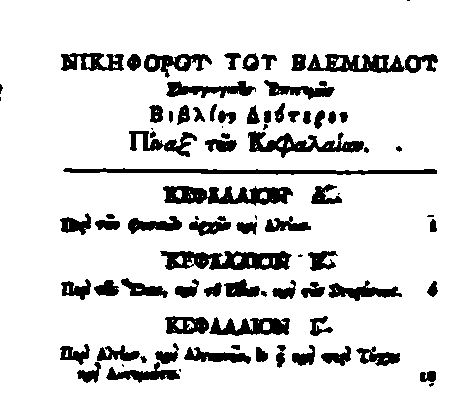} &
\includegraphics[width=0.45\columnwidth,scale=0.25]{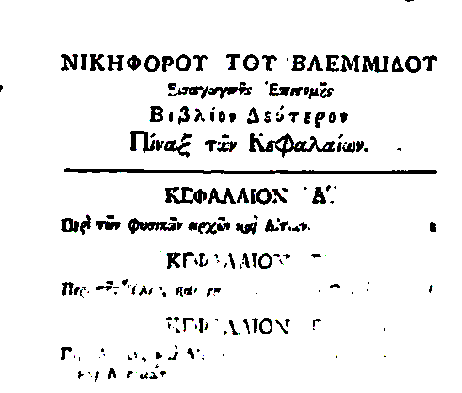} &
\includegraphics[width=0.45\columnwidth,scale=0.25]{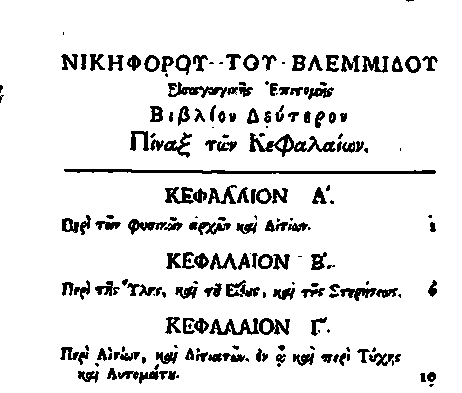} 
\\ \scriptsize e) Bradley~\citep{bradley2007adaptive} & \scriptsize f) DE-GAN~\citep{souibgui2020gan}  & \scriptsize g) DocEnTr~\citep{souibgui2022docentr}  &  \scriptsize h) \textbf{T2T-BinFormer}\\
\end{tabular}
\caption{{Qualitative performance of the various binarization techniques on sample no. 2 from the DIBCO 2019 dataset.}}
\label{fig:compDIBCO2019}
\end{figure*}

Subsection~\ref{sub_sec:QuantAnalysis} provided the quantitative results produced by the proposed model and did a comparative analysis with the existing binarization approaches. In this subsection, we will provide a qualitative evaluation. We begin the qualitative analysis by providing an intuitive comparison between the ground truth (GT) image and the output of the proposed model on handwritten document images taken from different DIBCO datasets. We chose various handwritten document images with varying types and degrees of degradation and compared the binarized output with the GT and the original document image. the illustration is shown in Fig.~\ref{fig:Output_Handwritten} and \ref{fig:Output_Types}, respectively. Next, we perform a qualitative analysis on samples taken from DIBCO 2009, 2010, 2011, 2012, 2013, 2018, and 2019 datasets. The comparative results are shown in Fig.~\ref{fig:compDIBCO2009},~\ref{fig:compDIBCO2010}, ~\ref{fig:compDIBCO2011},~\ref{fig:compDIBCO2012}, ~\ref{fig:compDIBCO2013},~\ref{fig:compDIBCO2018_1}, ~\ref{fig:compDIBCO2018_2}, and ~\ref{fig:compDIBCO2019}, respectively.        

In conclusion, the qualitative comparison shown in Fig.~\ref{fig:compDIBCO2009} to \ref{fig:compDIBCO2019} clearly demonstrates the proposed model's superiority in recovering a severely damaged picture over conventional thresholding, CNN, GAN, and current ViT-based approaches.
\section{Conclusion}
\label{sec:Con}
In this study, we introduced a Tokens-to-Token vision transformer network for document binarization. The Tokens-to-Token method is employed in the proposed network as opposed to the vanilla ViT network's straightforward tokenization. In this study, we went into great depth on how the model extracts both global and local feature dependencies and its impact on the results produced for the image binarization problem. In addition to providing a detailed description of the model's operating principles, we graphically and mathematically represented the complete proposed network. To the best of our knowledge, \textbf{T2T-BinFormer} is the first binarization method based on ViT to employ a soft split strategy for feature extraction as opposed to hard split patches, as vanilla ViT does. We conducted in-depth quantitative and qualitative experimental analysis on a variety of real-world document image binarization datasets, each of which presented its own set of challenges. The proposed model was successfully able to address almost all of these challenges and produced cutting-edge results. Numerous experiments show that the \textbf{T2T-BinFormer} model produces outcomes that are noticeably superior to those of the state-of-the-art. In our upcoming work, we want to expand this network to handle deblurring, dehazing, dewarping, and other activities related to document image enhancement. We would also keep researching the best ways to use ViT-based models to effectively extract the local information from an image.

\bibliographystyle{elsarticle-num}
\bibliography{refs}
\end{document}